\documentclass{article}

\PassOptionsToPackage{numbers, compress}{natbib}

\usepackage[preprint]{neurips_2024}

\usepackage{graphicx}
\usepackage[utf8]{inputenc} 
\usepackage[T1]{fontenc}    
\usepackage{hyperref}       
\usepackage{url}            
\usepackage{booktabs}       
\usepackage{amsfonts}       
\usepackage{nicefrac}       
\usepackage{microtype}      
\usepackage{xcolor}         
\usepackage[version=4]{mhchem}
\usepackage{multirow}
\usepackage{subcaption}

\usepackage{comment}
\usepackage{enumitem}

\title{Publishing FAIR and Machine-actionable Reviews in Materials Science: The Case for Symbolic Knowledge in Neuro-symbolic Artificial Intelligence}

\author{%
  Jennifer D'Souza \\
  TIB Leibniz Information Centre for Science and Technology\\
  Hannover, Germany\\
  \texttt{jennifer.dsouza@tib.eu}
  \And
  S{\"o}ren Auer \\
  TIB Leibniz Information Centre for Science and Technology\\
  Hannover, Germany\\
  L3S Research Center, Leibniz University Hannover\\
  Hannover, Germany
  \And
  Eleni Poupaki \\
  Eindhoven University of Technology (TU/e)\\
  Eindhoven, The Netherlands
  \And
  Alex Watkins \\
  University of Warwick\\
  Coventry, United Kingdom
  \And
  Anjana Devi \\
  Leibniz Institute for Solid State and Materials Research (IFW)\\
  Dresden, Germany
  \And
  Riikka L.~Puurunen \\
  School of Chemical Engineering\\
  Aalto University\\
  Espoo, Finland
  \And
  Bora Karasulu \\
  University of Warwick\\
  Coventry, United Kingdom
  \And
  Adrie Mackus \\
  Eindhoven University of Technology (TU/e)\\
  Eindhoven, The Netherlands
  \And
  Erwin Kessels \\
  Eindhoven University of Technology (TU/e)\\
  Eindhoven, The Netherlands
}

\begin{document}

\maketitle

\begin{abstract}
Scientific reviews are central to knowledge integration in materials science, yet their key insights remain locked in narrative text and static PDF tables, limiting reuse by humans and machines alike. This article presents a case study in atomic layer deposition and etching (ALD/E) where we publish review tables as FAIR, machine-actionable comparisons in the Open Research Knowledge Graph (ORKG), turning them into structured, queryable knowledge. Building on this, we contrast symbolic querying over ORKG with large language model–based querying, and argue that a curated symbolic layer should remain the backbone of reliable neurosymbolic AI in materials science, with LLMs serving as complementary, symbolically grounded interfaces rather than standalone sources of truth.
\end{abstract}

\begin{quotation}
“Consider a future device \ldots\ in which an individual stores all his books, records, and communications, and which is mechanized so that it may be consulted with exceeding speed and flexibility. It is an enlarged intimate supplement to his memory.”

\hfill --- Vannevar Bush, July 1945, \textit{As We May Think}~\cite{bush1945we}
\end{quotation}

\section{\label{sec:intro}Introduction}

Vannevar Bush’s 1945 vision of a personal, searchable ``memex''\cite{bush1945we} resonates strongly with today’s situation in materials science. Atomic layer deposition and etching (ALD/ALE) now generate vast literature of processes, mechanisms, and performance data, much of it communicated through narrative text and multi-page PDF tables. Review articles help by collating this information, yet answering even a simple comparative question (e.g., ``which reported ALD processes for material~X meet a given temperature window and growth-per-cycle range?'') often still involves substantial manual effort—reading, reconstructing tables in spreadsheets, and filtering entries. As a result, the extracted synthesis is frequently captured in ad hoc formats (e.g., personal notes or local spreadsheets) that are difficult to share, query, and reuse systematically.

In parallel, there is growing interest in making scientific knowledge \emph{machine-actionable}: structured in such a way that it can be searched, filtered, and recombined automatically, in line with the FAIR Guiding Principles (Findable, Accessible, Interoperable, Reusable) for scientific data~\cite{wilkinson2016fair}. The Open Research Knowledge Graph (ORKG)\cite{auer2020improving,auer2025open} is one such next-generation infrastructure. Instead of publishing only PDFs, authors and curators can represent key findings as structured ``comparisons'' that look like familiar tables—rows for individual studies, columns for materials, precursors, substrates, process conditions, and performance metrics—but are stored as a knowledge graph with stable identifiers and an explicit schema. Earlier work has shown how such “SmartReviews’’ can make survey-article tables both human- and machine-readable\cite{oelensmartreviews}, and related efforts such as MatKG\cite{venugopal2024matkg} demonstrate that large, literature-derived knowledge graphs are feasible in materials science.

At the same time, large language models (LLMs) have quickly become attractive tools for materials researchers. Recent work in ALD benchmarking, for example, evaluates GPT-4-class models as assistants for design and conceptual questions about ALD processes\cite{yanguas2025benchmarking}, while MOF-ChemUnity couples a literature-derived knowledge base with LLM-based assistance for metal–organic framework research\cite{pruyn2025mof}. These systems are appealing because they can read PDFs, answer questions in natural language, and help explain trends or generate hypotheses. However, LLMs are also stochastic generative models: they sample from a probability distribution over possible next tokens, so that the same question can yield slightly different answers from run to run. When precise numbers and exact table entries matter—as they do for process windows, growth-per-cycle values, or etch selectivities—this lack of determinism, together with known issues such as hallucinated facts and mis-read tables, limits their suitability as standalone tools for quantitative decision-making\cite{zhao2025surveylargelanguagemodels,fan2025towards}.

In this work we bring these two developments together for the specific case of ALD/ALE review articles. Our starting point is a symbolic layer: we take nine recent review papers (three in ALD, six in ALE) and convert 18 eligible review tables—those that summarise processes and conditions across multiple primary studies—into machine-actionable ORKG comparisons. Each comparison is a structured, human-curated table with typed columns for materials, precursor chemistries, substrates, process conditions, and performance descriptors, and each row is linked back to the cited reference. On top of this symbolic representation, we then study how well current LLMs can answer precise scientific questions about these tables, both when they only see the original PDFs and when they are explicitly \emph{grounded} in the ORKG comparisons. Across these settings, we find that symbolic querying over ORKG comparisons provides the most precise and reproducible answers for review-style questions, while PDF-only LLM querying frequently struggles with table structure and numerical accuracy. Expert feedback confirms that the machine-actionable comparisons are immediately useful for comparative reasoning in ALD/ALE research. Grounding LLMs in these structured tables substantially improves answer reliability, but does not fully replace symbolic querying as a gold standard.

Concretely, we address three research questions:
\begin{enumerate}
  \item[\textbf{RQ1}] \emph{How useful do ALD/ALE experts find machine-actionable ORKG comparison tables for answering complex review questions?} We treat the ORKG tables as a new form of review output and ask practising researchers to rate their meaningfulness and usefulness, and to comment on limitations and wish-list queries.
  \item[\textbf{RQ2}] \emph{To what extent can LLMs, operating purely over review-article PDFs, reproduce the precise, table-style answers obtained from symbolic querying?} Here we upload the PDFs to state-of-the-art LLMs, pose detailed questions that mirror SPARQL queries, and measure how closely the model-generated tables match the ORKG-based gold standard.
  \item[\textbf{RQ3}] \emph{How well can LLMs answer the same questions when explicitly grounded in machine-actionable ORKG comparison tables, and how does this compare to both the symbolic SPARQL baseline and purely PDF-based querying?} In this setting, models receive the exact ORKG comparison (exported as a CSV table) together with the natural-language question, allowing us to test whether structured symbolic context closes part of the gap to symbolic querying.
\end{enumerate}

Our contributions are threefold. First, we publish 18 curated ORKG comparison tables from nine ALD/ALE review articles, together with 33 natural-language questions and their SPARQL formulations, released as the \emph{ALD/E Query Dataset}\cite{dsouza_ald-e_2025}. Second, we report a domain-expert evaluation in which three ALD/ALE researchers judge the questions and SPARQL-derived tables to be scientifically meaningful and practically useful for comparative reasoning. Third, we benchmark three question-answering regimes over the same content—(i) symbolic SPARQL over ORKG, (ii) PDF-only LLM querying, and (iii) LLM querying grounded in ORKG tables—highlighting where each succeeds and where it breaks down. Our main message for JVSTA readers is that publishing review knowledge in a symbolic, machine-actionable form alongside PDFs enables reliable expert querying today and provides a foundation for more trustworthy neurosymbolic assistants in future ALD/ALE workflows.

\begin{figure*}[!htb]
    \centering
    \includegraphics[width=\linewidth]{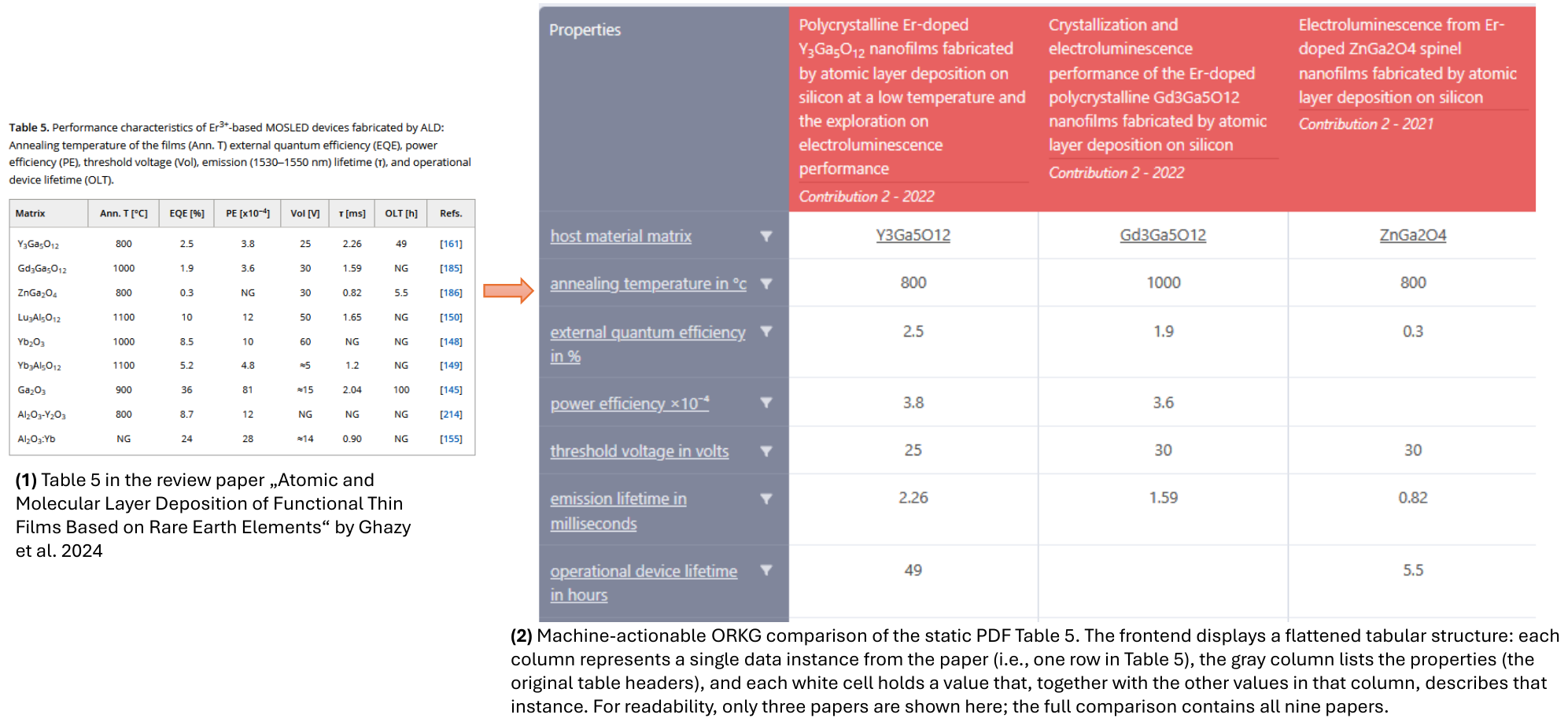}
    \caption{\label{fig:table-to-comparison} Modeling a survey table as machine-actionable data in the Open Research Knowledge Graph (ORKG). Panel~(1) shows Table~5 from the rare-earth ALD review by Ghazy et al.~\cite{ghazy2025atomic}. Panel~(2) shows the corresponding ORKG comparison \url{https://orkg.org/comparisons/R1469991}, where rows of the original table are represented as contribution columns and the original table column headers as ORKG properties. For readability, only three of the nine papers included in the full comparison are displayed. The following seven are modeled properties as web resources: \href{https://orkg.org/properties/P180031}{host material matrix}, \href{https://orkg.org/properties/P180032}{annealing temperature in $^\circ$C}, \href{https://orkg.org/properties/P180033}{external quantum efficiency in \%}, \href{https://orkg.org/properties/P180034}{power efficiency $\times 10^{-4}$}, \href{https://orkg.org/properties/P180035}{threshold voltage in volts}, \href{https://orkg.org/properties/P180036}{emission lifetime in milliseconds}, and \href{https://orkg.org/properties/P180037}{operational device lifetime in hours}.}
\end{figure*}

\section{\label{sec:fair-publishing}Background: The Open Research Knowledge Graph}

Large industrial knowledge graphs---such as those developed at Google, Microsoft, Facebook, eBay, and IBM---show how structuring information as \emph{entities} and \emph{relations} can transform access to knowledge~\cite{google-knowledgegraph-2012,noy2019industry}. In these settings, items like movies, people, or products are modeled as nodes (for example, a particular film, an actor, or a smartphone), while their properties and connections (such as ``directed by'', ``co-acted with'', ``has price'', or ``manufactured by'') are represented as labeled edges. This simple pattern---a set of comparable items, each described along a common set of properties---underlies many user-facing services: one can quickly retrieve ``all films directed by~X after~2010'' or ``all products with feature~Y below price~Z'' because the relevant facts are stored in a consistent, machine-readable structure rather than buried in free text.

A similar principle underlies the Open Research Knowledge Graph (ORKG)~\cite{auer2020improving,auer2025open}, an online platform (\url{https://orkg.org/}) that structures scientific literature in the same way: research papers, materials, processes, and experimental conditions are treated as entities, and their relationships are captured through explicitly defined properties. Instead of static PDF tables—in which column headings such as ``matrix'', ``annealing temperature'', or ``threshold voltage'' may recur across papers but remain disconnected and non-interoperable—the ORKG models these aspects as explicit, searchable properties. Relations such as ``precursor’’, ``co-reactant’’, or ``growth per cycle’’ become named predicates that can be aligned and reused across works~\cite{ehrlinger2016towards,hogan2021knowledge}.

\begin{figure*}[!htb]
    \centering
    \includegraphics[width=\linewidth]{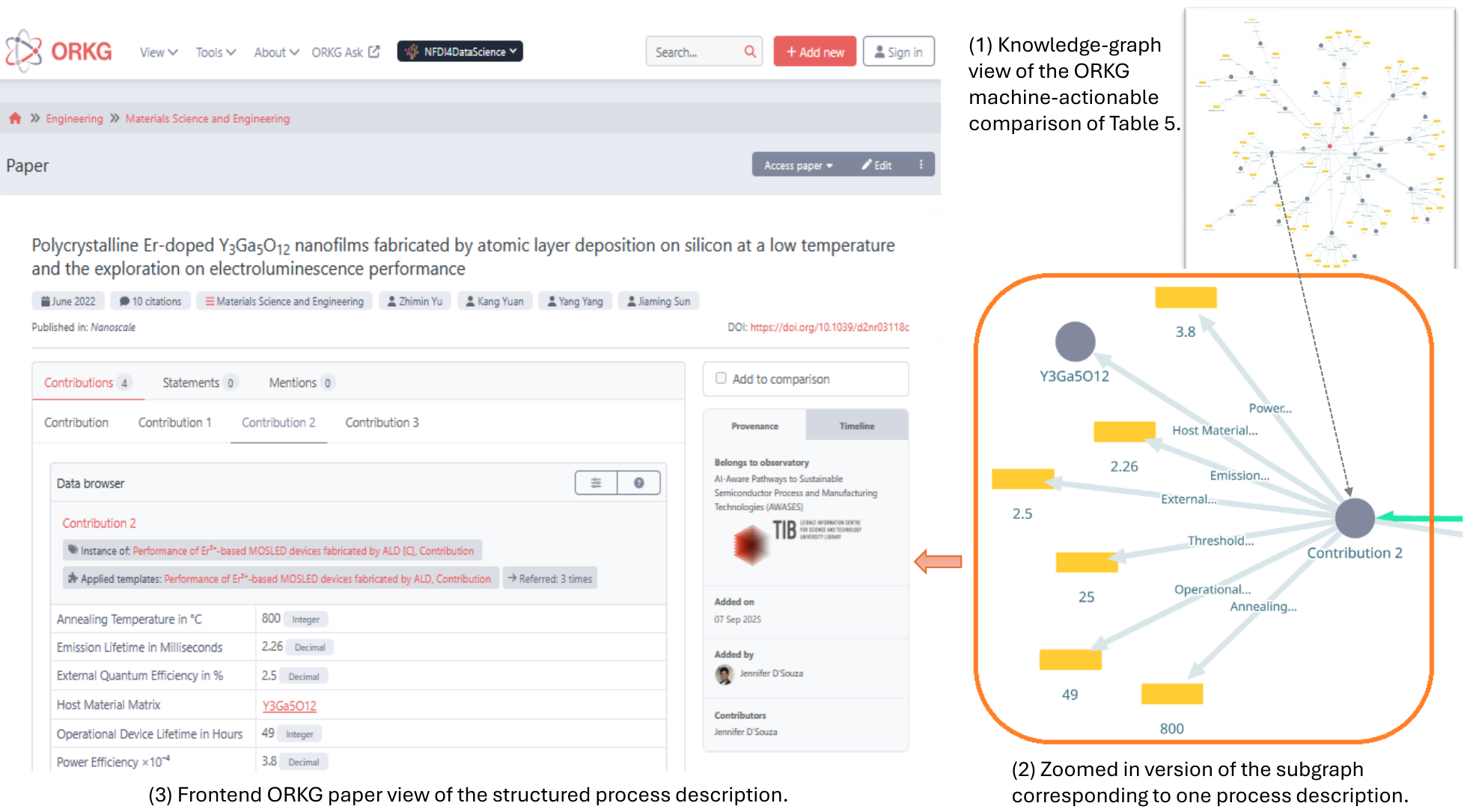}
    \caption{\label{fig:kg-to-paper} Continuation of \autoref{fig:table-to-comparison}. Starts with Panel (1) which is the knowledge graph view (the actual graph representation of the machine-actionable Table 5~\cite{ghazy2025atomic}). Panel 2 zooms in on the first contribution, linking the comparison entry to its individual paper record in the ORKG \url{https://orkg.org/papers/R1469778/R1469970} (Panel 3), which corresponds to the first data row in Table~5 in the review by Ghazy et al~\cite{ghazy2025atomic} and the article \textit{``Polycrystalline Er-doped Y\textsubscript{3}Ga\textsubscript{5}O\textsubscript{12} nanofilms fabricated by atomic layer deposition on silicon at a low temperature and the exploration on electroluminescence performance''}~\cite{yu2022polycrystalline}.}
\end{figure*}

\begin{figure*}[!htb]
    \centering
    \includegraphics[width=0.9\linewidth]{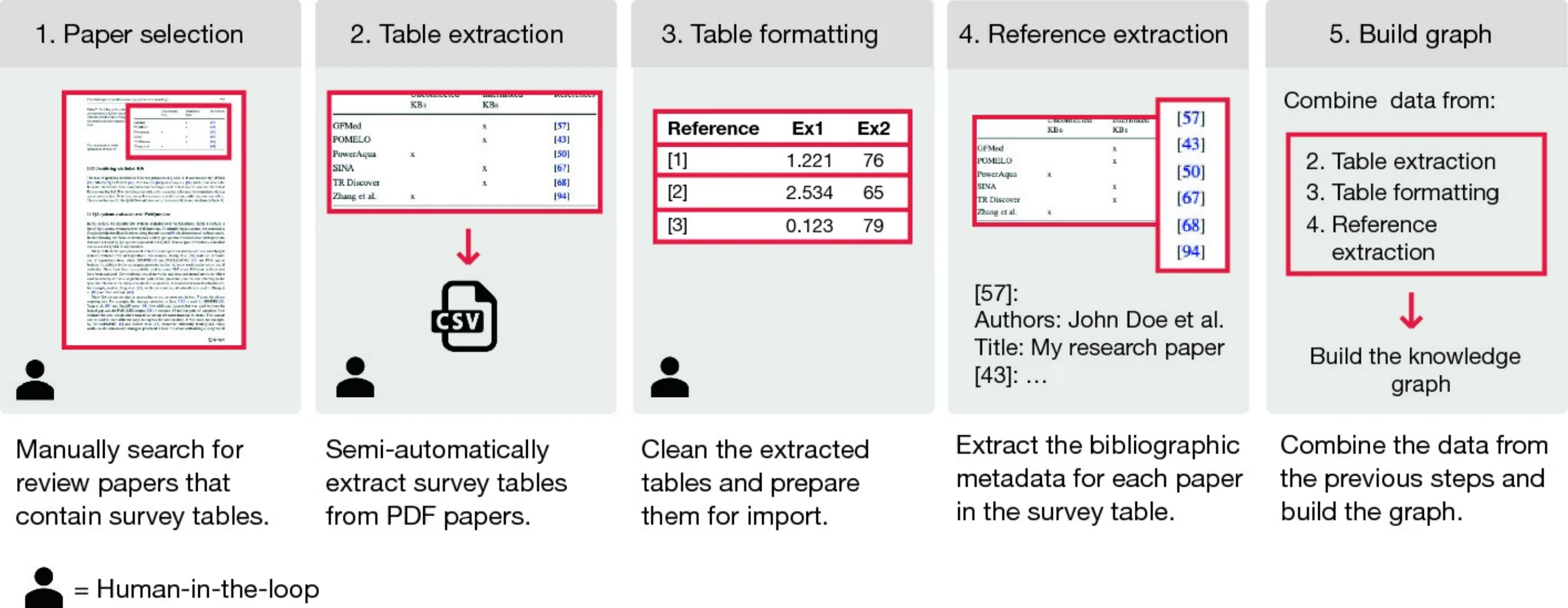}
    \caption{\label{fig:import-survey-tables}
        Methodology for importing survey tables into the Open Research Knowledge Graph (ORKG). 
        Figure reproduced from Oelen \emph{et al.} (2020), 
        ``Creating a Scholarly Knowledge Graph from Survey Article Tables'' \cite{oelen2020kgfromsurveytable}, 
        where this technique was first introduced.}
\end{figure*}

\begin{figure*}[!htb]
    \centering
    \includegraphics[width=0.75\linewidth]{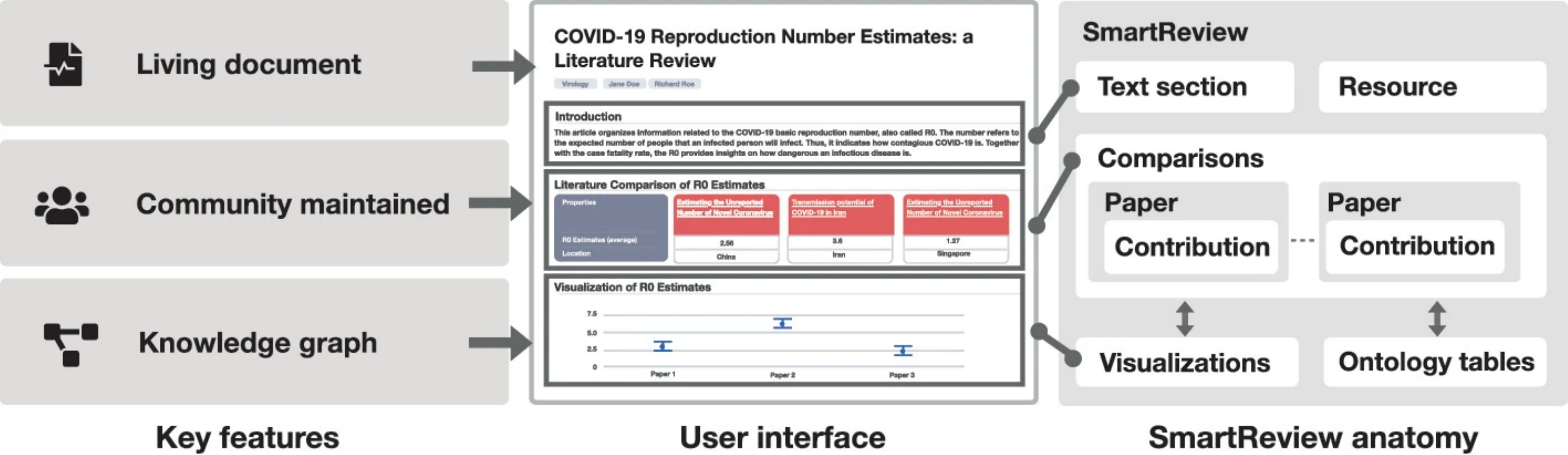}
    \caption{\label{fig:smart-review}
        Illustration of key features and anatomy of SmartReviews. 
        They are composed of several building blocks, including natural text, comparisons, and visualizations. 
        Figure reproduced from Oelen \emph{et al.} (2021), 
        ``SmartReviews: Towards Human- and Machine-Actionable Representation of Review Articles'' \cite{oelensmartreviews}, 
        where the SmartReview methodology was first introduced.}
\end{figure*}

In practice, a conventional survey or review table is decomposed so that each \emph{row} becomes a separate \emph{contribution} (one experimental configuration or reported device), and each \emph{column} becomes a well-defined property. Figure~\ref{fig:table-to-comparison} and \ref{fig:kg-to-paper} illustrates this for Table~5 in the rare-earth ALD review by Ghazy \textit{et al.} (2025)~\cite{ghazy2025atomic}. Panel~(1) reproduces the original table; Panel~(2) shows the corresponding ORKG comparison (\url{https://orkg.org/comparisons/R1469991}), where the nine table rows become nine contributions and the table headers appear as ORKG properties (gray column).

Each of these properties corresponds to a performance descriptor routinely interpreted by ALD experimentalists:
\begin{itemize}
    \item \textbf{Host material matrix} (\href{https://orkg.org/properties/P180031}{P180031}): the crystalline or amorphous material system into which Er\textsuperscript{3+} is incorporated (e.g., Y\textsubscript{3}Ga\textsubscript{5}O\textsubscript{12}, Ga\textsubscript{2}O\textsubscript{3}).
    \item \textbf{Annealing temperature} (\href{https://orkg.org/properties/P180032}{P180032}): the post-deposition thermal treatment used to activate dopants, control defects, or enhance crystallinity.
    \item \textbf{External quantum efficiency (EQE)} (\href{https://orkg.org/properties/P180033}{P180033}): the fraction of injected charge carriers converted into emitted photons.
    \item \textbf{Power efficiency (PE)} (\href{https://orkg.org/properties/P180034}{P180034}): the optical output power per unit electrical input, here reported as values scaled by $\times 10^{-4}$.
    \item \textbf{Threshold voltage} (\href{https://orkg.org/properties/P180035}{P180035}): the voltage at which electroluminescence begins.
    \item \textbf{Emission lifetime} (\href{https://orkg.org/properties/P180036}{P180036}): the decay time of the 1530--1550\,nm Er\textsuperscript{3+} emission, indicating radiative efficiency and defect environments.
    \item \textbf{Operational device lifetime (OLT)} (\href{https://orkg.org/properties/P180037}{P180037}): the duration for which the MOSLED device maintains usable output under operation.
\end{itemize}

Encoded as explicit predicates, these seven descriptors transform a single PDF table into a small, searchable performance dataset for Er\textsuperscript{3+}-based MOSLED devices. For the ALD community, the ORKG still behaves like a familiar comparison table, but—as a knowledge-graph representation—it can also be searched, filtered, linked to additional ALD/E data, and programmatically reused in ways not possible with static PDF tables.

For the ORKG platform, Oelen~\textit{et~al.}~(2020) first introduced \emph{ORKG Comparisons} as a dedicated content type for representing and comparing research contributions~\cite{oelen2020comparisons}. Conceptually, an ORKG Comparison is a structured comparison table in which:
\begin{itemize}
  \item each column is a \emph{contribution}, representing one reported device or experimental configuration from a specific research article (e.g.\ one of the nine Er\textsuperscript{3+}-based MOSLED devices listed in Table~5 of Ghazy \textit{et al.}~(2025)). Each contribution is linked directly to the ORKG entry of its source publication, ensuring full traceability so that the underlying article can be opened, inspected, and verified;  
  \item each row is a \emph{typed property}: numerical quantities such as annealing temperature, external quantum efficiency (EQE), power efficiency, threshold voltage, emission lifetime, and operational device lifetime (OLT) are stored as numbers with units, while categorical aspects such as the host material matrix are stored as controlled values rather than arbitrary free text;
  \item each cell can hold a \emph{resource-like value}, meaning that recurring entities (for example, a specific host material matrix such as Y\textsubscript{3}Ga\textsubscript{5}O\textsubscript{12}) are represented once as an identifiable resource (e.g.\ \url{https://orkg.org/resources/R1469904}) and reused across many contributions and comparisons;
  \item the comparison remains a \emph{living} object that can be updated or extended when new data or corrections become available (e.g.\ adding further Er\textsuperscript{3+}-based MOSLED devices as the literature grows);
  \item recurring column sets can be captured as \emph{templates}, making it easy to apply the same property schema (in this example: host material matrix, annealing temperature, EQE, power efficiency, threshold voltage, emission lifetime, and OLT) across multiple review tables.
\end{itemize}

In a follow-up study, Oelen~\textit{et~al.}~(2020) demonstrated how to populate such comparisons from already published survey tables~\cite{oelen2020kgfromsurveytable}. As illustrated in Figure~\ref{fig:import-survey-tables}, the workflow starts from PDF tables, extracts the cell contents, aligns the extracted columns to a simple schema (a description of what each column means), and then lets curators check and refine the result. Tables that already contain a dedicated ``Ref.'' or ``Reference'' column are particularly suitable: each row naturally corresponds to a contribution linked to a specific paper, and each column header becomes a candidate ORKG property. The outcome is \emph{FAIR} scientific data (findable, accessible, interoperable, reusable) \cite{wilkinson2016fair}: tables are no longer just images in a PDF, but machine-readable records that can be searched, filtered, and recombined in new analyses. While the survey-table setting provides a direct mapping from rows and columns to ORKG Comparisons, the same idea extends to other kinds of scientific knowledge (e.g.\ text summaries, figure panels, or protocol descriptions). In those cases, the comparison \emph{properties}—the aspects along which different studies can be compared—must first be identified or designed before a reusable comparison template can be created.

These ideas are further extended in the SmartReviews framework by Oelen~\textit{et~al.}~(2021)~\cite{oelensmartreviews}. A SmartReview is not a new type of comparison; rather, it is a way of packaging a traditional review article together with its underlying ORKG comparisons and the visualizations generated from them. In other words, a SmartReview bundles (i) ordinary prose, (ii) one or more ORKG comparisons, and (iii) figures that are driven directly by the structured data. This makes the review simultaneously readable as a normal scholarly article and usable as a small, curated data resource. For the ALD community, this simply means that tables such as the rare-earth MOSLED performance table or the SiO$_2$ ALE summaries are no longer isolated: they can be explored interactively, reused in new plots, or updated as new papers appear. 

In the remainder of this article, we adopt these ORKG concepts to encode ALD/E review tables on metal oxides and SiO$_2$ as comparisons, define reusable property templates for process parameters and mechanisms, and assemble them into SmartReview-style artefacts that support cross-table querying and long-term reuse in the ALD/E community.

\section{\label{sec:queries}Machine-actionable ALD/E Review Tables and Q\&A in the ORKG}

In this section, we demonstrate how review tables from the ALD/E literature can be modeled as machine-actionable comparisons in the Open Research Knowledge Graph (\href{https://orkg.org/}{ORKG}). Once a table is represented in this structured form, its entries can be retrieved, filtered, and combined using declarative queries rather than by manually scanning PDF documents. To illustrate this in practice, we encode seven ALD tables across three recent review papers and eleven ALE tables across six review papers as ORKG Comparisons. We then formulate a series of natural-language questions about their contents; for each question, we outline the corresponding query and show the complete or partial result. The goal is not to draw new scientific conclusions, but to demonstrate that typical ALD/E review content can be queried deterministically, ensuring factual accuracy and grounding of the retrieved measures once expressed as machine-actionable data.

Atomic layer deposition (ALD) has been extensively reviewed, from the early field-defining article by Suntola~\cite{suntola1989atomic} and subsequent foundational overviews of precursors and surface chemistry~\cite{leskela2002atomic,puurunen2005surface} through the comprehensive Kirk–Othmer Encyclopedia chapter by van Ommen, Goulas, and Puurunen~\cite{van2021atomic} and the widely cited field-wide summary by George~\cite{george2010atomic}. Process-focused surveys such as Miikkulainen~\textit{et al.}~\cite{miikkulainen2013crystallinity} and the recent updates by Popov~\textit{et al.}~\cite{popov2025recent} further map the evolution of ALD chemistries and temperature windows, while the \emph{Nature Reviews Methods Primers} article by Kessels~\textit{et al.} captures the current breadth of ALD applications~\cite{kessels2025atomic}.

To demonstrate machine-actionable modeling in a concrete ALD setting, we selected three recent review papers. The main criterion for selection was that each paper contains at least one substantive review table that compares parameters or performance measures across multiple studies and includes a dedicated ``Ref.'' or ``References'' column. This aligns naturally with the ORKG publishing model, where table rows are linked back to the original papers, as illustrated in \autoref{fig:kg-to-paper}. The decision to select the relevant papers to model was made by the first author of this work  In the following subsections, we convert these tables into ORKG Comparisons and query them using two types of example questions: \emph{Easy Queries}, which operate on a single table with simple filters, and \emph{Complex Queries}, which combine several properties, join multiple tables, or compute derived quantities.

\begin{figure*}[!htb]
\includegraphics[width=\textwidth]{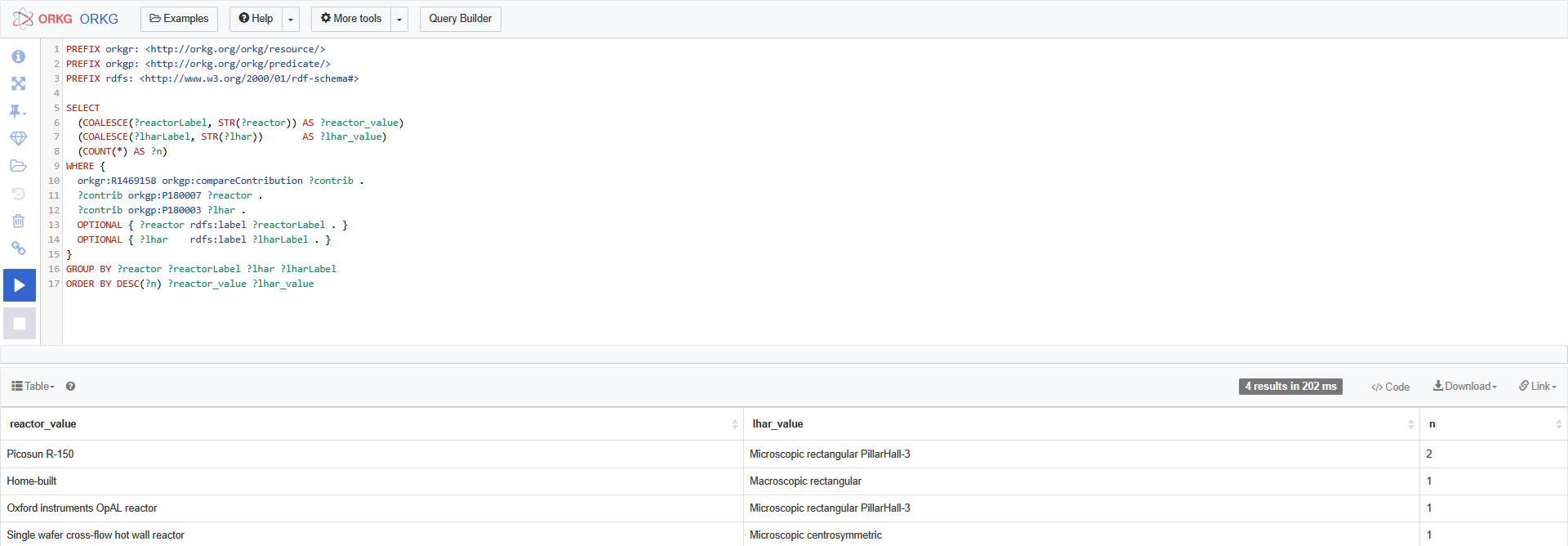}
\caption{\label{fig:reactor-lhar} Screenshot of the visual SPARQL editor provided on the ORKG platform (\url{https://orkg.org/sparql/}), showing the execution of the Q.1 for Table~2 from the research article \textit{``Saturation profile based conformality analysis for atomic layer deposition: aluminum oxide in lateral high-aspect-ratio channels''}~\cite{yim2020saturation}, modeled in the ORKG as the machine-actionable Comparison \href{https://orkg.org/comparisons/R1469158}{R1469158}. The query retrieves co-occurrences of reactor types and LHAR structures across studies, illustrating how different experimental setups pair deposition systems with test geometries. The persistent query URL can be reproduced at \url{https://tinyurl.com/lhr-reactor}.}
\end{figure*}

\subsection{\label{sec:ald-paper1} Paper — Saturation profile based conformality analysis for ALD in LHAR channels}

We begin with the paper by Yim \textit{et al.}~\cite{yim2020saturation}, whose Table~2 is modeled in the ORKG as the machine-actionable Comparison \url{https://orkg.org/comparisons/R1469158}. The table reports sticking coefficients for \ce{TMA} (\textit{c}\textsubscript{TMA}) and \ce{H2O} (\textit{c}\textsubscript{H2O}) during \ce{Al2O3} ALD in lateral high-aspect-ratio (LHAR) structures, together with process conditions such as deposition temperature, LHAR geometry, number of ALD cycles, and reactor configuration. Because the table includes a dedicated ``Reference'' column linking each row to its source article, it fits naturally into the ORKG comparison model, where every data row is traceable to an underlying publication.

In the ORKG, the Table~2 comparison \url{https://orkg.org/comparisons/R1469158} is published as a citable dataset~\cite{ald-paper1-table2} with its own DOI, \href{https://doi.org/10.48366/R1469158}{10.48366/R1469158}. As illustrated earlier in panel~2 of \autoref{fig:table-to-comparison}, the structure differs slightly from a conventional spreadsheet: each column is an ORKG \emph{contribution} (one device or experimental configuration from a specific article), and each row corresponds to a property (e.g., reactor type, LHAR geometry, \textit{c}\textsubscript{TMA}) whose values are compared across contributions. In other words, the original review table is “turned on its side,” but all the same numbers and labels are preserved and each data point remains linked to its source paper.

For the scientific community, this comparison-as-dataset publication model within the ORKG digital library offers two concrete incentives beyond a static CSV upload to a repository such as Zenodo. First, a carefully curated review table becomes a standalone research object with its own DOI that can be cited independently of the article, making the underlying quantitative synthesis more visible and reusable in later work, benchmark studies, or modeling efforts. Second, the comparison is a living resource: new findings are added as \emph{additional columns} (new contributions from subsequent studies) while the shared set of rows (properties) remains stable. Updated versions can be republished under new DOIs, and a versioning system in the ORKG backend preserves all earlier snapshots. Conceptually, this is similar to a Wikipedia-style update model—anyone can extend the comparison as new ALD results appear—but with the added benefit that each published state is frozen, timestamped, and citable. In fast-evolving areas such as LHAR conformality studies, as an example, this allows the community to maintain an up-to-date, machine-actionable comparison without losing track of the state of knowledge at earlier points in time.

\begin{table}[!htb]
\centering
\footnotesize
\caption{\label{tab:pillarhall3-results}
Answer table for Q.2 over Table~2 of Yim et al.~\cite{yim2020saturation},
modeled in the ORKG as Comparison
\href{https://orkg.org/comparisons/R1469158}{R1469158},
listing contributions reporting \textit{c}\textsubscript{TMA} values
at 300~\textdegree C in PillarHall-3 geometries.}
\resizebox{\textwidth}{!}{%
\begin{tabular}{@{}p{1.2cm} p{6.5cm} p{1cm} p{3.2cm} p{1cm}@{}}
\hline
\textbf{Contrib.} & \textbf{Research problem} & \textbf{Temp. [\textdegree C]} & \textbf{LHAR Type} & \textbf{cTMA} \\
\hline
\href{https://orkg.org/papers/R1469103/R1469105}{R1469105} &
\textit{Variation of precursor sticking coefficients in ALD on high-aspect-ratio structures} &
300 & Microscopic rectangular PillarHall-3 & 0.00572 \\
\href{https://orkg.org/papers/R1469109/R1469111}{R1469111} &
\textit{Variation of precursor sticking coefficients in ALD on high-aspect-ratio structures} &
300 & Microscopic rectangular PillarHall-3 & 0.00400 \\
\hline
\end{tabular}}
\end{table}

\paragraph{Easy Query.} The first query we introduce in this work is the following natural-language question: \emph{“Which combinations of reactor type and LHAR structure are reported across the studies, and how often does each occur?”} Once Table~2 is represented as an ORKG Comparison, a question of this kind can be answered programmatically rather than by manually scanning multiple PDF tables. We therefore implement it as a SPARQL query—the standard declarative query language for knowledge graphs, playing a role similar to SQL for relational databases—because it allows the structured table data to be searched, filtered, and aggregated in a transparent and reproducible way. In the specific query shown in \autoref{fig:reactor-lhar}, the \texttt{PREFIX} declarations at the top define shorthand names for ORKG resources (\texttt{orkgr:}), properties (\texttt{orkgp:}), and standard RDF terms (\texttt{rdfs:}); the \texttt{SELECT} clause specifies which variables appear as columns in the result table (here: reactor type, LHAR structure, and the count of their co-occurrences); and the \texttt{WHERE} block describes the pattern of links in the ORKG knowledge graph that connects each comparison contribution to its reactor and LHAR features. \texttt{OPTIONAL} clauses request human-readable labels when available, while \texttt{GROUP BY} and \texttt{ORDER BY} aggregate and sort the output. In this way, SPARQL provides a compact, readable way to formulate ``show me all reactor–geometry combinations and how often they appear'' directly over the machine-actionable representation of the review table.

A natural caveat for the reader may be: “How to obtain a SPARQL query for one's natural-language question?” There are two practical routes. First, one can write SPARQL manually by becoming familiar with these basic constructs; for readers used to SQL, the analogies above offer an accessible starting point, and the ORKG visual editor further lowers the entry barrier by exposing queries and their results side by side. Second, modern large language models (LLMs) are increasingly effective at translating natural-language questions into valid SPARQL queries when given the relevant property identifiers~\cite{xu2025harnessing,emonet2024llmbasedsparqlquerygeneration}; in particular, Avila~\textit{et~al.}~\cite{avila2024experiments} explicitly evaluate ChatGPT on text-to-SPARQL generation. In practice, a user can phrase their question in plain English, ask an LLM such as ChatGPT to propose a SPARQL query for the ORKG comparison of interest, and then run or adjust this query in the ORKG editor. This workflow substantially lowers the barrier for interacting with machine-actionable scientific data while preserving full transparency and reproducibility.

The SPARQL implementation of our first question can be rerun via \url{https://tinyurl.com/lhr-reactor}. \autoref{fig:reactor-lhar} displays this query in the \href{https://orkg.org/sparql/}{visual SPARQL editor} on the ORKG platform together with its result table, showing how all reactor–LHAR co-occurrences are extracted directly from the machine-actionable representation of Table~2.

\paragraph{Complex Query.} The second query we introduce in this work, and the first \emph{Complex Query}, is the following constrained question: \emph{“At 300~\textdegree C in PillarHall-3 structures, what \textit{c}\textsubscript{TMA} values are reported across studies?”} Throughout this paper, we refer to queries as \emph{Complex} when they go beyond simple single-property filters—by combining several properties in the same condition set, joining multiple comparisons, or computing derived quantities. In this first example, the query remains on a single comparison but simultaneously restricts three explicit properties in the ORKG knowledge graph: temperature (\href{https://orkg.org/properties/P37561}{P37561}), LHAR type (\href{https://orkg.org/properties/P180003}{P180003}), and the sticking coefficient for TMA, \textit{c}\textsubscript{TMA} (\href{https://orkg.org/properties/P180008}{P180008}); it also optionally records the associated research problem via \href{https://orkg.org/properties/P32}{P32}. We implement this question as a SPARQL query that can be rerun via \url{https://tinyurl.com/pillarhall3-ctma} over the same ORKG Comparison \url{https://orkg.org/comparisons/R1469158}. The resulting answers, summarized in \autoref{tab:pillarhall3-results}, reproduce the relevant rows from the original table in a fully machine-queryable form.

Together, these two questions show that a conventional ALD review table on LHAR conformality can be turned into a structured ORKG Comparison and queried deterministically using SPARQL, without changing the underlying experimental data. We emphasize that these queries are methodological demonstrators rather than endorsements of any specific test structure, reactor technology, or commercial implementation. As noted above, the underlying review papers were chosen by the first author via an independent literature search based on predefined inclusion criteria (presence of substantive multi-row review tables with a dedicated ``Ref.''/``References'' column and sufficient citation impact), before any of their authors were invited as co-authors on this article; the authors of the original reviews therefore did not influence the choice of example tables.

\subsection{\label{sec:ald-paper2} Paper -- Atomic layer deposition on particulate materials from 1988 through 2023: A quantitative review of technologies, materials and applications}

As our second review paper for ALD, we select the comprehensive quantitative review by Piechulla~\textit{et~al.}~(2025)~\cite{piechulla2025atomic}. This article synthesizes more than three decades of research on ALD applied to particulate materials, systematically analyzing 799 publications to map reactor technologies, coating materials, reactants, supports, and process conditions. The study’s quantitative approach aligns well with our goal of representing structured scientific knowledge in the ORKG, as it captures recurring experimental parameters and cross-references them across a large body of literature.  

From this paper, we model its Tables 3 and 4, both of which include a dedicated \textit{Reference} column linking reported data to source publications—making them particularly suitable for ORKG comparison modeling and traceable, queryable data integration.

\subsubsection{\label{sec:ald-paper2-table3} Machine-actionable modeling of Table 3}

Table~3 in the review by Piechulla~\textit{et~al.}~(2025)~\cite{piechulla2025atomic} summarizes ALD coating studies on phosphor materials. It lists, for each study, the phosphor host (e.g., BaMgAl\textsubscript{10}O\textsubscript{17}:Eu\textsuperscript{2+}, Sr[LiAl\textsubscript{3}N\textsubscript{4}]:Eu\textsuperscript{2+}, Zn\textsubscript{2}SiO\textsubscript{4}:Mn\textsuperscript{2+}), the ALD coating material (such as SiO\textsubscript{2}, Al\textsubscript{2}O\textsubscript{3}, TiO\textsubscript{2}, MgO, ZnO, TiN), the precursor scheme (e.g., TEOS/H\textsubscript{2}O, TMA/H\textsubscript{2}O, TiCl\textsubscript{4}/H\textsubscript{2}O, DEZ/H\textsubscript{2}O), coating thickness (typically 2–50~nm or given in cycles), and the corresponding reference.

We model this table as the ORKG Comparison \emph{``Comparative Studies of ALD Coatings on Phosphor Materials''} at \url{https://orkg.org/comparisons/R1469383} \cite{ald-paper2-table3}. In this machine-actionable form, each entry is linked to its source publication and all fields are represented as explicit properties, so that structured questions—such as ``Which phosphors were coated with SiO\textsubscript{2}?'' or ``Which precursor schemes were used for Al\textsubscript{2}O\textsubscript{3} coatings on Eu\textsuperscript{2+}-activated hosts?''—can be answered directly via SPARQL queries rather than by re-reading the underlying PDF table.

\begin{table}[!htb]
\centering
\caption{\label{tab:phosphor-sio2-results} Answer table for Q.3 over Table~3 of Piechulla et al.~\cite{piechulla2025atomic}, modeled in the ORKG as Comparison \href{https://orkg.org/comparisons/R1469383}{R1469383}.}
\begin{tabular}{@{}lll@{}}
\hline
\textbf{Contribution} & \textbf{Phosphor} & \textbf{Coating} \\
\hline
\href{https://orkg.org/papers/R1469160/R1469163}{R1469163} & BaMgAl\textsubscript{10}O\textsubscript{17}:Eu\textsuperscript{2+} & SiO\textsubscript{2} \\
\href{https://orkg.org/papers/R1469171/R1469173}{R1469173} & BaMgAl\textsubscript{10}O\textsubscript{17}:Eu\textsuperscript{2+} & SiO\textsubscript{2} \\
\href{https://orkg.org/papers/R1469214/R1469216}{R1469216} & Sr[LiAl\textsubscript{3}N\textsubscript{4}]:Eu\textsuperscript{2+} & SiO\textsubscript{2} \\
\hline
\end{tabular}
\end{table}

\paragraph{Easy Query.} In our question--answering dataset, the third question is: \emph{``Which phosphors were coated with SiO\textsubscript{2}?''} over the ORKG comparison of Table~3. Once the table is modeled in the ORKG, this question can be answered programmatically rather than by re-reading the original review. We implement it as a SPARQL query that can be rerun via \url{https://tinyurl.com/Phosphor-SiO2-ALD}. In contrast to traditional QA tasks that return free-text responses, our system produces a synthesized, evidence-traceable answer table: \autoref{tab:phosphor-sio2-results} lists all phosphor materials in the comparison reported with SiO\textsubscript{2} ALD coatings, each explicitly linked to its original contribution. This tabular output mitigates long, potentially redundant narratives and supports transparent verification of the underlying data.

\begin{table}[!htb]
\centering
\caption{\label{tab:red-eu2-ald} Answer table for Q.4 over Table~3 of Piechulla et al.~\cite{piechulla2025atomic}, modeled in the ORKG as Comparison \href{https://orkg.org/comparisons/R1469383}{R1469383}.}
\begin{tabular}{llllp{0.5cm} p{0.6cm}}
\hline
contrib. & phosphor & coating & precursors & temp (°C) & thick. (nm) \\
\hline
\href{https://orkg.org/papers/R1469205/R1469207}{R1469207} & Sr$_2$Si$_5$N$_8$:Eu$^{2+}$ & Al$_2$O$_3$ & TMA/H$_2$O & 30 & 8 \\
\href{https://orkg.org/papers/R1469205/R1469207}{R1469207} & Sr$_2$Si$_5$N$_8$:Eu$^{2+}$ & Al$_2$O$_3$ & TMA/O$_3$  & 30 & 8 \\
\hline
\end{tabular}
\end{table}

\paragraph{Complex Query.} The fourth question is: \emph{``Among Eu$^{2+}$-doped red phosphors, which ALD coatings deposited at $\leq 150~^{\circ}\mathrm{C}$ achieved an optimal thickness of $\leq 20$~nm, and which precursors were used?''} The SPARQL query implementation for this natural language question can be rerun via \url{https://tinyurl.com/Red-Eu2-ALD-thinlowT}. As explained earlier, answers in this work are tabulated, synthesized results based on the original question; \autoref{tab:red-eu2-ald} lists all Eu$^{2+}$-doped red phosphors, coatings, precursors, deposition temperatures, and optimal thicknesses that satisfy these query conditions.

\subsubsection{\label{sec:ald-paper2-table4} Machine-actionable modeling of Table 4}

The next ORKG comparison, available at \url{https://orkg.org/comparisons/R1469594}~\cite{ald-paper2-table4}, is modeled after Table~4 of Piechulla~\textit{et~al.}~(2025)~\cite{piechulla2025atomic}. This table compiles reported thickness ranges and growth-per-cycle (GPC) values for ALD coatings on pharmaceutically relevant particulate supports. It contains 25 experimental entries from ten publications, covering a variety of drug carriers and biomolecular substrates (e.g., lactose, budesonide, indomethacin, HPV capsomers, and proteins), together with ALD process parameters such as precursor chemistries, deposition temperatures, coating thicknesses, and GPC values. In the ORKG, this information is represented as a machine-actionable comparison that underpins the question–answering examples in this section.

\begin{table}[!htb]
\centering
\caption{\label{tab:pharma-easy-results-condensed} Answer table for Q.5 over Table~4 of Piechulla et al.~\cite{piechulla2025atomic}, modeled in the ORKG as Comparison \href{https://orkg.org/comparisons/R1469594}{R1469594}. Rows with identical support, precursor pair, and temperature are merged for compactness.}
\begin{tabular}{p{2cm} p{2cm} p{1cm} c p{2.5cm}}
\hline
support & precursors & temp (°C) & thick. (nm) & contrib. \\
\hline
budesonide & SiCl$_4$/H$_2$O & 40 & 1.5--10.0 & \href{https://orkg.org/papers/R1469549/R1469559}{R1469559}, \href{https://orkg.org/papers/R1469552/R1469560}{R1469560}, \href{https://orkg.org/papers/R1469546/R1469558}{R1469558} \\
budesonide & TMA/H$_2$O & 40 & 0.8--2.0 & \href{https://orkg.org/papers/R1469538/R1469545}{R1469545} \\
budesonide & TMA/O$_3$ & 40 & 20.0--50.0 & \href{https://orkg.org/papers/R1469549/R1469551}{R1469551}, \href{https://orkg.org/papers/R1469546/R1469548}{R1469548}, \href{https://orkg.org/papers/R1469552/R1469554}{R1469554} \\
budesonide & TiCl$_4$/H$_2$O & 40 & 3.0--15.0 & \href{https://orkg.org/papers/R1469552/R1469557}{R1469557}, \href{https://orkg.org/papers/R1469546/R1469555}{R1469555}, \href{https://orkg.org/papers/R1469549/R1469556}{R1469556} \\
lactose & TMA/H$_2$O & 30 & 2.5--11.0 & \href{https://orkg.org/papers/R1469541/R1469543}{R1469543}, \href{https://orkg.org/papers/R1469538/R1469540}{R1469540} \\
lactose & TMA/O$_3$ & 30 & 3.8--10.3 & \href{https://orkg.org/papers/R1469541/R1469544}{R1469544} \\
myoglobin & TMA/H$_2$O & 30 & 8.3 & \href{https://orkg.org/papers/R1469578/R1469580}{R1469580} \\
posaconazole/ HPMCAS & TMA/H$_2$O & 35 & 8.0--40.0 & \href{https://orkg.org/papers/R1469572/R1469589}{R1469589} \\
posaconazole/ HPMCAS & TMA/H$_2$O & 35 & 12 & \href{https://orkg.org/papers/R1469572/R1469588}{R1469588} \\
\hline
\end{tabular}
\end{table}

\paragraph{Easy Query.} The fifth question is: \emph{``Which support materials were coated at $\leq 40~^{\circ}\mathrm{C}$, and which precursor pairs were used, along with the reported coating thickness?''} The SPARQL query implementation for this question can be rerun via \url{https://tinyurl.com/less-than-40}. The tabulated answer is in \autoref{tab:pharma-easy-results-condensed}, which lists all pharmaceutically relevant supports, precursor pairs, deposition temperatures, and thickness ranges that satisfy the $\leq 40~^{\circ}\mathrm{C}$ constraint.

\begin{table}[!htb]
\centering
\caption{\label{tab:pharma-hard-results-condensed} Answer table for Q.6 over Table~4 of Piechulla et al.~\cite{piechulla2025atomic}, modeled in the ORKG as Comparison \href{https://orkg.org/comparisons/R1469594}{R1469594}. Rows with identical support–precursor–temperature combinations are merged for compactness.}
\begin{tabular}{p{1.5cm} p{1.2cm} c c c c p{2cm}}
\hline
support & precursors & temp (°C) & thick. (nm) & GPC (nm) & class & contrib. \\
\hline
lactose & TMA/H$_2$O & 30 & 2.5 & 0.47 & average & \href{https://orkg.org/papers/R1469538/R1469540}{R1469540}, \href{https://orkg.org/papers/R1469541/R1469543}{R1469543} \\
lactose & TMA/O$_3$  & 30 & 3.8 & 0.47 & average & \href{https://orkg.org/papers/R1469541/R1469544}{R1469544} \\
budesonide & TMA/H$_2$O & 40 & 0.8 & 0.20 & slow & \href{https://orkg.org/papers/R1469538/R1469545}{R1469545} \\
budesonide & TiCl$_4$/H$_2$O & 40 & 3.0 & 0.30 & n/a & \href{https://orkg.org/papers/R1469546/R1469555}{R1469555}, \href{https://orkg.org/papers/R1469549/R1469556}{R1469556}, \href{https://orkg.org/papers/R1469552/R1469557}{R1469557} \\
budesonide & SiCl$_4$/H$_2$O & 40 & 1.5 & 0.10 & n/a & \href{https://orkg.org/papers/R1469546/R1469558}{R1469558}, \href{https://orkg.org/papers/R1469549/R1469559}{R1469559}, \href{https://orkg.org/papers/R1469552/R1469560}{R1469560} \\
myoglobin & TMA/H$_2$O & 30 & 8.3 & 0.21 & slow & \href{https://orkg.org/papers/R1469578/R1469580}{R1469580} \\
posaconazole/ HPMCAS & TMA/H$_2$O & 35 & 12 & 0.40 & average & \href{https://orkg.org/papers/R1469572/R1469588}{R1469588} \\
posaconazole/ HPMCAS & TMA/H$_2$O & 35 & 8.0 & 0.40 & average & \href{https://orkg.org/papers/R1469572/R1469589}{R1469589} \\
\hline
\end{tabular}
\end{table}

\paragraph{Complex Query.} The sixth question is: \emph{``Among low-temperature runs ($< 70~^{\circ}\mathrm{C}$) that produced thin coatings ($< 20$~nm), which support materials, precursor pairs, temperatures, thicknesses, and growth-per-cycle (GPC) values were reported, and how do the GPC values classify as slow, average, or fast for alumina coatings?''} The SPARQL query implementation can be rerun via \url{https://tinyurl.com/pharma-hard-query}. The tabulated answer is in \autoref{tab:pharma-hard-results-condensed}, which lists all support materials, precursor pairs, temperatures, thicknesses, GPC values, and corresponding GPC classes satisfying the query constraints.

\subsection{\label{sec:ald-paper3} Paper -- Atomic and molecular layer deposition of functional thin films based on rare earth elements}

We now consider the review by Ghazy~\textit{et~al.}~(2025)~\cite{ghazy2025atomic}, titled ``Atomic and molecular layer deposition of functional thin films based on rare earth elements.'' In contrast to the first paper, which focused on conformality in high-aspect-ratio test structures, and the second, which provided a large-scale quantitative synthesis of particulate ALD systems, this review centers on the chemistry and functional properties of rare-earth-based ALD and MLD thin films. From this paper, we model Tables~2, 3, 4, and 5 as machine-actionable ORKG comparisons, each accompanied by Easy and Complex Queries. Together with the previous two review papers, these models cover a broad spectrum of ALD-related information and show that diverse experimental and materials data can be represented within a unified, queryable knowledge graph framework in the ORKG.

\subsubsection{\label{sec:ald-paper3-table2} Machine-actionable modeling of Table~2}

We first convert Table~2 of Ghazy~\textit{et~al.}~(2025) into an ORKG comparison, available at \url{https://orkg.org/comparisons/R1469955}~\cite{ald-paper3-table2}. This makes it citable, reusable, and versionable as a \emph{living review table} with full provenance and access to earlier versions. Table~2, titled \textit{``Use of different R species as dopants in ALD''}, lists host materials and rare-earth dopants used in ALD processes. It spans oxides (e.g., Al$_2$O$_3$, TiO$_2$, HfO$_2$), sulfides (e.g., ZnS, SrS, CaS), and garnets (e.g., Y$_3$Ga$_5$O$_{12}$, Lu$_3$Al$_5$O$_{12}$), together with reported applications such as luminescence, magnetics, gate dielectrics, electrolytes, and thermometry. Each record links a specific host–dopant combination to one or more literature references, providing the structured basis for the Easy and Complex Queries discussed next.

\begin{table}[!htb]
\centering
\caption{\label{tab:versatile-dopants} Answer table for Q.7 over Table~2 of Ghazy et al.~\cite{ghazy2025atomic}, modeled in the ORKG as Comparison \href{https://orkg.org/comparisons/R1469955}{R1469955}. For each application, the table lists rare-earth dopants and the number of distinct host materials.}
\begin{tabular}{p{1.2cm} p{4.3cm} p{2cm}}
\hline
application & dopant(s) & hosts \\
\hline
Electrolyte & Y, Dy, La & 2, 1, 1 \\
Gate dielectrics & La, Dy, Er, Gd, Pr, Y & 2, 1, 1, 1, 1, 1 \\
Interphase & Dy, La, Y & 1, 1, 1 \\
Luminescence & Er, Eu, Tb, Nd, Ce, Dy, Ho, La, Pr, Sm, Tm, Yb, (Er,Yb) & 12, 5, 5, 3, 2, 2, 2, 2, 2, 2, 2, 2, 1 \\
Magnetics & Dy, Er, Eu, Ho, La, Nd, Pr, Sm, Tb, Tm, Yb & 1 each \\
Memory & Gd, La, Y, Dy, Er, Pr & 2, 2, 2, 1, 1, 1 \\
Optoelectronics & La & 1 \\
Thermometry & Dy, Er, Eu, Ho, Nd, Sm, Tb, Tm, Yb & 1 each \\
Waveguides & (Er,Yb), Er, Yb & 1 each \\
\hline
\end{tabular}
\end{table}

\paragraph{Easy Query.} The seventh question is: \emph{``For each application (luminescence, memory, gate dielectrics, electrolytes, etc.), which rare-earth dopants appear on the largest number of distinct hosts?''} The SPARQL query implementation for this question can be rerun via \url{https://tinyurl.com/versatile-dopants} over the ORKG comparison \href{https://orkg.org/comparisons/R1469955}{R1469955} of Table~2 in the review article. The tabulated answer is in \autoref{tab:versatile-dopants}, which lists, for each application, the rare-earth dopants and their corresponding host counts.

\begin{table}[!htb]
\centering
\caption{\label{tab:cross-functional-hosts} Answer table for Q.8 over Table~2 of Ghazy et al.~\cite{ghazy2025atomic}, modeled in the ORKG as Comparison \href{https://orkg.org/comparisons/R1469955}{R1469955}. For each host material appearing in at least two application domains, the table lists its applications and rare-earth dopants.}
\begin{tabular}{p{1cm} p{3cm} p{5cm}}
\hline
host & applications & dopants \\
\hline
Al$_2$O$_3$ & Luminescence, Waveguides & (Er,Yb), Er, Yb \\
HfO$_2$ & Gate dielectrics, Memory & La, Dy, Er, Gd, Pr, Y \\
TiO$_2$ & Luminescence, Magnetics & Er, Eu, Dy, Ho, La, Nd, Pr, Sm, Tb, Tm, Yb \\
ZrO$_2$ & Electrolyte, Interphase & Dy, La, Y \\
\hline
\end{tabular}
\end{table}

\paragraph{Complex Query.} The eighth question is: \emph{``Which host materials appear in two or more different application domains, and with which dopants?''} The SPARQL query implementation for this question can be rerun via \url{https://tinyurl.com/cross-functional-hosts} over the ORKG comparison \href{https://orkg.org/comparisons/R1469955}{R1469955} of Table~2 in the review article. The tabulated answer is in \autoref{tab:cross-functional-hosts}, which lists all host materials occurring in at least two application domains, together with their applications and associated rare-earth dopants.

\subsubsection{\label{sec:ald-paper3-table3} Machine-actionable modeling of Table~3}

We modeled Table~3 of Ghazy~\textit{et~al.}~(2025) as an ORKG comparison available at \url{https://orkg.org/comparisons/R1471077}~\cite{ald-paper3-table3}. The comparison records, for each reported rare-earth (R$^{3+}$) ALD process, the deposited material, metal precursor(s), co-reactant(s), growth per cycle (GPC), deposition temperature window, and literature references, yielding a structured representation of 254 compared ALD processes. The table spans a broad range of binary and ternary systems, including rare-earth combinations with Al, Ti, Ni, Ga, W, Co, Mn, Fe, Hf, Y, Sc, Lu, and Li, many of which employ supercycle strategies to control element ratios. In the ORKG, this information becomes a machine-actionable basis for the Easy and Complex Queries on rare-earth ALD chemistry introduced next.

\begin{table}[!htb]
\centering
\caption{\label{tab:table3-highgpc-lowtemp} Answer table for Q.9 over Table~3 of Ghazy et al.~\cite{ghazy2025atomic}, modeled in the ORKG as Comparison \href{https://orkg.org/comparisons/R1471077}{R1471077}. For each ligand family, the table lists representative materials, GPC and temperature ranges, and co-reactants for high-GPC ($\geq$1~\AA{}) and low-temperature ($\leq250~^{\circ}\mathrm{C}$) rare-earth ALD processes.}
\begin{tabular}{p{0.15\columnwidth}p{0.33\columnwidth}p{0.12\columnwidth}p{0.13\columnwidth}p{0.14\columnwidth}}
\hline
\textbf{Ligand family} & \textbf{Representative materials} & \textbf{GPC [\AA]} & \textbf{Temp [$^\circ$C]} & \textbf{Co-reactant} \\
\hline
Cp-derived & \ce{Y2O3}, \ce{Er2O3}, \ce{Gd2O3}, \ce{Tm2O3}, \ce{GdAlO3} & 1.0--4.5 & 175--250 & \ce{H2O}, \ce{O3}, \ce{TMA} \\
Amidinate / Formamidinate & \ce{La2O3}, \ce{CeO2}, \ce{Pr2O3}, \ce{Gd2O3} & 1.1--2.8 & 200--240 & \ce{H2O}, \ce{O3} \\
Guanidinate & \ce{CeO2}, \ce{Y2O3}, \ce{Dy2O3}, \ce{Gd2O3} & 1.0--2.1 & 150--200 & \ce{H2O} \\
$\beta$-Diketonate & \ce{LaF3}, \ce{YF3}, \ce{HoF3} & 1.0--2.0 & 200--225 & \ce{TiF4}, \ce{NbF5} \\
\hline
\end{tabular}
\end{table}

\paragraph{Easy Query.} The ninth question is: \emph{``List all rare-earth ALD processes that achieve high growth per cycle (GPC~$\geq1$~\AA{}) at low deposition temperature ($\leq250~^{\circ}\mathrm{C}$). Report material, metal precursor family, co-reactant, GPC, and temperature; sort by GPC.''} The SPARQL query implementation for this question can be rerun via \url{https://tinyurl.com/ald-highgpc-lowtemp} over the ORKG comparison \href{https://orkg.org/comparisons/R1471077}{R1471077} of Table~3 in the review article. The tabulated answer is in \autoref{tab:table3-highgpc-lowtemp}, which groups the resulting rare-earth ALD processes by ligand family and summarizes representative materials, GPC ranges, temperature ranges, and co-reactants.

\begin{table}[!htb]
\centering
\caption{\label{tab:table3-y2o3-gpc200} Answer table for Q.10 over Table~3 of Ghazy et al.~\cite{ghazy2025atomic}, modeled in the ORKG as Comparison \href{https://orkg.org/comparisons/R1471077}{R1471077}. For \ce{Y2O3} processes in the 200--300~$^{\circ}\mathrm{C}$ window, the table reports ALD mode, average GPC, record counts, and example co-reactants.}
\begin{tabular}{p{0.23\columnwidth}p{0.22\columnwidth}p{0.18\columnwidth}p{0.3\columnwidth}}
\hline
\textbf{Mode} & \textbf{Avg. GPC [\AA]} & \textbf{Records} & \textbf{Example co-reactant} \\
\hline
Thermal ALD & 1.95 & 28 & \ce{Y(thd)3} \\
PE-ALD      & 1.65 & 3  & \ce{O2\ plasma} \\
\hline
\end{tabular}
\end{table}

\paragraph{Complex Query.} The tenth question is: \emph{``For \ce{Y2O3} only, compare the average growth per cycle (GPC) between plasma-enhanced ALD (PE-ALD) and thermal ALD within the 200--300~$^{\circ}\mathrm{C}$ window. Return mode (PE vs.\ thermal), average GPC, count of records, and a sample list of co-reactants.''} The SPARQL query implementation for this question can be rerun via \url{https://tinyurl.com/ald-y2o3-gpc200} over the ORKG comparison \href{https://orkg.org/comparisons/R1471077}{R1471077} of Table~3 in the review article. The tabulated answer is in \autoref{tab:table3-y2o3-gpc200}, which summarizes the average GPC, record counts, and example co-reactants for thermal and PE-ALD \ce{Y2O3} processes in this temperature range.

\subsubsection{\label{sec:ald-paper3-table4} Machine-actionable modeling of Table~4}

We modeled Table~4 of Ghazy~\textit{et~al.}~(2025) as an ORKG comparison available at \url{https://orkg.org/comparisons/R1470110}~\cite{ald-paper3-table4}. The comparison records, for each reported ALD/MLD process, the rare-earth–organic hybrid material, metal precursor (e.g., R(thd)$_3$, R(dpdmg)$_3$), organic linker (e.g., TPA, PDC, NDC), growth per cycle (GPC), deposition temperature, and literature reference.

The table thus covers classical R(thd)$_3$ routes with aromatic linkers as well as newer guanidinate-based chemistries reporting higher GPC. In the ORKG, this structured representation forms the basis for the Easy and Complex Queries on rare-earth–organic hybrid films introduced next.

\begin{table}[!htb]
\centering
\caption{\label{tab:table4-highgpc-lowtemp} Answer table for Q.11 over Table~4 of Ghazy et al.~\cite{ghazy2025atomic}, modeled in the ORKG as Comparison \href{https://orkg.org/comparisons/R1470110}{R1470110}. The table lists rare-earth ALD/MLD hybrid films with GPC~$\geq$5~\AA{} at deposition temperatures $\leq$250~$^{\circ}\mathrm{C}$, including material system, metal precursor, organic precursor, GPC, and temperature.}
\begin{tabular}{p{0.2\columnwidth}p{0.2\columnwidth}p{0.2\columnwidth}p{0.1\columnwidth}p{0.1\columnwidth}}
\hline
\textbf{Material system} & \textbf{Metal precursor} & \textbf{Organic precursor} & \textbf{GPC [\AA]} & \textbf{T [$^\circ$C]} \\
\hline
Ce--TPA     & Ce(DPDMG)$_3$ & TPA    & 5.4 & 200 \\
Er--3,5PDC  & Er(DPDMG)$_3$ & 3,5PDC & 6.4 & 245 \\
Eu--HQA     & Eu(thd)$_3$   & HQA    & 7.3 & 210 \\
\hline
\end{tabular}
\end{table}

\paragraph{Easy Query.} The eleventh question is: \emph{``List all rare-earth ALD/MLD hybrid films that achieve high growth per cycle (GPC~$\geq$5~\AA{}) at low deposition temperature ($\leq250~^{\circ}\mathrm{C}$). Report the material system, metal precursor family, organic precursor, GPC, and temperature; sort by GPC.''} The SPARQL query implementation for this question can be rerun via \url{https://tinyurl.com/HiGLoT-query} over the ORKG comparison \href{https://orkg.org/comparisons/R1470110}{R1470110} of Table~4 in the review article. The tabulated answer is in \autoref{tab:table4-highgpc-lowtemp}, which lists the corresponding high-GPC, low-temperature hybrid film processes and their deposition conditions.

\paragraph{Complex Query.} The twelfth question is: \emph{``Group rare-earth ALD/MLD hybrid films by organic linker family (terephthalate, pyridinedicarboxylate, naphthalenedicarboxylate, pyrazine-based, other) and compute the average growth per cycle (GPC) for each, considering only films deposited at temperatures $\leq250~^{\circ}\mathrm{C}$. Report linker family, average GPC, and number of films; sort by average GPC.''} The SPARQL query implementation for this question can be rerun via \url{https://tinyurl.com/aldmld-linker-gpc} over the ORKG comparison \href{https://orkg.org/comparisons/R1470110}{R1470110} of Table~4 in the review article.

\subsubsection{\label{sec:ald-paper3-table5} Machine-actionable modeling of Table~5}

We modeled Table~5 of Ghazy~\textit{et~al.}~(2025) as an ORKG comparison available at \url{https://orkg.org/comparisons/R1469991}~\cite{ald-paper3-table5}, instantiated with the template \url{https://orkg.org/templates/R1469957}. The comparison records, for each Er$^{3+}$-doped MOSLED device, the host matrix, annealing temperature, external quantum efficiency (EQE), power efficiency (PE), threshold voltage, emission lifetime ($\tau$), and operational device lifetime (OLT). This machine-actionable representation provides a machine-actionable basis for queries on how host composition and processing conditions relate to device performance in Er$^{3+}$-doped MOSLEDs.

\begin{table}[!htb]
\centering
\caption{\label{tab:table5-high-eqe-lowvol} Answer table for Q.13 over Table~5 of Ghazy et al.~\cite{ghazy2025atomic}, modeled in the ORKG as Comparison \href{https://orkg.org/comparisons/R1469991}{R1469991}. The table lists Er$^{3+}$-based MOSLED host matrices with EQE~$\geq$10\% and a reported threshold voltage, together with EQE, voltage, annealing temperature, emission lifetime ($\tau$), and operational lifetime (OLT).}
\begin{tabular}{p{0.15\columnwidth}p{0.1\columnwidth}p{0.1\columnwidth}p{0.12\columnwidth}p{0.12\columnwidth}p{0.1\columnwidth}}
\hline
\textbf{Host matrix} & \textbf{EQE [\%]} & \textbf{Voltage [V]} & \textbf{Anneal [$^\circ$C]} & \textbf{Lifetime $\tau$ [ms]} & \textbf{OLT [h]} \\
\hline
\ce{Al2O3\text{:}Yb}     & 24  & 14 & --   & 0.9  & --  \\
\ce{Ga2O3}        & 36  & 15 & 900  & 2.04 & 100 \\
\ce{Lu3Al5O12}    & 10  & 50 & 1100 & 1.65 & --  \\
\hline
\end{tabular}
\end{table}

\paragraph{Easy Query.} The thirteenth question is: \emph{``Among Er$^{3+}$ MOSLEDs in the comparison, which host matrices achieved high external quantum efficiency (EQE~$\geq$10\%) at the lowest threshold voltage, and what annealing temperatures and lifetimes ($\tau$, OLT) were reported?''} The SPARQL query implementation for this question can be rerun via \url{https://tinyurl.com/mosled-high-eqe} over the ORKG comparison \href{https://orkg.org/comparisons/R1469991}{R1469991} of Table~5 in the review article. The tabulated answer is in \autoref{tab:table5-high-eqe-lowvol}, which lists the corresponding host matrices together with EQE, threshold voltage, annealing temperature, emission lifetime, and operational lifetime.

\begin{table}[!htb]
\centering
\caption{\label{tab:table5-eqe-per-volt} Answer table for Q.14 over Table~5 of Ghazy et al.~\cite{ghazy2025atomic}, modeled in the ORKG as Comparison \href{https://orkg.org/comparisons/R1469991}{R1469991}. For each Er$^{3+}$-based MOSLED host matrix, the table lists EQE, operating voltage, the derived EQE/Vol metric, annealing temperature, emission lifetime ($\tau$), and operational lifetime (OLT).}
\begin{tabular}{p{0.15\columnwidth}p{0.1\columnwidth}p{0.08\columnwidth}p{0.08\columnwidth}p{0.13\columnwidth}p{0.08\columnwidth}p{0.08\columnwidth}}
\hline
\textbf{Host matrix} & \textbf{EQE [\%]} & \textbf{Vol [V]} & \textbf{EQE/Vol} & \textbf{Ann.\ T [$^\circ$C]} & \textbf{$\tau$ [ms]} & \textbf{OLT [h]} \\
\hline
\ce{Ga2O3}       & 36.0 & 15  & 2.40  & 900  & 2.04 & 100 \\
\ce{Al2O3\text{:}Yb}     & 24.0 & 14  & 1.71  & --   & 0.90 & --  \\
\ce{Yb3Al5O12}   & 5.2  & 5   & 1.04  & 1100 & 1.20 & --  \\
\ce{Lu3Al5O12}   & 10.0 & 50  & 0.20  & 1100 & 1.65 & --  \\
\ce{Yb2O3}       & 8.5  & 60  & 0.14  & 1000 & --   & --  \\
\ce{Y3Ga5O12}    & 2.5  & 25  & 0.10  & 800  & 2.26 & 49  \\
\ce{Gd3Ga5O12}   & 1.9  & 30  & 0.06  & 1000 & 1.59 & --  \\
\ce{ZnGa2O4}     & 0.3  & 30  & 0.01  & 800  & 0.82 & 5.5 \\
\hline
\end{tabular}
\end{table}

\paragraph{Complex Query.} The fourteenth question is: \emph{``Compute an efficiency-per-volt metric (EQE/Vol) to approximate the EQE–voltage trade-off, and rank host matrices accordingly.''} The SPARQL query implementation for this question can be rerun via \url{https://tinyurl.com/mosled-pareto-score} over the ORKG comparison \href{https://orkg.org/comparisons/R1469991}{R1469991} of Table~5 in the review article. The tabulated answer is in \autoref{tab:table5-eqe-per-volt}, which lists, for each host matrix, EQE, operating voltage, EQE/Vol, annealing temperature, emission lifetime, and operational lifetime.

\subsubsection{\label{sec:ald-paper3-cross-tables} Cross querying across the modeled tables}

Machine-actionable modeling of the review tables enables not only within-table queries but also cross-table queries that combine complementary information into a unified analytical view. This allows users to derive aggregate, precise insights that cannot be obtained from any table in isolation. In this section, we illustrate how Tables~2, 3, and 5 of Ghazy~\textit{et~al.}~(2025) can be jointly queried: Table~2 provides host materials that are Er-doped or luminescent, Table~3 supplies their corresponding ALD process recipes, and Table~5 reports MOSLED device metrics for Er-doped systems. Table~4, which focuses on rare-earth–organic hybrid films, has limited overlap with the MOSLED data in Table~5 and is therefore treated as optional and not included in the examples that follow.

\begin{table*}[!htb]
\centering
\caption{\label{tab:t3t5-easy-lowT} Answer table for Q.15 joining Tables~3 and~5 of Ghazy et al.~\cite{ghazy2025atomic}, modeled in the ORKG as Comparisons \href{https://orkg.org/comparisons/R1471077}{R1471077} (Table~3) and \href{https://orkg.org/comparisons/R1469991}{R1469991} (Table~5). For each Er$^{3+}$-based MOSLED host matrix, the table lists the corresponding ALD process with the lowest reported deposition temperature, including process material, metal precursor, co-reactant, precursor family, GPC, and deposition temperature, together with EQE.}
\resizebox{\textwidth}{!}{%
\begin{tabular}{p{0.16\textwidth}p{0.07\textwidth}p{0.16\textwidth}p{0.16\textwidth}p{0.12\textwidth}p{0.09\textwidth}p{0.1\textwidth}p{0.06\textwidth}}
\hline
\textbf{Host matrix} & \textbf{EQE [\%]} & \textbf{Process material} & \textbf{Metal precursor} & \textbf{Co-reactant} & \textbf{Family} & \textbf{GPC [\AA/cycle]} & \textbf{T [$^\circ$C]} \\
\hline
\ce{Gd3Ga5O12}  & 1.9 & \ce{Gd3Ga5O12}  & Gd(thd)$_3$  & TEGa   & R(thd)x & 0.19 & 350 \\
\ce{Lu3Al5O12}  & 10.0 & \ce{Lu3Al5O12} & Lu(thd)$_3$  & TMA    & R(thd)x & 0.19 & 350 \\
\ce{Y3Ga5O12}   & 2.5 & \ce{Y3Ga5O12}   & Y(thd)$_3$   & Et$_3$Ga & R(thd)x & 0.23 & 350 \\
\ce{Yb2O3}      & 8.5 & \ce{Yb2O3}      & Yb(MeCp)$_3$ & H$_2$O & other   & 1.00 & 180 \\
\ce{Yb3Al5O12}  & 5.2 & \ce{Yb3Al5O12}  & Yb(thd)$_3$  & TMA    & R(thd)x & 0.20 & 350 \\
\hline
\end{tabular}}
\end{table*}

\paragraph{Easy Query~1.} The fifteenth question is: \emph{``Show the ALD recipe alongside device performance for Er-MOSLED hosts.''} The SPARQL query implementation for this question can be rerun via \url{https://tinyurl.com/t3t5-lowT} over the ORKG comparisons \href{https://orkg.org/comparisons/R1471077}{R1471077} (Table~3) and \href{https://orkg.org/comparisons/R1469991}{R1469991} (Table~5). For each Er$^{3+}$-based MOSLED host matrix, it retrieves the highest reported EQE in Table~5 and joins it with the corresponding ALD process parameters from Table~3 (process material, metal precursor, co-reactant, precursor family, GPC, and deposition temperature). The tabulated answer is in \autoref{tab:t3t5-easy-lowT}.

\begin{table*}[!htb]
\centering
\caption{\label{tab:t2t5-easy-luminescence} Answer table for Q.16 joining Tables~2 and~5 of Ghazy et al.~\cite{ghazy2025atomic}, modeled in the ORKG as Comparisons \href{https://orkg.org/comparisons/R1469955}{R1469955} (Table~2) and \href{https://orkg.org/comparisons/R1469991}{R1469991} (Table~5). For each luminescent rare-earth–doped material appearing in both tables, the table lists EQE, threshold voltage, power efficiency (PE), annealing temperature, emission lifetime ($\tau$), and operational lifetime (OLT).}
\begin{tabular}{p{0.18\textwidth}p{0.1\textwidth}p{0.1\textwidth}p{0.1\textwidth}p{0.1\textwidth}p{0.1\textwidth}p{0.1\textwidth}}
\hline
\textbf{Material} & \textbf{EQE [\%]} & \textbf{Threshold [V]} & \textbf{PE [$\times10^{-4}$]} & 
\textbf{Ann. T [$^\circ$C]} & \textbf{$\tau$ [ms]} & \textbf{OLT [h]} \\
\hline
\ce{Ga2O3}       & 36  & $\sim$15 & 81  & 900  & 2.04 & 100 \\
\ce{Gd3Ga5O12}   & 1.9 & 30       & 3.6 & 1000 & 1.59 & --  \\
\ce{Lu3Al5O12}   & 10  & 50       & 12  & 1100 & 1.65 & --  \\
\ce{Y3Ga5O12}    & 2.5 & 25       & 3.8 & 800  & 2.26 & 49  \\
\ce{Yb2O3}       & 8.5 & 60       & 10  & 1000 & --   & --  \\
\ce{Yb3Al5O12}   & 5.2 & $\sim$5  & 4.8 & 1100 & 1.20 & --  \\
\ce{ZnGa2O4}     & 0.3 & 30       & --  & 800  & 0.82 & 5.5 \\
\hline
\end{tabular}
\end{table*}

\paragraph{Easy Query~2.} The sixteenth question is: \emph{``Which luminescent materials listed as doped systems in the ALD dopant overview also have measured MOSLED performance reported in the rare-earth MOSLED performance comparison?''} The SPARQL query implementation for this question can be rerun via \url{https://tinyurl.com/LumMOSLED-query} over the ORKG comparisons \href{https://orkg.org/comparisons/R1469955}{R1469955} (Table~2) and \href{https://orkg.org/comparisons/R1469991}{R1469991} (Table~5). For each luminescent host material that appears in both tables, it retrieves the measured performance parameters—EQE, threshold voltage, power efficiency (PE), annealing temperature, emission lifetime ($\tau$), and operational lifetime (OLT). The tabulated answer is in \autoref{tab:t2t5-easy-luminescence}.

\begin{table*}[!htb]
\centering
\caption{\label{tab:t3t5-complex-join} Answer table for Q.17 joining Tables~3 and~5 of Ghazy et al.~\cite{ghazy2025atomic}, modeled in the ORKG as Comparisons \href{https://orkg.org/comparisons/R1471077}{R1471077} (Table~3) and \href{https://orkg.org/comparisons/R1469991}{R1469991} (Table~5). For each luminescent MOSLED host matrix, the table lists EQE together with the aligned ALD process parameters: process material, metal precursor, co-reactant, GPC, and deposition temperature.}
\resizebox{\textwidth}{!}{%
\begin{tabular}{p{0.12\textwidth}p{0.14\textwidth}p{0.07\textwidth}p{0.18\textwidth}p{0.12\textwidth}p{0.12\textwidth}p{0.10\textwidth}}
\hline
\textbf{Matrix} & \textbf{Process material} & \textbf{EQE [\%]} & \textbf{Metal precursor} & \textbf{Co-reactant} & \textbf{GPC [\AA/cycle]} & \textbf{T [$^\circ$C]} \\
\hline
\ce{Lu3Al5O12} & \ce{Lu3Al5O12} & 10.0 & Lu(thd)$_3$ & TMA   & 0.19      & 350 \\
\ce{Yb2O3}     & \ce{Yb2O3}     & 8.5  & Yb(thd)$_3$ & O$_3$ & 0.15--0.20 & 250--400 \\
\ce{Yb2O3}     & \ce{Yb2O3}     & 8.5  & Yb(thd)$_3$ & O$_3$ & 0.15--0.20 & 250--400 \\
\ce{Yb2O3}     & \ce{Yb2O3}     & 8.5  & Yb(thd)$_3$ & O$_3$ & 0.15--0.20 & 250--400 \\
\ce{Yb2O3}     & \ce{Yb2O3}     & 8.5  & Yb(MeCp)$_3$ & H$_2$O & 1.00     & 180--375 \\
\ce{Yb3Al5O12} & \ce{Yb3Al5O12} & 5.2  & Yb(thd)$_3$ & TMA   & 0.20      & 350 \\
\ce{Y3Ga5O12}  & \ce{Y3Ga5O12}  & 2.5  & Y(thd)$_3$  & Et$_3$Ga & 0.23   & 350 \\
\ce{Gd3Ga5O12} & \ce{Gd3Ga5O12} & 1.9  & Gd(thd)$_3$ & TEGa  & 0.19      & 350 \\
\hline
\end{tabular}}
\end{table*}

\paragraph{Complex Query~1.} The seventeenth question is: \emph{``For each luminescent MOSLED host material listed in the rare-earth MOSLED performance comparison, retrieve the corresponding ALD process parameters from the rare-earth ALD process overview—metal precursor, co-reactant, growth-per-cycle rate, and deposition temperature—and report them alongside the external quantum efficiency (EQE).''} The SPARQL query implementation for this question can be rerun via \url{https://tinyurl.com/t3t5-complex} over the ORKG comparisons \href{https://orkg.org/comparisons/R1471077}{R1471077} (Table~3) and \href{https://orkg.org/comparisons/R1469991}{R1469991} (Table~5). The tabulated answer is in \autoref{tab:t3t5-complex-join}.

\begin{table}[!htb]
\centering
\caption{\label{tab:t3t5-correlation} Answer table for Q.18 joining Tables~3 and~5 of Ghazy et al.~\cite{ghazy2025atomic}, modeled in the ORKG as Comparisons \href{https://orkg.org/comparisons/R1471077}{R1471077} (Table~3) and \href{https://orkg.org/comparisons/R1469991}{R1469991} (Table~5). For each matrix appearing in both tables, the synthesis temperature, annealing temperature, EQE, power efficiency (PE), and emission lifetime are listed.}
\begin{tabular}{p{0.14\textwidth}p{0.18\textwidth}p{0.18\textwidth}p{0.1\textwidth}p{0.08\textwidth}p{0.16\textwidth}}
\hline
\textbf{Matrix} & \textbf{Syn. Temp [$^\circ$C]} & \textbf{Ann. Temp [$^\circ$C]} & \textbf{EQE [\%]} & \textbf{PE [\%]} & \textbf{Lifetime [ms]} \\
\hline
\ce{Ga2O3}          & 350 & 900  & 36.0 & 81.0 & 2.04 \\
\ce{Al2O3\text{:}Yb} & 350 &      & 24.0 & 28.0 & 0.90 \\
\ce{Lu3Al5O12}      & 350 & 1100 & 10.0 & 12.0 & 1.65 \\
\ce{Al2O3-Y2O3}     & 350 & 800  & 8.7  & 12.0 &      \\
\ce{Yb2O3}          & 350 & 1000 & 8.5  & 10.0 &      \\
\ce{Yb3Al5O12}      & 350 & 1100 & 5.2  & 4.8  & 1.20 \\
\ce{Y3Ga5O12}       & 350 & 800  & 2.5  & 3.8  & 2.26 \\
\ce{Gd3Ga5O12}      & 350 & 1000 & 1.9  & 3.6  & 1.59 \\
\hline
\end{tabular}
\end{table}

\paragraph{Complex Query~2.} The eighteenth question is: \emph{``Which rare-earth oxide matrices share process-level synthesis information and device-level performance data across Tables~3 and~5, and how do synthesis and annealing conditions relate to their optical efficiencies and lifetimes?''} The SPARQL query implementation for this question can be rerun via \url{https://tinyurl.com/t3t5-correlation} over the ORKG comparisons \href{https://orkg.org/comparisons/R1471077}{R1471077} (Table~3) and \href{https://orkg.org/comparisons/R1469991}{R1469991} (Table~5). For each matrix occurring in both tables, it retrieves the synthesis temperature, annealing temperature, external quantum efficiency (EQE), power efficiency (PE), and emission lifetime. The tabulated answer is in \autoref{tab:t3t5-correlation}.

\begin{table}[!htb]
\centering
\caption{\label{tab:t2t3t5-complex} Answer table for Q.19 joining Tables~2, 3, and~5 of Ghazy et al.~\cite{ghazy2025atomic}, modeled in the ORKG as Comparisons \href{https://orkg.org/comparisons/R1469955}{R1469955} (Table~2), \href{https://orkg.org/comparisons/R1471077}{R1471077} (Table~3), and \href{https://orkg.org/comparisons/R1469991}{R1469991} (Table~5). For Er-doped luminescent oxides appearing across all three tables, the synthesis temperature, EQE, and derived efficiency index (EQE/synthesis temperature~$\times$~100) are listed.}
\begin{tabular}{p{0.12\textwidth}p{0.18\textwidth}p{0.10\textwidth}p{0.18\textwidth}}
\hline
\textbf{Material} & \textbf{Syn. Temp [$^\circ$C]} & \textbf{EQE [\%]} & \textbf{Efficiency Index} \\
\hline
\ce{Ga2O3}         & 250 & 36.0 & 14.4 \\
\ce{Al2O3}         & 250 & 24.0 & 9.6 \\
\ce{Y2O3}          & 140 & 8.7  & 6.21 \\
\ce{Yb2O3}         & 180 & 8.5  & 4.72 \\
\ce{Lu3Al5O12}     & 230 & 10.0 & 4.35 \\
\ce{Yb3Al5O12}     & 350 & 5.2  & 1.49 \\
\ce{Y3Ga5O12}      & 350 & 2.5  & 0.71 \\
\ce{Gd3Ga5O12}     & 350 & 1.9  & 0.54 \\
\hline
\end{tabular}
\end{table}

\paragraph{Complex Query~3.} The nineteenth question is: \emph{``From Table~2, select oxide materials doped with Er for luminescence. Combine their synthesis temperatures (Table~3) and EQEs (Table~5), then compute an efficiency index = (EQE / synthesis temperature~$\times$~100) to identify materials that deliver high efficiency at lower fabrication temperatures. Rank materials by this index.''} The SPARQL query implementation for this question can be rerun via \url{https://tinyurl.com/t2t3t5-complex} over the ORKG comparisons \href{https://orkg.org/comparisons/R1469955}{R1469955} (Table~2), \href{https://orkg.org/comparisons/R1471077}{R1471077} (Table~3), and \href{https://orkg.org/comparisons/R1469991}{R1469991} (Table~5). The tabulated answer is in \autoref{tab:t2t3t5-complex}.

As a final note to this subsection, we publish all four machine-actionable ORKG comparisons derived from the tables of the Ghazy~\textit{et~al.}~(2025) review article \textit{``Atomic and molecular layer deposition of functional thin films based on rare earth elements''}~\cite{ghazy2025atomic}: 
Table~2 as \url{https://orkg.org/comparisons/R1469955}, 
Table~3 as \url{https://orkg.org/comparisons/R1471077}, 
Table~4 as \url{https://orkg.org/comparisons/R1470110}, and 
Table~5 as \url{https://orkg.org/comparisons/R1469991}. 
These four comparisons are jointly curated in the ORKG as a mini review article in the Smart Review format (\url{https://orkg.org/reviews/R1471098}). 
The Smart Review serves as a concise, machine-actionable counterpart to the original review, illustrating how structured, queryable representations of ALD data can complement and extend traditional scholarly publishing.

With this, we conclude the section on ALD machine-actionable comparisons in the ORKG. In total, seven tables from three recent ALD review papers were converted from their original PDF form into citable, machine-actionable ORKG Comparisons. Over these representations, we demonstrated the benefits of machine-actionable data by executing 19 SPARQL queries.

Next, we turn to Atomic Layer Etching (ALE). ALE—the subtractive analogue of ALD—uses sequential, self-limiting surface reactions to remove material with \AA-level precision. Plasma-based ALE has become well established in semiconductor processing, while thermal ALE has advanced rapidly through mechanistic studies and expanding materials coverage~\cite{kanarik2015overview,kanarik2018atomic,george2020mechanisms,fischer2021thermal}. In the remainder of this section, we apply the same ORKG-based modeling and querying approach to representative ALE review tables.

To demonstrate machine-actionable modeling for ALE, we selected six review-style articles that each contain at least one substantive table comparing process conditions, materials, or etch outcomes across multiple studies and including a dedicated ``Ref.'' or ``References'' column. Such tables align well with the ORKG publishing model, where each row is linked to its source article and each column corresponds to a well-defined property. Together, these papers cover both plasma and thermal ALE, from general overviews to materials-specific studies. In the following subsections, we convert their tables into ORKG Comparisons and illustrate Easy and Complex Queries over the reported etching parameters and outcomes.

\subsection{\label{sec:ale-paper1} Paper -- Atomic layer etching at the tipping point: an overview}

In this seminal review, Oehrlein~\textit{et~al.}~(2015)~\cite{oehrlein2015atomic} provide an early comprehensive overview of ALE, outlining key concepts, mechanisms, and materials systems, and comparing halogen- and fluorine-based chemistries across oxides, III–V semiconductors, and related substrates. From this work, we model Table~I (\textit{``Overview of materials and ALE investigations''}) as a machine-actionable ORKG Comparison, capturing the surveyed ALE investigations in structured form as a starting point for an ALE knowledge graph.

\subsubsection{\label{sec:ale-paper1-table1} Machine-actionable modeling of Table I}

Based on the ALE survey summarized in Table~I of Oehrlein~\textit{et~al.}~(2015)~\cite{oehrlein2015atomic}, we modeled the reported materials, chemistries, and energy sources as an ORKG Comparison \url{https://orkg.org/comparisons/R1562672}~\cite{ale-paper1-table1}. This machine-actionable comparison captures, for each ALE investigation, the etched material, the precursor chemistry used in the adsorption step, and the energy source driving etching or desorption, yielding a structured, queryable representation of the table. The underlying data span oxides, compound semiconductors, nitrides, polymers, and carbon-based materials, illustrating the diversity of ALE mechanisms developed to achieve atomic-scale surface removal.

\begin{table}[!htb]
\centering
\caption{\label{tab:orkg-ale-materials} Answer table for Q.20 over Table~I of Oehrlein et al.~\cite{oehrlein2015atomic}, modeled in the ORKG as Comparison \href{https://orkg.org/comparisons/R1562672}{R1562672}. The table lists, for each ALE investigation, the material, adsorption precursor chemistry, and energy source used.}
\begin{tabular}{p{0.23\columnwidth}p{0.25\columnwidth}p{0.43\columnwidth}}
\hline
\textbf{Material} & \textbf{Precursor Chemistry} & \textbf{Energy Source} \\
\hline
\ce{Al2O3}              & \ce{BCl3}                      & Ar neutral beam \\
\ce{BeO}                & \ce{BCl3}                      & Ar neutral beam \\
\ce{GaAs}               & \ce{Cl2}                       & Electron bombardment \\
\ce{GaAs}               & \ce{Cl2}                       & 248~nm KrF excimer and Ti:sapphire lasers \\
\ce{InP}                & Tertiarybutylphosphine         & Halogen lamp desorption \\
\ce{Ge}                 & \ce{Cl2}                       & Ar ions from ECR plasma \\
\ce{Graphene}           & \ce{O2} plasma                 & Ar neutral beam \\
\ce{HfO2}               & \ce{BCl3}                      & Ar neutral beam \\
Polymer (polystyrene)   & \ce{O2}                        & Ar ions from CCP plasma \\
\ce{Si}                 & \ce{Cl2}                       & Ar ions from ICP source \\
\ce{TiO2}               & \ce{BCl3}                      & Ar neutral beam \\
\hline
\end{tabular}
\end{table}

\paragraph{Easy Query.} The twentieth question is: \emph{``For each ALE investigation summarized in Table~I, list the material, adsorption precursor chemistry, and energy source used for etching or desorption.''} The SPARQL query implementation for this question can be rerun via \url{https://tinyurl.com/orkg-ale-materials-nolimit}. As explained earlier, answers in this work are tabulated, synthesized results based on the original question; \autoref{tab:orkg-ale-materials} lists all materials, precursor chemistries, and energy sources recorded in the ORKG comparison.

\paragraph{Complex Query.} The 21st question is: \emph{``Group all ALE materials by the dominant energy-source category (neutral beam, plasma ions, photon, or thermal), and count how many distinct precursor chemistries are used within each category.''} The SPARQL query implementation for this question can be rerun via \url{https://tinyurl.com/ale-energy-classes}. As with the other questions, the answer is a synthesized, machine-derived aggregation over the ORKG comparison: all ALE entries are grouped by energy-source category, and for each category the number of distinct precursor chemistries is returned. In the current comparison, plasma- or ion-driven ALE processes are associated with 12 distinct precursor chemistries, neutral-beam approaches with four, and photon-assisted and other activation schemes with one each.

\subsection{\label{sec:ale-paper2} Paper -- Thermal atomic layer etching: A review}

The review by Fischer~\textit{et al.}~(2021) provides an overview of \emph{thermal ALE}, focusing on self-limiting, plasmaless reaction mechanisms based on sequential surface fluorination and ligand-exchange steps. It surveys experimental and theoretical work across oxides, nitrides, semiconductors, and metals, and discusses underlying chemical principles, reactor designs, and mechanistic classifications. Among its contents, only Table~III met our structural and provenance criteria for machine-actionable modeling in the ORKG, as it systematically lists ALE publications by etched material and reactant chemistry. We therefore converted Table~III into a semantically structured ORKG comparison to enable reproducible, queryable synthesis of ALE chemistries across the literature.

\subsubsection{\label{sec:ale-paper2-table3} Machine-actionable modeling of Table III}

Table~III of the review article \textit{``Thermal Atomic Layer Etching: A Review''} lists ALE publications by etched material and reactant chemistry. Each row corresponds to one experimentally reported process, recording the substrate material, primary and secondary reactants, and bibliographic reference. In its machine-actionable form, this table is modeled in the ORKG as Comparison \url{https://orkg.org/comparisons/R1563034}~\cite{ale-paper2-table3}, where each publication entry is represented as a \texttt{compareContribution} with four key predicates: \href{https://orkg.org/properties/P183144}{P183144} (\textit{material~etched}), \href{https://orkg.org/properties/P183145}{P183145} (\textit{reactant~1}), \href{https://orkg.org/properties/P183146}{P183146} (\textit{reactant~2}), and \href{https://orkg.org/properties/P183147}{P183147} (\textit{reactant~3}). This encoding turns the literature summary into a semantically queryable dataset that underlies the Easy and Complex Queries discussed below.

\begin{table}[!htb]
\centering
\caption{\label{tab:thermal-ale-easy} Answer table for Q.22 over Table~III of Fischer et al.~\cite{fischer2021thermal}, modeled in the ORKG as Comparison \href{https://orkg.org/comparisons/R1563034}{R1563034}. The table shows, for each material class, representative materials, dominant reactant tuples (Reactant~1–3), and the number of reported ALE processes using those tuples.}
\begin{tabular}{p{0.45\columnwidth}p{0.35\columnwidth}p{0.11\columnwidth}}
\hline
\textbf{Material class / representative} & \textbf{Dominant reactant tuples} & \textbf{\# Reports} \\
\hline
Oxides (\ce{Al2O3}, \ce{HfO2}, \ce{ZrO2}) 
  & \ce{HF} + (\ce{TMA}, \ce{DMAC}, \ce{Sn(acac)2}, \ce{TiCl4}) 
  & 25 \\
Fluorides / Nitrides (\ce{AlF3}, \ce{AlN}) 
  & \ce{HF} + \ce{Sn(acac)2} 
  & 2 \\
III–V semiconductors (\ce{Ga2O3}, \ce{GaN}, \ce{InGaAs}) 
  & \ce{HF/BCl3} or \ce{XeF2} cycles with \ce{DMAC/TMA} 
  & 4 \\
Transition metals (\ce{Co}, \ce{Ni}, \ce{Fe}, \ce{Cu}) 
  & Oxidation/chelation: \ce{O2/O3} or \ce{Cl2} + \ce{Hhfac/Hacac} 
  & 4 \\
Si-based systems (\ce{Si}, \ce{SiO2}, \ce{Si3N4}) 
  & \ce{HF/TMA}, \ce{Cl2}, or \ce{O2/O3} plasma combinations 
  & 6 \\
Refractory / conductive films (\ce{TiN}, \ce{W}, \ce{WO3}, \ce{TiO2}) 
  & \ce{WF6}/\ce{BCl3}/\ce{O2/O3} ligand-exchange routes 
  & 5 \\
Other / emerging (\ce{VO2}, \ce{ZnO}) 
  & \ce{HF/SF4} or \ce{Hacac} + \ce{O2 plasma} 
  & 3 \\
\hline
\end{tabular}
\end{table}

\paragraph{Easy Query.} The 22nd question is: \emph{``List, for each material etched, the distinct reactant tuples reported (Reactant~1–3) and how many entries report each tuple.''} The SPARQL query implementation for this question can be rerun via \url{https://tinyurl.com/t3-easy-ale}. The tabulated, synthesized answer is shown in \autoref{tab:thermal-ale-easy}. It lists material classes with their dominant reactant tuples and the number of reported ALE processes for each. In this dataset, the query returns 46 unique material–reactant tuples.

\begin{table}[!htb]
\centering
\caption{\label{tab:thermal-ale-complex} Answer table for Q.23 over Table~III of Fischer et al.~\cite{fischer2021thermal}, modeled in the ORKG as Comparison \href{https://orkg.org/comparisons/R1563034}{R1563034}. For each mechanistic archetype, the table reports the number of distinct ALE chemistries and the number of etched materials.}
\begin{tabular}{p{0.43\columnwidth}p{0.16\columnwidth}p{0.16\columnwidth}}
\hline
\textbf{Mechanism archetype} & \textbf{\# Chemistries} & \textbf{\# Materials} \\
\hline
Fluorination + ligand-exchange & 23 & 17 \\
Other / unclassified & 11 & 10 \\
Oxidation + chelation & 3 & 4 \\
Halogenation \& conversion & 1 & 1 \\
\hline
\end{tabular}
\end{table}

\paragraph{Complex Query.} The 23rd question is: \emph{``Classify each ALE chemistry by its dominant mechanistic archetype—(i) Fluorination~+~ligand-exchange (\ce{HF} with \ce{TMA}/\ce{DMAC}/\ce{Sn(acac)2}), (ii) Oxidation~+~chelation (\ce{O2}/\ce{O3} with \ce{Hacac}/\ce{Hhfac}), (iii) Halogenation~\&~conversion (\ce{Cl2}/\ce{XeF2} with \ce{BCl3}/\ce{WF6}/\ce{WCl6}/\ce{TiCl4}), or (iv) Other.''} The SPARQL query implementation for this question can be rerun via \url{https://tinyurl.com/t3-mechanism-buckets}. As explained earlier, answers in this work are tabulated, synthesized results based on the original question; \autoref{tab:thermal-ale-complex} lists, for each mechanistic archetype, the number of distinct chemistries and etched materials in the ORKG comparison.

\subsection{\label{sec:ale-paper3} Paper -- Thermal atomic layer etching: Mechanism, materials and prospects}

Fang~\textit{et~al.}~(2018)~\cite{fang2018thermal} present a widely cited review on \emph{thermal atomic layer etching} (ALE), consolidating chemical principles, mechanistic pathways, and material systems. The article contrasts thermal ALE with ALD and plasma-based ALE, and develops mechanistic classifications such as fluorination–ligand exchange, conversion–etch, and oxidation–fluorination cycles. It also surveys materials and application challenges, including temperature-dependent EPC behavior, precursor design, and process selectivity. Within this broader context, Table~3 provides a cross-material summary of experimentally verified thermal ALE processes, which we use as the basis for a machine-actionable ORKG comparison.

\subsubsection{\label{sec:ale-paper3-table3} Machine-actionable modeling of Table 3}

Table~3 in Fang~\textit{et~al.}~(2018)~\cite{fang2018thermal} summarizes representative thermal ALE processes across multiple material classes. For each process, it reports the etched material, surface-adsorption reactant, surface-removal reactant, etch-per-cycle (EPC) rate, and process temperature. We modeled this table in the ORKG as Comparison \url{https://orkg.org/comparisons/R1560222}~\cite{ale-paper3-table3}, with each process represented as a separate \texttt{compareContribution} and the tabular fields encoded as comparison predicates. This machine-actionable representation provides the basis for the Easy and Complex Queries discussed below.

\begin{table}[!htb]
\centering
\caption{\label{tab:t3fang-easy-results} Answer table for Q.24 over Table~3 of Fang et al.~\cite{fang2018thermal}, modeled in the ORKG as Comparison \href{https://orkg.org/comparisons/R1560222}{R1560222}. The table lists all thermal ALE processes with EPC $>$ 0.5~\AA/cycle, including the material, surface-adsorption reactant, surface-removal reactant, EPC, and etching temperature.}
\begin{tabular}{p{0.1\columnwidth}p{0.2\columnwidth}p{0.22\columnwidth}p{0.15\columnwidth}p{0.12\columnwidth}}
\hline
\textbf{Material} & \textbf{Surface adsorption} & \textbf{Surface removal} & \textbf{EPC [\AA/cycle]} & \textbf{Temp [$^\circ$C]} \\
\hline
\ce{W}     & \ce{O2}           & \ce{WF6}                     & 6.30 & 300 \\
\ce{WO3}   & \ce{BCl3}         & \ce{HF}                      & 4.19 & 207 \\
\ce{W}     & \ce{O3}/\ce{BCl3} & \ce{HF}                      & 2.50 & 207 \\
\ce{ZnO}   & \ce{HF}           & TMA                          & 2.19 & 295 \\
\ce{AlN}   & \ce{HF}           & \ce{Sn(acac)2}/H$_2$ plasma  & 1.96 & 250 \\
\ce{HfO2}  & \ce{HF}           & \ce{Al(CH3)2Cl}              & 0.77 & 250 \\
\ce{Al2O3} & \ce{HF}           & \ce{Sn(acac)2}               & 0.61 & 250 \\
\ce{TiO2}  & \ce{WF6}          & \ce{BCl3}                    & 0.60 & 170 \\
\hline
\end{tabular}
\end{table}

\paragraph{Easy Query.} The 24th question is: \emph{``List all thermal ALE processes with EPC $>$ 0.5~\AA/cycle; return the material, surface-adsorption and surface-removal reactants, EPC, and etching temperature; sort by EPC.''} The SPARQL query implementation for this question can be rerun via \url{https://tinyurl.com/t3-easy-fang}. The resulting high-EPC processes from the ORKG comparison are shown in \autoref{tab:t3fang-easy-results}.

\begin{table}[!htb]
\centering
\caption{\label{tab:t3fang-complex-results} Answer table for Q.25 over Table~3 of Fang et al.~\cite{fang2018thermal}, modeled in the ORKG as Comparison \href{https://orkg.org/comparisons/R1560222}{R1560222}. For each mechanistic archetype, the table reports the number of distinct materials and the mean EPC (in~\AA/cycle).}
\begin{tabular}{p{0.34\columnwidth}p{0.15\columnwidth}p{0.23\columnwidth}}
\hline
\textbf{Mechanism archetype} & \textbf{\# Materials} & \textbf{Mean EPC [\AA/cycle]} \\
\hline
Other & 2 & 5.245 \\
Oxidation + fluorination & 2 & 1.350 \\
Fluorination + ligand-exchange & 7 & 0.617 \\
Halogenation + conversion & 1 & 0.600 \\
\hline
\end{tabular}
\end{table}

\paragraph{Complex Query.} The 25th question is: \emph{``Group all thermal ALE processes by inferred mechanism archetype—(i) Fluorination + ligand-exchange (\ce{HF} with \ce{TMA}/\ce{DMAC}/\ce{Sn(acac)2}/\ce{Al(CH3)2Cl}/\ce{SiCl4}), (ii) Halogenation + conversion (\ce{WF6} with \ce{BCl3}), (iii) Oxidation + fluorination (\ce{O2}/\ce{O3} with \ce{HF}), or (iv) Other—and compute, for each group, the number of distinct materials and the mean EPC.''} The SPARQL query implementation for this question can be rerun via \url{https://tinyurl.com/t3-complex-fang}. The aggregated results are shown in \autoref{tab:t3fang-complex-results}.

\subsection{\label{sec:ale-paper4} Paper -- Physical and chemical effects in directional atomic layer etching}

Sang and Chang (2020)~\cite{sang2020physical} review the physical and chemical factors that govern directional ALE, with emphasis on how ion–surface interactions, energy thresholds, sputtering, surface oxidation, ligand exchange, and reaction thermodynamics interact to produce anisotropy and selectivity. The article discusses plasma-based, thermal, and hybrid plasma–thermal ALE routes, linking fundamental ion-driven phenomena and surface reactions to integration requirements in advanced semiconductor patterning.

\subsubsection{\label{sec:ale-paper4-table1} Machine-actionable modeling of Table 1}

Table~1 in Sang and Chang (2020)~\cite{sang2020physical} summarizes representative ALE processes for elemental and compound semiconductors, including Si, Ge, GaAs, InP, GaN, InGaAs, and AlGaN. For each process, it records the etched material, activation mechanism (typically plasma-driven), surface-modification chemistry, removal step, and key process conditions. The table also includes two purely thermal ALE systems, \ce{GaN} (\ce{XeF2}/\ce{BCl3}) and \ce{InGaAs} (HF–pyridine/\ce{AlCl(CH3)2}). We modeled this table in the ORKG as Comparison \url{https://orkg.org/comparisons/R1560825}~\cite{ale-paper4-table1}. This machine-actionable representation underlies the Easy and Complex Queries discussed below.

\begin{table}[!htb]
\centering
\caption{\label{tab:t1sang-easy-results} Answer table for Q.26 over Table~1 of Sang and Chang~\cite{sang2020physical}, modeled in the ORKG as Comparison \href{https://orkg.org/comparisons/R1560825}{R1560825}. The table lists semiconductor ALE processes with their modification, removal, and activation modes.}
\begin{tabular}{p{0.18\columnwidth}p{0.16\columnwidth}p{0.14\columnwidth}p{0.14\columnwidth}}
\hline
\textbf{Semiconductor} & \textbf{Modification} & \textbf{Removal} & \textbf{Activation} \\
\hline
\ce{AlGaN}   & \ce{Cl2}          & \ce{Ar}             & plasma \\
\ce{GaAs}    & \ce{Cl2}          & \ce{Ar}             & plasma \\
\ce{GaN}     & \ce{Cl2}          & \ce{Ar}             & plasma \\
\ce{Ge}      & \ce{Cl2}          & \ce{Ar}             & plasma \\
\ce{InGaAs}  & \ce{Cl2}          & \ce{Ar}             & plasma \\
\ce{InP}     & \ce{Cl2}          & \ce{Ne}             & plasma \\
\ce{Si}      & \ce{Cl2}          & \ce{Ar}             & plasma \\
\ce{GaN}     & \ce{XeF2}         & \ce{BCl3}           & Thermal \\
\ce{InGaAs}  & HF-pyridine       & \ce{AlCl(CH3)2}     & Thermal \\
\hline
\end{tabular}
\end{table}

\paragraph{Easy Query.} The 26th question is: \emph{``List all semiconductor ALE processes and return, for each, the modification, removal, and activation types; sort by activation mode.''} The SPARQL query implementation for this question can be rerun via \url{https://tinyurl.com/t1-semi-ale}. The resulting semiconductor ALE processes are shown in \autoref{tab:t1sang-easy-results}.

\begin{table}[!htb]
\centering
\caption{\label{tab:t1sang-complex-results} Answer table for Q.27 over Table~1 of Sang and Chang~\cite{sang2020physical}, modeled in the ORKG as Comparison \href{https://orkg.org/comparisons/R1560825}{R1560825}. For each activation type, the table lists modification–removal pairs and the number of distinct semiconductor materials using each pair.}
\begin{tabular}{p{0.14\columnwidth}p{0.3\columnwidth}p{0.2\columnwidth}}
\hline
\textbf{Activation} & \textbf{Modification / Removal pair} & \textbf{\# Semiconductors} \\
\hline
plasma  & \ce{Cl2} / \ce{Ar}                  & 6 \\
plasma  & \ce{Cl2} / \ce{Ne}                  & 1 \\
Thermal & HF-pyridine / \ce{AlCl(CH3)2}       & 1 \\
Thermal & \ce{XeF2} / \ce{BCl3}               & 1 \\
\hline
\end{tabular}
\end{table}

\paragraph{Complex Query.} The 27th question is: \emph{``Group semiconductor ALE processes by activation type (plasma vs thermal) and count the number of distinct materials using each modification–removal pair.''} The SPARQL query implementation for this question can be rerun via \url{https://tinyurl.com/t1-sang-complex}. The aggregated results are shown in \autoref{tab:t1sang-complex-results}.



\subsection{\label{sec:ale-paper5} Paper -- Anisotropic/Isotropic Atomic Layer Etching of Metals}

Kim~\textit{et~al.}~(2020)~\cite{san2020anisotropic} review recent progress in ALE of metals, contrasting anisotropic, ion-assisted pathways with purely thermal, isotropic routes. Motivated by scaling challenges in advanced semiconductor devices and 3D architectures such as GAA and vertical FETs, the article positions ALE as a low-damage, angstrom-level alternative to conventional reactive ion etching. It surveys reported ALE chemistries for metals and metal nitrides (e.g., \ce{Cu}, \ce{W}, \ce{Co}, \ce{Cr}, \ce{Fe}, \ce{Pt}, \ce{TiN}, \ce{AlN}), discussing adsorption steps, ion-energy windows, thermal reactions, and process ideality. Within this context, Table~2 summarizes directional behavior, reactant pairings, process temperatures, and EPC values for metal ALE systems, and serves as the basis for our machine-actionable ORKG comparison.

\subsubsection{\label{sec:ale-paper5-table2} Machine-actionable modeling of Table 2}

Table~2 in Kim~\textit{et~al.}~(2020)~\cite{san2020anisotropic} summarizes reported ALE processes for metals and metal nitrides. For each entry, the table lists the etched material, directionality (anisotropic or isotropic), surface-modification and removal reactants, process temperature, and EPC. Materials covered include \ce{Co}, \ce{Fe}, \ce{Pd}, \ce{Pt}, \ce{Cr}, \ce{W}, \ce{Cu}, and nitrides such as \ce{TiN} and \ce{AlN}. We modeled this table in the ORKG as Comparison \url{https://orkg.org/comparisons/R1563131}~\cite{ale-paper5-table2}, forming a machine-actionable dataset that supports the Easy and Complex Queries presented below.

\begin{table}[!htb]
\centering
\footnotesize
\caption{\label{tab:t2metals-easy-results}
Answer table for Q.28 over Table~2 of Kim et al.~\cite{san2020anisotropic}, modeled in the ORKG as Comparison \href{https://orkg.org/comparisons/R1563131}{R1563131}. The table lists metal ALE processes with EPC~$\geq 2~\AA/\text{cycle}$, including direction, modification and removal chemistries, process temperature, and cycle time.}
\begin{tabular}{p{0.08\columnwidth}p{0.11\columnwidth}p{0.07\columnwidth}p{0.12\columnwidth}p{0.06\columnwidth}p{0.1\columnwidth}p{0.15\columnwidth}}
\hline
\textbf{Material} & \textbf{Direction} & \textbf{Reaction} & \textbf{Removal} & \textbf{EPC} & \textbf{Temp ($^\circ$C)} & \textbf{Cycle time (sec)} \\
\hline
Fe & Anisotropy & O\textsubscript{2} & Formic acid & 42   & 80  & 350 \\
Co & Anisotropy & O\textsubscript{2} & Formic acid & 28   & 80  & 350 \\
Pd & Anisotropy & O\textsubscript{2} & Formic acid & 12   & 80  & 350 \\
W  & Isotropy   & O\textsubscript{2} & WF\textsubscript{6} & 6.3  & 300 & 320 \\
Pt & Anisotropy & O\textsubscript{2} & Formic acid & 5    & 80  & 350 \\
W  & Anisotropy & NF\textsubscript{3} & O\textsubscript{2} & 2.6  &      & 40 \\
W  & Isotropy   & O\textsubscript{2}/O\textsubscript{3} & BCl\textsubscript{3}$\rightarrow$HF & 2.5 & 207 & 325 \\
W  & Anisotropy & Cl\textsubscript{2} & Ar & 2.1 & 60 & 25 \\
Co & Isotropy   & Cl\textsubscript{2} & hfacH & 2 & 140 & 95 \\
\hline
\end{tabular}
\end{table}

\paragraph{Easy Query.} The 28th question is: \emph{``List all metal ALE processes with EPC~$\geq 2~\AA/\text{cycle}$; return the material, direction (anisotropic vs.\ isotropic), modification and removal chemistries, EPC, process temperature, and cycle time; sort by EPC in descending order.''} The SPARQL query implementation for this question can be rerun via \url{https://tinyurl.com/t2-metals-high-epc}. The resulting high-EPC metal ALE processes are shown in \autoref{tab:t2metals-easy-results}.

\begin{table}[!htb]
\centering
\caption{\label{tab:t2metals-complex-results}
Answer table for Q.29 over Table~2 of Kim et al.~\cite{san2020anisotropic}, modeled in the ORKG as Comparison \href{https://orkg.org/comparisons/R1563131}{R1563131}. For each etching direction, the table reports the number of distinct metals and the mean EPC (in~\AA/cycle).}
\begin{tabular}{p{0.15\columnwidth}p{0.2\columnwidth}p{0.25\columnwidth}}
\hline
\textbf{Direction} & \textbf{Number of Metals} & \textbf{Mean EPC (Å/cycle)} \\
\hline
Anisotropy & 7 & 11.6875 \\
Isotropy   & 5 & 2.035   \\
\hline
\end{tabular}
\end{table}

\paragraph{Complex Query.} The 29th question is: \emph{``Group metal ALE processes by direction (anisotropic vs.\ isotropic) and compute, for each group, the number of distinct metals covered and the mean EPC; sort by mean EPC in descending order.''} The SPARQL query implementation for this question can be rerun via \url{https://tinyurl.com/t2-metals-complex}. The aggregated results are shown in \autoref{tab:t2metals-complex-results}.

\subsection{\label{sec:ale-paper6} Paper -- Atomic layer etching of SiO$_2$ for nanoscale semiconductor devices: A review}

Hong~\textit{et~al.}~(2024)~\cite{hong2024atomic} review ALE strategies for SiO$_2$ in advanced semiconductor patterning. They contrast fluorocarbon-based anisotropic ALE—controlled by precursor choice, ion energy, etch selectivity, and chamber-wall effects—with isotropic ALE routes that convert SiO$_2$ into volatile intermediates such as Al$_2$O$_3$ or ammonium fluorosilicate under thermal or plasma activation. By linking mechanistic steps to device-relevant process metrics, the review discusses how SiO$_2$ ALE can be tuned for emerging 3D architectures and sub-10~nm fabrication requirements.

\subsubsection{\label{sec:ale-paper6-tables} Machine-actionable modeling of Tables I--VI}

Tables~I--VI in the SiO$_2$ ALE review together cover fluorocarbon-based anisotropic ALE and thermal or plasma-assisted isotropic mechanisms. Tables~I and~II (\url{https://orkg.org/comparisons/R1560949}~\cite{ale-paper6-table1}, \url{https://orkg.org/comparisons/R1560977}~\cite{ale-paper6-table2}) list fluorination precursors, removal gases, etch rates, temperatures, and ion-energy windows for anisotropic SiO$_2$ ALE. Tables~III and~IV (\url{https://orkg.org/comparisons/R1561025}~\cite{ale-paper6-table3}, \url{https://orkg.org/comparisons/R1561023}~\cite{ale-paper6-table4}) summarize selectivity (SiO$_2$/Si and SiO$_2$/Si$_3$N$_4$), methods for improving selectivity, and chamber-wall conditioning effects. Tables~V and~VI (\url{https://orkg.org/comparisons/R1561046}~\cite{ale-paper6-table5}, \url{https://orkg.org/comparisons/R1562478}~\cite{ale-paper6-table6}) cover thermal and plasma-assisted isotropic ALE, organizing multi-step reaction pathways together with etch rates.

As in the other ALE tables modeled in this work, all six SiO$_2$ tables are represented in the ORKG as machine-actionable comparisons using a shared, harmonized predicate schema. For Tables~I--IV, core descriptors include \href{https://orkg.org/properties/P183129}{P183129} (fluorination precursor chemistry), \href{https://orkg.org/properties/P183125}{P183125} (removal chemistry), \href{https://orkg.org/properties/P183130}{P183130} (process temperature), and \href{https://orkg.org/properties/P183131}{P183131} (etch rate), with Table~II adding the plasma-relevant \href{https://orkg.org/properties/P183132}{P183132} (ion energy in the removal step). Table~III introduces selectivity-focused predicates \href{https://orkg.org/properties/P183133}{P183133} (selectivity material pair), \href{https://orkg.org/properties/P173032}{P173032} (selectivity value), and \href{https://orkg.org/properties/P183134}{P183134} (selectivity improvement method), while Table~IV contributes \href{https://orkg.org/properties/P183135}{P183135} (method of chamber-wall effect). The isotropic ALE mechanism tables (V and VI) use \href{https://orkg.org/properties/P183136}{P183136} (etching mechanism class) together with step-level predicates \href{https://orkg.org/properties/P183137}{P183137}, \href{https://orkg.org/properties/P183138}{P183138}, and \href{https://orkg.org/properties/P183139}{P183139}. Within this shared schema, cross-table SPARQL queries can link precursor families, ion-energy windows, selectivity ranges, chamber-wall treatments, and isotropic mechanism pathways across the entire ALE corpus.

The six SiO$_2$ ALE comparisons are also curated in an ORKG Smart Review (\url{https://orkg.org/reviews/R1562498}), which provides a single entry point to the modeled tables from Hong~\textit{et~al.}~(2024)~\cite{hong2024atomic}. While cross-table SPARQL queries can be executed directly on the comparisons, the Smart Review documents the modeling choices, highlights example queries, and connects the semantic data model back to the original article.

\begin{table}[!htb]
\centering
\caption{\label{tab:crossq1-easy-results} Answer table for Q.30 (Easy Cross-Query~1) over Tables~I and~II of Hong et al.~\cite{hong2024atomic}, modeled in the ORKG as Comparisons \href{https://orkg.org/comparisons/R1560949}{R1560949} and \href{https://orkg.org/comparisons/R1560977}{R1560977}. The table lists fluorocarbon ALE processes that appear in both tables under near-room-temperature (RT~$\pm 20^\circ$C) conditions, together with removal chemistry, etching rate, and ion-energy window.}
\begin{tabular}{p{0.22\columnwidth}p{0.12\columnwidth}p{0.12\columnwidth}p{0.15\columnwidth}p{0.21\columnwidth}}
\hline
\textbf{Precursor} & \textbf{Removal} & \textbf{Temp ($^\circ$C)} & \textbf{EPC (\AA/cycle)} & \textbf{Ion-energy window} \\
\hline
C4F6/Ar plasma                & Ar plasma     & $-10$ & 14.2 & 10--100~V \\
CHF$_3$ plasma               & Ar plasma     & $-40$--20 & 9.0 & 0--50~V \\
C4F8 plasma                  & Ar plasma     & 20  & 5.8 & 30--90~V \\
CF$_3$CF$_2$CF$_2$CH$_2$OH plasma & Ar plasma & RT  & 5.2 & 10--80~V \\
CHF$_3$ plasma               & O$_2$ plasma  & 20  & 4.1 & 30--90~V \\
n-C$_3$F$_7$OCH$_3$ plasma   & Ar plasma     & RT  & 2.1 & 10--80~V \\
C4F8 plasma                  & Ar ion beam   & RT  & 1.9 & 30--200~V \\
i-C$_3$F$_7$OCH$_3$ plasma   & Ar plasma     & RT  & 1.8 & 10--80~V \\
\hline
\end{tabular}
\end{table}

\paragraph{Easy Cross-Query 1.} The 30th question is: \emph{``For each fluorocarbon precursor system that appears in both Table~I and Table~II, list the precursor chemistry, removal gas, process temperature, etching rate, and ion-energy window used in the removal step; restrict to conditions near room temperature (RT~$\pm 20^\circ$C) and sort by etching rate in descending order.''} The SPARQL query implementation for this question can be rerun via \url{https://tinyurl.com/pap6-crossq1-easy}. The resulting fluorocarbon ALE processes are shown in \autoref{tab:crossq1-easy-results}.

\begin{table}[!htb]
\centering
\caption{\label{tab:crossq2-easy-results} Answer table for Q.31 (Easy Cross-Query~2) over Tables~III and~IV of Hong et al.~\cite{hong2024atomic}, modeled in the ORKG as Comparisons \href{https://orkg.org/comparisons/R1561025}{R1561025} and \href{https://orkg.org/comparisons/R1561023}{R1561023}. The table lists C$_4$F$_8$/Ar-based anisotropic SiO$_2$ ALE processes that appear in both tables, including their selectivity pair, selectivity range, selectivity method, chamber-wall treatment, and etch rate.}
\begin{tabular}{p{0.11\columnwidth}p{0.12\columnwidth}p{0.26\columnwidth}p{0.2\columnwidth}p{0.15\columnwidth}}
\hline
\textbf{Selectivity pair} & \textbf{Selectivity} & \textbf{Selectivity method} & \textbf{Chamber-wall treatment} & \textbf{EPC (\AA/cycle)} \\
\hline
SiO$_2$/Si$_3$N$_4$ & 0.2--15.0 & Precursor selection; Fluorocarbon film thickness; Ion energy; Etching step time & Chamber cleaning with O$_2$ plasma; Chamber wall heating & 3 \\
SiO$_2$/Si$_3$N$_4$ & $>7.0$    & Precursor selection & Chamber cleaning with O$_2$ plasma; Chamber wall heating & 3 \\
SiO$_2$/Si          & $>10.0$   & Precursor selection & Chamber cleaning with O$_2$ plasma; Chamber wall heating & 3 \\
\hline
\end{tabular}
\end{table}

\paragraph{Easy Cross-Query 2.} The 31st question is: \emph{``Collect all anisotropic SiO$_2$ ALE processes based on C$_4$F$_8$/Ar plasma from Table~III and Table~IV, and list for each the target selectivity pair (SiO$_2$/Si or SiO$_2$/Si$_3$N$_4$), selectivity range (if available), chamber-wall treatment (none, O$_2$ cleaning, wall heating), and etching rate; sort by etching rate.''} The SPARQL query implementation for this question can be rerun via \url{https://tinyurl.com/pap6-crossq2-easy}. The matched processes are shown in \autoref{tab:crossq2-easy-results}.

\begin{table}[!htb]
\centering
\caption{\label{tab:crossq1-complex-results} Answer table for Q.32 (Complex Cross-Query~1) over Tables~I--III of Hong et al.~\cite{hong2024atomic}, modeled in the ORKG as Comparisons \href{https://orkg.org/comparisons/R1560949}{R1560949}, \href{https://orkg.org/comparisons/R1560977}{R1560977}, and \href{https://orkg.org/comparisons/R1561025}{R1561025}. For each fluorocarbon precursor family, the table reports mean and maximum SiO$_2$ etch rate, the combined ion-energy windows, and the maximum reported SiO$_2$/Si and SiO$_2$/Si$_3$N$_4$ selectivity.}
\begin{tabular}{p{0.12\columnwidth}p{0.12\columnwidth}p{0.12\columnwidth}p{0.18\columnwidth}p{0.1\columnwidth}p{0.1\columnwidth}}
\hline
\textbf{Precursor family} & \textbf{Mean etch rate} & \textbf{Max.\ etch rate} & \textbf{Ion-energy windows} & \textbf{Max.\ Sel.\ SiO$_2$/Si} & \textbf{Max.\ Sel.\ SiO$_2$/Si$_3$N$_4$} \\
\hline
C$_4$F$_8$   & \textemdash & \textemdash & \textemdash      & 0 & 0 \\
C$_3$H$_3$F$_3$ & \textemdash & \textemdash & \textemdash      & 0 & 0 \\
C$_4$F$_6$   & \textemdash & \textemdash & 10--100~V        & \textemdash & \textemdash \\
CF$_3$I      & \textemdash & \textemdash & \textemdash      & \textemdash & \textemdash \\
CHF$_3$      & \textemdash & \textemdash & \textemdash      & 0 & 0 \\
\hline
\end{tabular}
\end{table}

\paragraph{Complex Cross-Query 1.} The 32nd question is: \emph{``Group all anisotropic SiO$_2$ ALE processes by fluorocarbon precursor family (e.g., C$_4$F$_8$-based, CHF$_3$-based, C$_3$F$_7$OCH$_3$ isomers) across Tables~I, II, and III. For each family, compute (i) the mean and maximum SiO$_2$ etching rate, (ii) the union of reported ALE ion-energy windows in the removal step, and (iii) the maximum reported SiO$_2$/Si and SiO$_2$/Si$_3$N$_4$ selectivity; return one row per precursor family, ordered first by maximum selectivity, then by mean etching rate.''} The SPARQL query implementation for this question can be rerun via \url{https://tinyurl.com/pap6-crossq1-complex}. The resulting precursor-family aggregation is shown in \autoref{tab:crossq1-complex-results}, and reflects the limited amount of parseable numeric etch-rate and selectivity data currently available in the modeled tables.

\begin{table*}[!htb]
\centering
\caption{\label{tab:crossq2-complex-results} Answer table for Q.33 (Complex Cross-Query~2) over Tables~V and~VI of Hong et al.~\cite{hong2024atomic}, modeled in the ORKG as Comparisons \href{https://orkg.org/comparisons/R1561046}{R1561046} and \href{https://orkg.org/comparisons/R1562478}{R1562478}. Isotropic SiO$_2$ ALE processes are grouped by etching mechanism class; for each class, the table reports the number of variants, minimum/maximum/mean etching rate, temperature range associated with rate-limiting steps, and whether plasma is required and high-rate operation at low temperature is achieved.}
\resizebox{\textwidth}{!}{%
\begin{tabular}{lccccccc}
\hline
\textbf{Mechanism class} & \textbf{\# variants} & \textbf{Min.\ rate} & \textbf{Max.\ rate} & \textbf{Mean rate} & \textbf{Temp range [$^\circ$C]} & \textbf{Plasma required} & \textbf{High-rate, low-$T$} \\
\hline
AFS-based plasma-assisted   & 3 & 27   & 75   & 43   & 3--4     & yes & yes \\
AFS-based thermal           & 1 & 9    & 9    & 9    & 20--20   & no  & yes \\
Al$_2$O$_3$-conversion cycles & 2 & 0.07 & 0.35 & 0.21 & 300--350 & no  & no  \\
\hline
\end{tabular}}
\end{table*}

\paragraph{Complex Cross-Query 2 (Tables V + VI: thermal vs.\ plasma-assisted isotropic ALE).} The 33rd question is: \emph{``Combine all isotropic SiO$_2$ ALE processes and group by mechanism class; report number of variants, min/max/mean etching rate, temperature range, and plasma requirement.''} The SPARQL query implementation for this question can be rerun via \url{https://tinyurl.com/pap6-crossq2-complex-fixed}. The aggregated mechanism-class statistics are shown in \autoref{tab:crossq2-complex-results}.

With this, we conclude the section on ALE machine-actionable comparisons in the ORKG.

Across all ALD and ALE review papers considered in this work, we modeled a total of \textbf{18 machine-actionable tables} as ORKG Comparisons: \textbf{seven} from three ALD papers and \textbf{eleven} from six ALE papers. These structured representations provide a unified semantic layer for precursor and reactant chemistries, process parameters, etch and growth metrics, selectivity behavior, ion-energy windows, chamber-wall effects, and isotropic multi-step mechanisms. In total, \textbf{33 natural-language questions}---\textbf{19} for ALD and \textbf{14} for ALE---were implemented as SPARQL queries, each producing a reproducible, machine-derived result table. Together, these queries form a comprehensive benchmark for neurosymbolic querying over materials-science knowledge in the ORKG.

As supplementary data to this paper, a consolidated catalog of papers, tables, comparisons, NL questions, and SPARQL links is available in our GitHub repository: \url{https://github.com/sciknoworg/ald-ale-orkg-review}.

In the remainder of this paper, we evaluate several LLM-based question answering settings on this dataset, using the SPARQL query results as gold-standard answers.

\section{\label{sec:eval}Evaluations and Results}

With all published machine-actionable, symbolic representations of selected parts of review articles in materials science in place, we next perform quantitative evaluations to answer the three research questions introduced at the beginning of the article: (i) how useful domain experts find these ORKG-based comparison representations (\textbf{RQ1}), (ii) how purely neural, PDF-based LLM querying compares to the symbolic SPARQL baseline (\textbf{RQ2}), and (iii) how neurosymbolic querying over ORKG comparisons performs relative to both (\textbf{RQ3}). Together, these evaluations make the case for the importance of \emph{symbolic knowledge representations of scientific knowledge to facilitate neurosymbolic AI}—the central research premise of this paper.

\begin{figure*}[!htb]
    \centering
    \includegraphics[width=\linewidth]{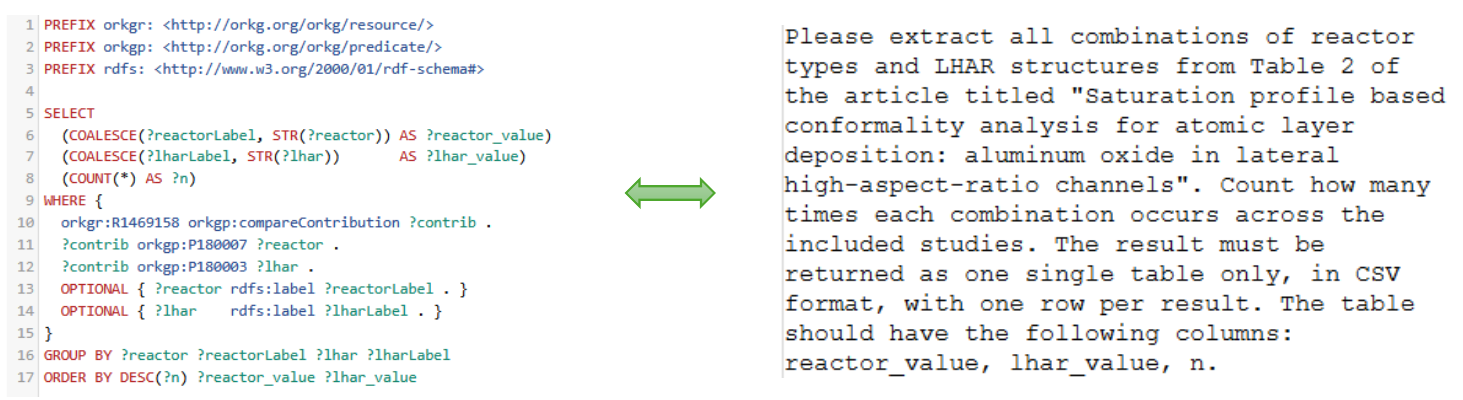}
    \caption{\label{fig:sparql-to-NL}Illustration of a SPARQL query operating over the machine-actionable ORKG comparison graph \url{https://orkg.org/comparisons/R1469158} and its natural-language equivalent. The latter can be used in LLM-based querying scenarios over different forms of input, such as the article PDF or a serialized version of the comparison. This SPARQL query was introduced in \autoref{sec:ald-paper1}.}
\end{figure*}

\subsection{Evaluation Setup}

We begin by describing our evaluation setup: \autoref{sec:exp-dataset} introduces the experimental dataset, and \autoref{sec:exp-setting} outlines the experimental settings.

\subsubsection{\label{sec:exp-dataset} Experimental dataset}

Our evaluations are based on the 33 SPARQL queries introduced in Sec.~\ref{sec:queries}: 19 questions over the ALD comparisons and 14 over the ALE comparisons. Each query runs on one or more ORKG comparisons and returns a machine-generated answer table, which we use as the gold standard.

For the neural (LLM-based) experiments, we then created a natural-language version of each SPARQL query. These natural-language queries explicitly specify (i) which review table in the paper to consult, (ii) which columns in that table to use, and (iii) which columns the answer table should contain and that the output must be formatted as a table. In other words, they convey the same information as the corresponding SPARQL query, but in a form that an LLM can directly follow. An example is shown in \autoref{fig:sparql-to-NL}. We generated these natural-language queries programmatically using \href{https://huggingface.co/mistralai/Mistral-Large-Instruct-2407}{\texttt{mistral-large}}; the Python script is available in our GitHub repository:
\url{https://github.com/sciknoworg/ald-ale-orkg-review/blob/main/llm-experiments/nl_query_generator_csv_onetable.py}.
This yields 33 detailed natural-language queries that map one-to-one to the 33 SPARQL queries and their gold-standard answer tables. Note that these differ from the earlier (\autoref{sec:queries}), more informal question formulations, which are intended for readers rather than for enforcing a consistent output structure from LLMs. For quick reference, the GitHub repository also lists, for each ALD and ALE review table (including cross-table queries), both the short and detailed natural-language queries and the corresponding \href{https://github.com/sciknoworg/ald-ale-orkg-review/}{SPARQL query links}.

For one of the experimental settings (Setting~3, described below), the LLM does not see the full review PDF. (In Setting~2, by contrast, the model is given the review PDF or its extracted text and must read it to answer the question.) In Setting~3, to bring the neural experiments closer in spirit to the pure SPARQL (symbolic) gold standard of Setting~1, we instead provide the ORKG comparison table itself in a simple, spreadsheet-like format (a CSV file). Each row in this CSV corresponds to one row of the original review table, and each column corresponds to one of the typed properties we modeled in ORKG (for example, material, precursor chemistry, substrate, temperature, etch rate, or selectivity). In other words, the model sees the same structured information that the SPARQL queries in Setting~1 operate on, but now as a flat text table rather than as a figure embedded in a PDF. We obtain these CSV files via the built-in ORKG export function, which also supports additional formats such as \LaTeX, PDF, RDF, and citation metadata. Settings~2 and~3 therefore both evaluate question answering via large language models, in contrast to the direct, structured graph querying in Setting~1. While SPARQL queries are fully deterministic and always return the same result for a given ORKG snapshot, LLMs are probabilistic models: their answers can vary slightly from run to run, even when the underlying evidence (here, the curated comparison tables) remains fixed. High agreement with the SPARQL answers in Settings~2 and~3 should therefore be interpreted as an approximation to the symbolic gold standard, rather than an exact guarantee.

\subsubsection{\label{sec:exp-setting} Experimental settings}

\textit{Setting 1: Symbolic querying of structured comparison tables.} In this first setting, we do not use any language models at all. Instead, we directly query the machine-actionable ORKG comparison tables using SPARQL, a structured query language for graphs. For each of the 33 questions, we wrote a corresponding SPARQL query that selects exactly the relevant rows and columns from the curated comparison tables. Because these queries run over structured data and execute deterministically, the resulting answer tables form the gold-standard reference against which we compare all LLM-based settings (Settings~2 and~3).

\textit{Setting 2: Neural querying from PDF-based scientific evidence.} 
In this setting, we ask how well large language models can answer the 33 questions when their only evidence is the review article itself in PDF form. Both proprietary and open-weights models receive the detailed natural-language equivalent of the SPARQL query together with the article PDF (or text and tables extracted from it), but they have \emph{no} direct access to the ORKG representations. The models must therefore internally parse the PDF, locate the relevant passages and tables, and produce an answer table in a single neural inference step. By comparing their outputs against the SPARQL gold standard from Setting~1, we can quantify how closely purely neural, PDF-based querying can approach symbolic performance for precise, table-style scientific questions.

\begin{enumerate}[label=2.\arabic*]
    \item \textit{Proprietary language model (neural) querying.} 
    We first evaluate three state-of-the-art commercially hosted conversational models—ChatGPT-5.1 (OpenAI, release 12.11.2025)~\cite{chatgpt51}, Gemini-3-Pro (Google, release 18.11.2025)~\cite{gemini3}, and Claude-Sonnet-4.5 (Anthropic, release 29.09.2025)~\cite{anthropic2025sonnet45}. As of the date of this submission (13.12.2025), these systems are among the most capable proprietary large language models. Each model is provided with the detailed natural-language equivalent of the SPARQL query and the PDF of the corresponding review article. This setting tests whether such black-box systems, with their own internal PDF-parsing pipelines, can directly answer precise, table-oriented scientific questions from the PDFs alone. A limitation of this setting is that the parsing and context-integration steps cannot be inspected or modified, and runs cannot be exactly reproduced beyond what the providers expose in their interfaces.

    \item \textit{Open-weights language model (neural) querying.} 
    To gain full experimental control and reproducibility, we complement the proprietary systems with high-performing \emph{open-weights} large language models. Unlike proprietary models accessed via paid cloud services, these open-weights models expose their parameters and can be downloaded from platforms such as \url{https://huggingface.co/models} or Ollama \url{https://ollama.com/} and run locally. Once suitable hardware is available, this removes per-query usage costs and additionally provides full control over model versioning and configuration, strict reproducibility of experiments, and the option to keep all data within a local computing environment (which can be important for sensitive datasets).

    For each paper, we first convert the PDF into machine-readable text and tables using \href{https://github.com/jsvine/pdfplumber}{\texttt{pdfplumber}} and our processing script \url{https://github.com/sciknoworg/ald-ale-orkg-review/blob/main/llm-experiments/pdf_extractor.py}. In practical terms, \texttt{pdfplumber} reads each PDF page and extracts the visible text plus the grid structure of tables into plain-text and tabular form. This conversion step is required so that the language models can work with the article content as sequences of text tokens and rows/columns, rather than as a page-layout format optimized for display but whose information is not directly accessible for text processing. We then evaluate five open-weights models:
    \href{https://huggingface.co/google/gemma-3-27b-it}{\texttt{gemma-3-27b-it}},
    \href{https://huggingface.co/meta-llama/Llama-3.3-70B-Instruct}{\texttt{llama-3.3-70b-instruct}},
    \href{https://huggingface.co/Qwen/Qwen2.5-VL-72B-Instruct}{\texttt{qwen2.5-vl-72b-instruct}},
    \href{https://huggingface.co/Qwen/Qwen3-30B-A3B-Instruct-2507}{\texttt{qwen3-30b-a3b-instruct-2507}}, and
    \href{https://huggingface.co/mistralai/Mistral-Large-Instruct-2407}{\texttt{mistral-large-instruct}},
    selected based on their strong performance on recent tabular reasoning benchmarks~\cite{cremaschi2025mammotab}.

    For open-weights models, we consider two variants:
    \begin{enumerate}[label=\alph*)]
        \item \textit{Full-document context (context-injected document QA).} 
        In this variant, all extracted text and tables from a given paper are concatenated into a single long context and passed directly to the model together with the detailed natural-language query (see \href{https://github.com/sciknoworg/ald-ale-orkg-review/blob/main/llm-experiments/context_injected_doc_qa.py}{\texttt{context\_injected\_doc\_qa.py}} in our repository). The model must answer using this complete, but potentially noisy, representation of the article. This mirrors how a human might read the entire paper and then answer the question.
        \item \textit{Retrieval-augmented generation (RAG) over PDF segments.} 
        In the RAG variant, we first split the extracted paper into contiguous text segments (roughly a few pages each) and index these segments using a dense semantic embedding model. At query time, we retrieve the most relevant segment(s) for the detailed natural-language query and supply only these retrieved segments, plus the query, to the model (see \href{https://github.com/sciknoworg/ald-ale-orkg-review/blob/main/llm-experiments/rag_pdf_segments_doc_qa.py}{\texttt{rag\_pdf\_segments\_doc\_qa.py}}). This reduces context length and focuses the model on the parts of the paper most likely to contain the answer, but it can miss information if key table rows or explanations are spread across distant segments.
    \end{enumerate}
\end{enumerate}

A detailed, technical version of setting 2 description including our implementation choices for PDF extraction, segmentation, embedding, and retrieval hyperparameters are provided in the Appendix~\ref{appsec:setting2}.

\textit{Setting 3: Symbolic-context-augmented neural querying.} 
In this setting, we bring the language models as close as possible to the symbolic baseline by grounding them directly in the machine-actionable ORKG comparison tables instead of the raw PDFs. Concretely, we export each ORKG comparison into a flat, spreadsheet-like format (CSV), as described in Sec.~\ref{sec:exp-dataset}, and provide this CSV together with the detailed natural-language query as input to the models. This data is publicly released~\cite{dsouza_ald-e_2025}. We evaluate the same proprietary and open-weights models as in Setting~2, but here they never see the PDFs or their extracted text. All reasoning is performed over the CSV-encoded comparison tables, which are \emph{identical in content} to those queried via SPARQL in Setting~1; only the query engine changes, from a deterministic symbolic executor to a probabilistic neural model.

We refer to this as \emph{symbolic-context-augmented} querying because the language model is not only given a natural-language question, but also a \emph{symbolic} representation of the evidence: the ORKG comparison table. In AI and data science, ``symbolic'' typically denotes knowledge represented as explicit symbols and relations (for example, rows and columns in a table or nodes and edges in a graph) that can be queried with formal languages such as SPARQL, rather than as free-running text. In our case, each material, precursor, temperature, etch rate, or EQE value appears in a dedicated column with a well-defined meaning, instead of being embedded in sentences in the PDF. Setting~3 therefore lets us study how well language models can approximate the SPARQL gold standard when their input context includes this structured, symbol-like table representation, rather than only heterogeneous PDF layouts. We do not, however, build a specialised neurosymbolic architecture (such as a separate symbolic reasoning module inside the model); the symbolic information is provided purely as richer input context.

It is important to note that the query engine in Setting~1 is fully deterministic: the same SPARQL query over a fixed ORKG snapshot always returns the same result. By contrast, the neural query engines in Settings~2 and~3 are non-deterministic because their text generation involves random choices; repeated runs can yield slightly different answers, even when the input is unchanged. In Setting~3, the model must still perform the essential operations that SPARQL carries out explicitly: identify the relevant rows in the comparison table (for example, all entries with a given material or precursor), select the correct columns for the requested output, and sometimes compare or combine values across rows. Our evaluation therefore asks to what extent a symbolically grounded neural Q\&A system can match the numerical accuracy of the SPARQL baseline (Setting~1), while offering a flexible natural-language front end.

Across all ALD and ALE questions, we applied the 33 queries to 21 distinct model–setting configurations (for example, each model evaluated in Settings~2a, 2b, and~3), resulting in a total of 693 individual experiments. This provides a sizeable basis for comparing symbolic, neural, and symbolic-context-augmented querying.

\subsubsection{\label{sec:eval-metrics} Evaluation metrics}

We use different evaluation metrics aligned with our three research questions.

\paragraph{Expert judgments on SPARQL queries (RQ1).}
To address RQ1, we conducted a small survey with three ALD/ALE domain experts (two of whom are co-authors of this paper). For each of the 33 queries (19 ALD, 14 ALE), experts ran the SPARQL query on the corresponding ORKG comparison using a provided ORKG link and SPARQL link, inspected the resulting table, and then rated it along two dimensions:
\begin{itemize}
    \item \emph{Meaningfulness}: Does the question make scientific sense (for example, does it ask about plausible relationships, use appropriate concepts, and align with current ALD/ALE knowledge)?
    \item \emph{Usefulness}: Is the resulting table a helpful synthesis for researchers (for example, by saving time, surfacing patterns that are hard to see by eye, or providing non-trivial insight beyond reading the review once)?
\end{itemize}
Both dimensions were rated on a 5-point Likert scale (1 = strongly disagree, 5 = strongly agree). Experts evaluated all 33 queries spanning seven ALD and eleven ALE comparison tables. For each query, they first judged whether the natural-language question itself was meaningful, and then whether the corresponding SPARQL result table would be useful as a synthesis tool under the assumption that it could be continuously updated with additional studies in the future. Separating meaningfulness from usefulness helps diagnose whether potential improvements are needed in (i) how the question is formulated, (ii) how the underlying data are modeled in the comparison, or (iii) how the synthesized results are presented. Whenever the usefulness score was below~3, experts were asked to briefly justify their rating in a free-text field, providing qualitative feedback on limitations or missing information. The survey questionnaire is available at
\url{https://forms.gle/qSDUG1WzSZSz2864A}; it is shared for transparency, and interested readers are welcome to inspect it or contribute their own responses.

\paragraph{Agreement between neural and SPARQL tables (RQ2 and RQ3).}
To address RQ2 and RQ3, we measure how closely the tables produced by the neural systems match the SPARQL result tables, on a per-query basis. For this, we adopt the \emph{Relative Mapping Similarity} (RMS) metric introduced by Liu et al.~\cite{liu-etal-2023-deplot}, originally developed for evaluating table and chart question answering systems.

Intuitively, RMS treats each table as a set of entries of the form
\((\text{row header}, \text{column header}, \text{cell value})\) and compares the predicted and gold tables by matching these entries. A high RMS score is obtained only when (i) the model selects the correct rows and columns (for example, the right materials or processes) and (ii) the corresponding values (for example, temperatures, etch rates, or EQEs) are close to those in the SPARQL table. RMS produces precision, recall, and F1 scores in the range \([0,1]\), and is by construction insensitive to row/column order or table transposition.

In our experiments, for each query we treat the SPARQL result as the gold table and the output of each neural system (PDF-based and symbolic-context-augmented settings) as the prediction. We compute RMS precision, recall, and F1 using the official DePlot implementation from Google Research, adapted to our ALD/ALE dataset; our RMS evaluation script and configuration are documented in
\url{https://github.com/sciknoworg/ald-ale-orkg-review/blob/main/eval_deplot_rms/README.md}.
The full mathematical definition of RMS and implementation details are provided in the Appendix~\ref{appsec:rms}.

\begin{figure*}[!htb]
    \centering
    \begin{subfigure}[t]{\linewidth}
        \centering
        \includegraphics[width=\linewidth]{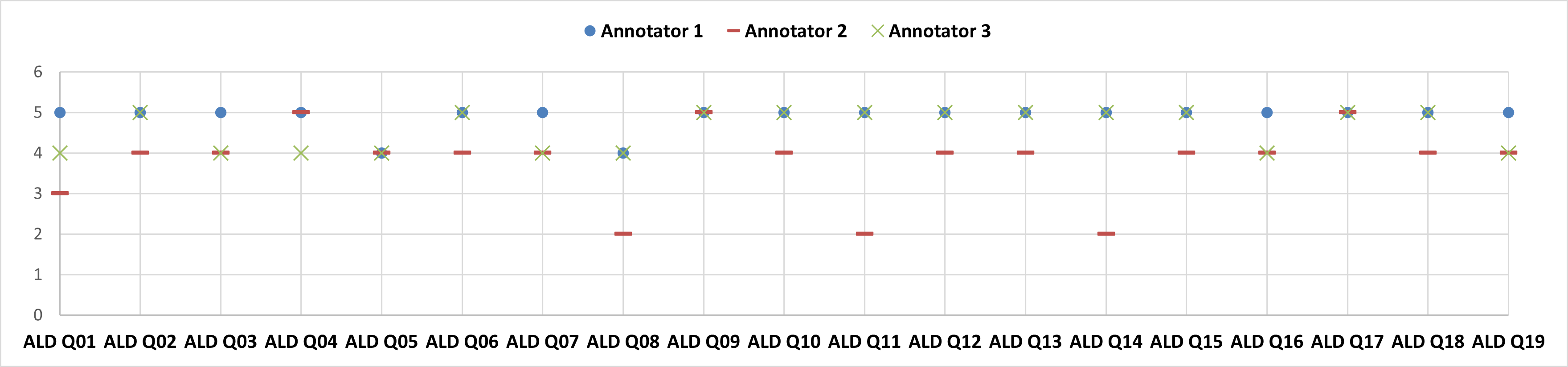}
        \caption{\label{fig:ALD-meaningful}Per-query meaningfulness ratings for the 19 ALD queries.}
    \end{subfigure}
    \begin{subfigure}[t]{\linewidth}
        \centering
        \includegraphics[height=3.3cm, keepaspectratio]{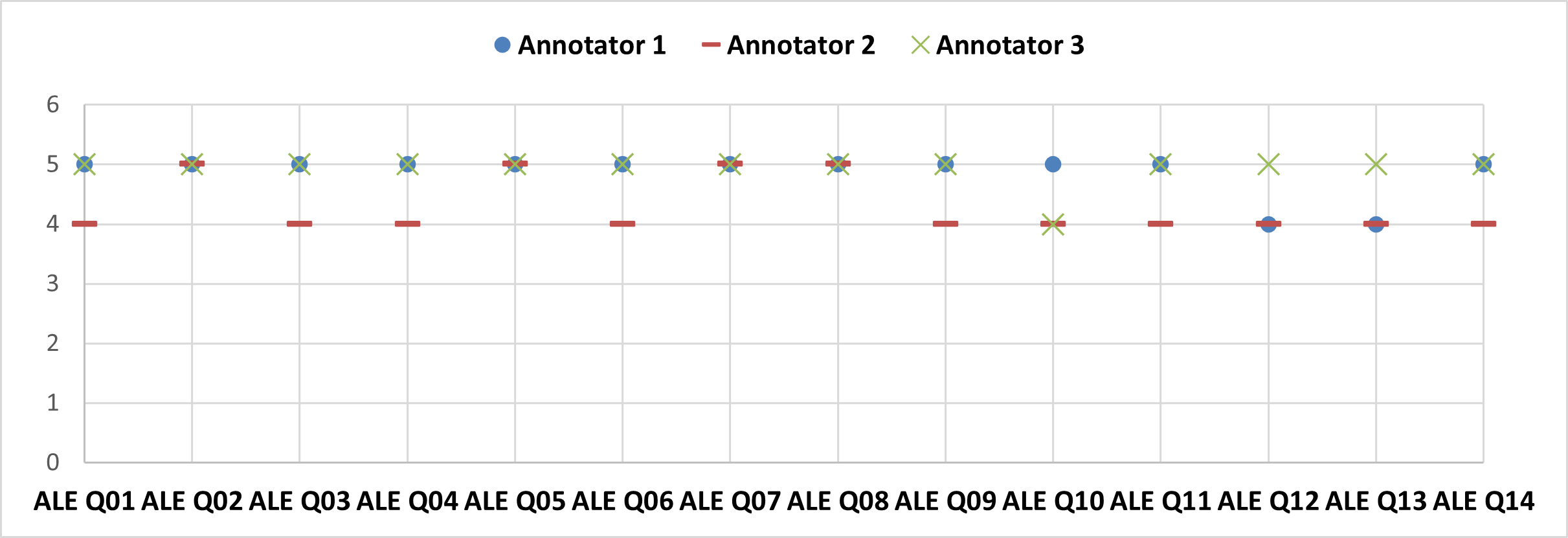}
        \caption{\label{fig:ALE-meaningful}Per-query meaningfulness ratings for the 14 ALE queries.}
    \end{subfigure}
    \caption{\label{fig:meaningful}Meaningfulness ratings assigned by three domain experts for all ALD (a) and ALE (b) queries, on a 5-point Likert scale (1 = strongly disagree, 5 = strongly agree) in response to the prompt ``the query is meaningful''.}
\end{figure*}

\begin{figure*}[!htb]
    \centering
    \begin{subfigure}[t]{\linewidth}
        \centering
        \includegraphics[width=\linewidth]{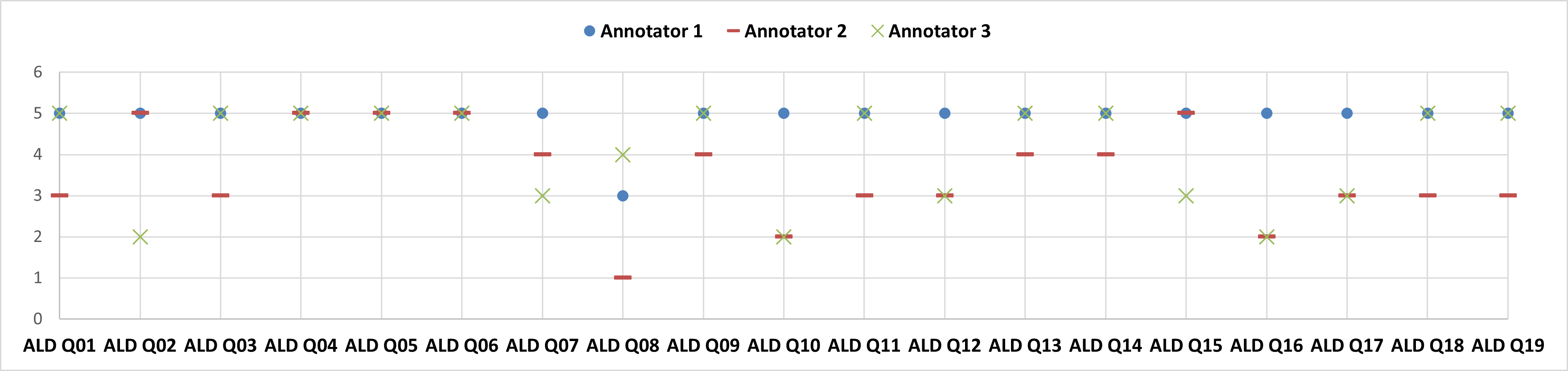}
        \caption{\label{fig:ALD-useful}Per-query usefulness ratings for the 19 ALD queries.}
    \end{subfigure}
    \begin{subfigure}[t]{\linewidth}
        \centering
        \includegraphics[height=3.3cm, keepaspectratio]{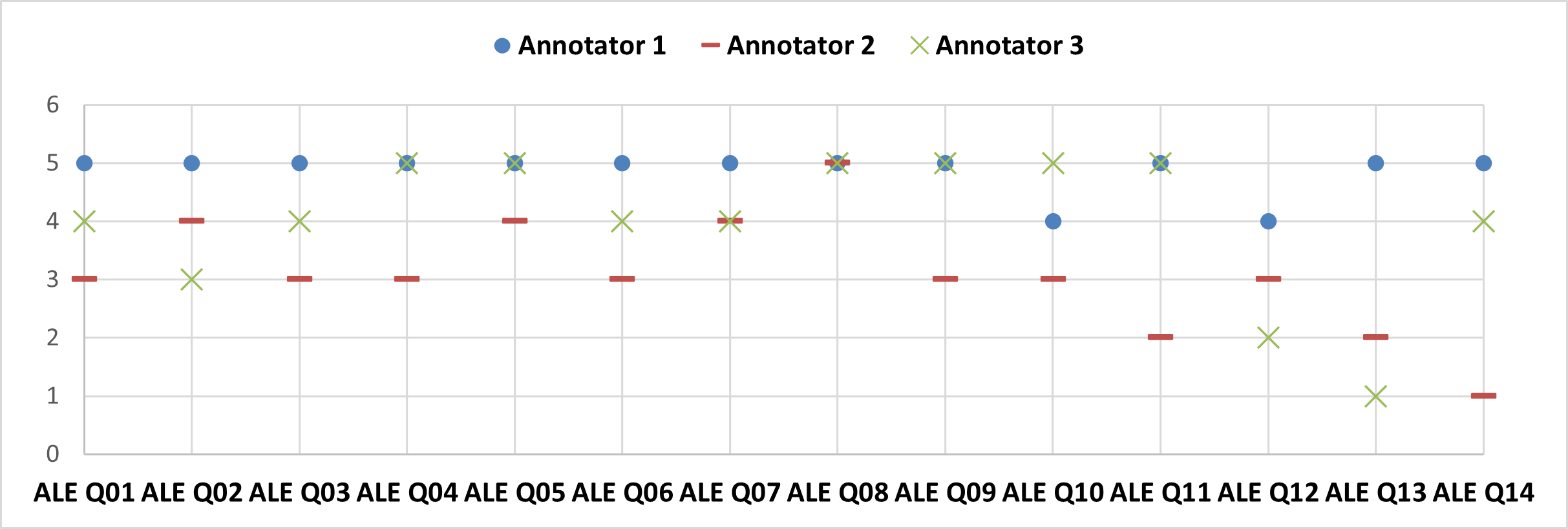}
        \caption{\label{fig:ALE-useful}Per-query usefulness ratings for the 14 ALE queries.}
    \end{subfigure}
    \caption{\label{fig:useful}Usefulness ratings assigned by three domain experts for all ALD (a) and ALE (b) queries, on a 5-point Likert scale (1 = strongly disagree, 5 = strongly agree) in response to the prompt: ``The query result provides a useful synthesis of the insights in the table'' (with the instruction to imagine a richer, continuously updated version of the table when the original data are sparse).}
\end{figure*}

\subsection{Results}

\subsubsection{Domain expert impressions of machine-actionable knowledge synthesis}

Here we address RQ1: how useful ALD/ALE experts find ORKG-based comparison tables for answering complex review questions. As described in the Evaluation Metrics section \ref{sec:eval-metrics}, three ALD/ALE domain experts (two co-authors and one external senior researcher, with between 1--5 and $>10$ years of experience in materials science) completed an online survey in which they evaluated all 33 queries (19 ALD, 14 ALE). For each query, they ran the SPARQL query on the corresponding ORKG comparison, inspected the resulting table, and rated (i) whether the question was scientifically meaningful and (ii) whether the table provided a useful synthesis.

Overall, the response was clearly positive. Experts judged almost all questions to be scientifically well-posed: the average meaningfulness score across all queries was $4.54/5$, with 96\% of ratings at~$\geq 4$, and 32 out of 33 queries achieving an average score of at least~4. The per-query distributions of meaningfulness ratings for ALD and ALE are shown in \autoref{fig:meaningful}. Usefulness scores were slightly lower but still favourable (mean $4.08/5$), with 25 of 33 queries reaching an average usefulness of at least~4 and only two queries falling below~3; \autoref{fig:useful} visualises the corresponding usefulness ratings. In a separate overall question on the value of machine-actionable scientific knowledge, all three experts rated it between 4 and 5 (mean $4.67$). In other words, experts not only regard the questions as meaningful but also see the resulting SPARQL-derived tables as genuinely helpful knowledge syntheses. Detailed per-query statistics and inter-annotator agreement analyses are provided in Appendix~\ref{appsec:stats-for-survey}.

The qualitative feedback explains why. Experts repeatedly highlighted that being able to call up comparison tables ``in one click'' would substantially speed up literature review and make it easier to compare process conditions across many studies, especially for tasks such as precursor screening or identifying suitable process windows. They emphasised that such machine-actionable tables could support both academic and industrial work, for example by quickly surveying which precursors and co-reactants have been tried for a given material, or by mapping out temperature and plasma power ranges where acceptable growth or etch rates are reported.

At the same time, the low usefulness scores and critical remarks point to concrete areas for improvement rather than fundamental objections. Most reservations concern:
\begin{itemize}
    \item \textbf{Missing or incomplete parameters}, such as sticking coefficients, surface orientation and termination, clearer precursor/co-reactant distinctions, dopant information, error bars on growth-per-cycle, or more detailed timing parameters per ALD/ALE cycle.
    \item \textbf{Schema and categorisation refinements}, including more consistent treatment of mechanisms and precursor families, and clearer separation of processes that practitioners would not class as ``true'' ALD/ALE.
    \item \textbf{Table readability}, such as avoiding duplicate rows, making very long tables more compact, and clearly exposing key identifiers (e.g., material and precursor names) rather than hiding them behind opaque resource IDs.
    \item \textbf{Data completeness and trust}, especially in sparsely populated comparisons or in scenarios where tables might be (partially) extracted automatically from PDFs.
\end{itemize}

The ``wish-list'' questions provide a particularly concrete picture of how experts would like to use such tables. Examples include querying which surface orientations and terminations are most commonly reported for a given material, grouping ALD recipes by precursor/co-reactant pair and visualising growth-per-cycle as a function of temperature, plasma power, and reactor type, or aggregating ALE conditions by material, precursor, reactant, and process temperature. These examples go beyond simple lookups and mirror the multi-parameter reasoning that ALD/ALE researchers already perform manually.

Experts also identified several factors that would affect adoption. Barriers include the additional effort required to enter machine-actionable tables alongside traditional PDFs, uneven reporting practices in the literature, concerns about quality control for automatically extracted data, and the current lack of community standards for which process and performance parameters should be captured. On the positive side, clear incentives include demonstrable time savings in literature review, community-wide schema standardisation, user-friendly tools for authoring and querying ORKG comparisons, and visible benefits for day-to-day process development (for example, being able to answer complex comparative questions ``in one click'').

Taken together, these findings answer RQ1 positively: ALD/ALE experts view ORKG-based machine-actionable reviews as both scientifically meaningful and practically useful. Their feedback mainly points to next steps in schema refinement, data coverage, and tooling, rather than to conceptual objections. In the remainder of this section, we therefore treat the SPARQL-derived tables as an expert-validated symbolic baseline and investigate how closely neural systems operating over PDFs and over ORKG comparisons can approximate this baseline (RQ2 and RQ3). The complete domain expert survey responses, including individual ratings and free-text comments, are available in our Zenodo repository~\cite{dsouza_ald-e_2025}.

\begin{table}[!htb]
\centering
\footnotesize
\caption{\label{tab:all-settings-results}
Evaluation results across all experimental settings, reported as macro-averaged Relative Mapping Similarity (RMS) precision, recall, and F$_1$ (in \%). Setting~1 is the symbolic SPARQL baseline over ORKG comparisons. Setting~2 comprises neural querying over PDFs (2.1: proprietary models; 2.2a: open-weights with a single concatenated PDF context; 2.2b: open-weights with RAG over embedded PDF chunks). Setting~3 comprises symbolic-context-augmented neural QA over ORKG comparison CSVs (3.1: proprietary models; 3.2: open-weights). The ``\#queries'' column indicates, for each system, how many of the 33 ALD+ALE queries produced a valid prediction that could be evaluated against the SPARQL gold-standard tables. Open-weights shorthand names: ``gemma-3'' = gemma-3-27b-it; ``llama-3.3'' = Llama-3.3-70B-Instruct; ``mistral'' = Mistral-Large-Instruct-2407; ``qwen2.5-vl'' = Qwen2.5-VL-72B-Instruct; ``qwen3'' = Qwen3-30B-A3B-Instruct-2507.}
\resizebox{\textwidth}{!}{%
\begin{tabular}{clp{0.32\textwidth}lcccc}
\hline
\textbf{Setting} & \textbf{Model type} & \textbf{Context} & \textbf{Model} & \textbf{\#queries} & \textbf{RMS-prec. [\%]} & \textbf{RMS-rec. [\%]} & \textbf{RMS-F1 [\%]} \\
\hline
1 & Symbolic & ORKG comparison (SPARQL over graph) & SPARQL 
  & 33 & 100.0 & 100.0 & 100.0 \\ \hline \hline

\multirow{3}{*}{2.1} 
  & \multirow{3}{*}{Prop.} 
  & \multirow{3}{0.32\textwidth}{Review article PDF (black-box proprietary PDF processing)} 
  & chatgpt-5.1 & 33 & 46.0 & 39.5 & 40.6 \\
 &  &  & claude-sonnet & 33 & 61.4 & 63.7 & 61.7 \\
 &  &  & gemini-3 & 33 & 59.6 & 60.0 & 59.1 \\ \hline

\multirow{5}{*}{2.2a} 
  & \multirow{5}{*}{Open} 
  & \multirow{5}{0.32\textwidth}{PDF text and tables (single concatenated context)} 
  & gemma-3     & 33 & 25.5 & 25.1 & 24.6 \\
 &  &  & llama-3.3   & 33 & 32.6 & 35.9 & 32.8 \\
 &  &  & mistral     & 32 & 28.4 & 27.1 & 25.9 \\
 &  &  & qwen2.5-vl  & 33 & 34.1 & 27.6 & 28.2 \\
 &  &  & qwen3       & 32 & 29.1 & 33.9 & 29.7 \\ \hline

\multirow{5}{*}{2.2b} 
  & \multirow{5}{*}{Open} 
  & \multirow{5}{0.32\textwidth}{PDF text and tables (RAG over embedded chunks)} 
  & gemma-3     & 33 & 21.4 & 18.9 & 19.0 \\
 &  &  & llama-3.3   & 33 & 25.0 & 23.8 & 22.8 \\
 &  &  & mistral     & 33 & 23.8 & 20.9 & 21.1 \\
 &  &  & qwen2.5-vl  & 33 & 28.2 & 24.1 & 24.3 \\
 &  &  & qwen3       & 33 & 24.7 & 24.6 & 23.2 \\ \hline \hline

\multirow{3}{*}{3.1} 
  & \multirow{3}{*}{Prop.} 
  & \multirow{3}{0.32\textwidth}{ORKG comparison exported as CSV (symbolic table context)} 
  & chatgpt-5.1 & 33 & 67.8 & 61.5 & 63.0 \\
 &  &  & claude-sonnet & 33 & \textbf{74.0} & \textbf{76.1} & \textbf{74.2} \\
 &  &  & gemini-3      & 33 & 72.2 & 71.5 & 70.3 \\ \hline

\multirow{5}{*}{3.2} 
  & \multirow{5}{*}{Open} 
  & \multirow{5}{0.32\textwidth}{ORKG comparison exported as CSV (symbolic table context)} 
  & gemma-3     & 33 & 44.8 & 47.4 & 44.7 \\
 &  &  & llama-3.3   & 33 & \textbf{48.5} & \textbf{52.8} & \textbf{49.3} \\
 &  &  & mistral     & 31 & 51.4 & 51.2 & 48.8 \\
 &  &  & qwen2.5-vl  & 32 & 46.0 & 37.9 & 39.3 \\
 &  &  & qwen3       & 32 & 40.8 & 46.0 & 40.8 \\
\hline
\end{tabular}}
\end{table}

\subsubsection{Neural querying over PDFs: performance on precise scientific Q\&A tasks}

Here we address RQ2, i.e., how well large language models that only see the review-article PDFs can reproduce the precise, table-style answers obtained from symbolic SPARQL querying. The corresponding Relative Mapping Similarity (RMS) scores (introduced in \autoref{sec:eval-metrics}) are summarised in \autoref{tab:all-settings-results} for Settings~2.1, 2.2a, and 2.2b. In Setting~2.1, we use the chat interfaces of three proprietary models (ChatGPT-5.1, Gemini-3-Pro, and Claude-Sonnet-4.5), uploading the full PDF and treating their internal PDF handling as a black box. In Settings~2.2a and~2.2b, we evaluate open-weights models using a transparent pipeline: text and tables are first extracted from each PDF and then provided as input, either as one long context (2.2a) or via a simple retrieval step over document segments (2.2b). Implementation details are in the Appendix.

Within this purely PDF-based regime, the proprietary models in Setting~2.1 perform best. Claude-Sonnet-4.5 reaches the highest RMS F$_1$ at 61.7\%, followed by Gemini-3-Pro at 59.1\%, with ChatGPT-5.1 at 40.6\%. Even this weakest proprietary model outperforms all open-weights systems in Settings~2.2a and~2.2b, indicating that the proprietary PDF-processing and document-structure handling currently offer a substantial advantage over our simple open-source pipeline. Among the open-weights models with a single concatenated context (Setting~2.2a), the best performer, Llama-3.3-70B-Instruct, attains 32.8\% RMS F$_1$, with the remaining models clustered around 24–30\%. Moving to the retrieval-augmented variant (Setting~2.2b) consistently \emph{decreases} performance: all five open-weights models drop into the 19–24\% RMS F$_1$ range, suggesting that our straightforward “retrieve a few segments and answer from them’’ strategy is not yet sufficient for reliable table-style question answering and can even omit important evidence.

Overall, these findings show that purely neural PDF-based querying, although workable to some extent, remains substantially below the symbolic SPARQL upper bound and is tightly constrained by the quality of the PDF-processing and context-selection pipeline. The proprietary systems currently set the bar in this regime, but even the best of them recovers only about 60\% of the SPARQL RMS F$_1$ signal. For open-weights models, the gap widens further, and naïve retrieval does not close it. This motivates the next subsection (RQ3), where we remove the PDF bottleneck and ground the same models directly in the machine-actionable ORKG comparison tables.

\begin{figure*}[!htb]
    \centering
    \includegraphics[width=\linewidth]{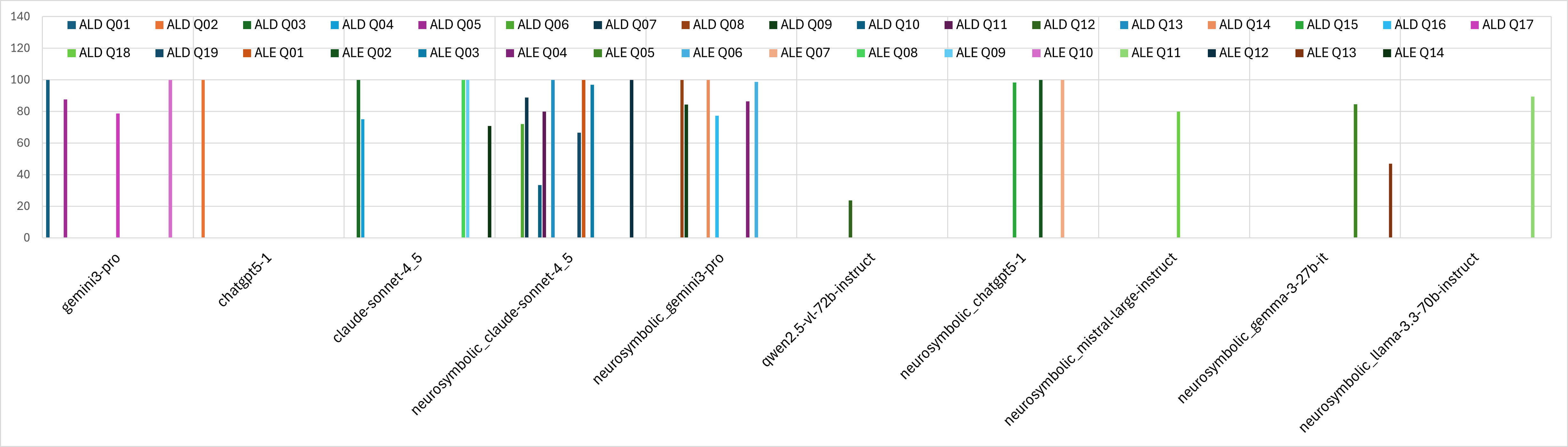}
    \caption{\label{fig:neurosymbolic}Best RMS F$_1$ score obtained for each query. For every ALD and ALE query, we identify the system with the highest RMS F$_1$ score when compared against the SPARQL gold-standard table. Among the 21 systems evaluated, 10 achieve the best RMS F$_1$ score on at least one query. Here, ``best'' denotes the highest RMS F$_1$ score attained for a given query across the 21 systems, not necessarily a perfect score of 100\%.}
\end{figure*}

\begin{figure*}[!htb]
    \centering
    \begin{subfigure}[t]{0.49\linewidth}
        \centering
        \includegraphics[width=\linewidth]{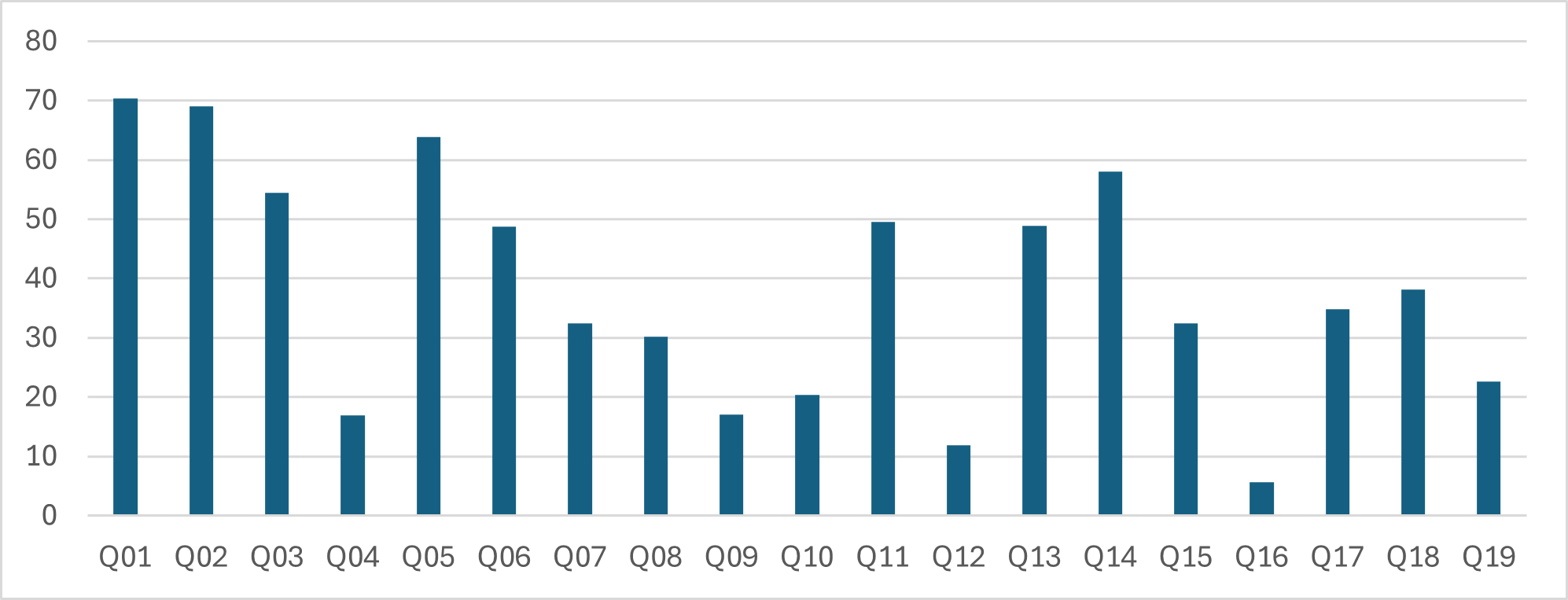}
        \caption{\label{fig:ALD-per-query}Mean RMS F$_1$ scores for the 19 ALD queries, averaged across 21 systems.}
    \end{subfigure}
    \begin{subfigure}[t]{0.49\linewidth}
        \centering
        \includegraphics[height=2.6cm, keepaspectratio]{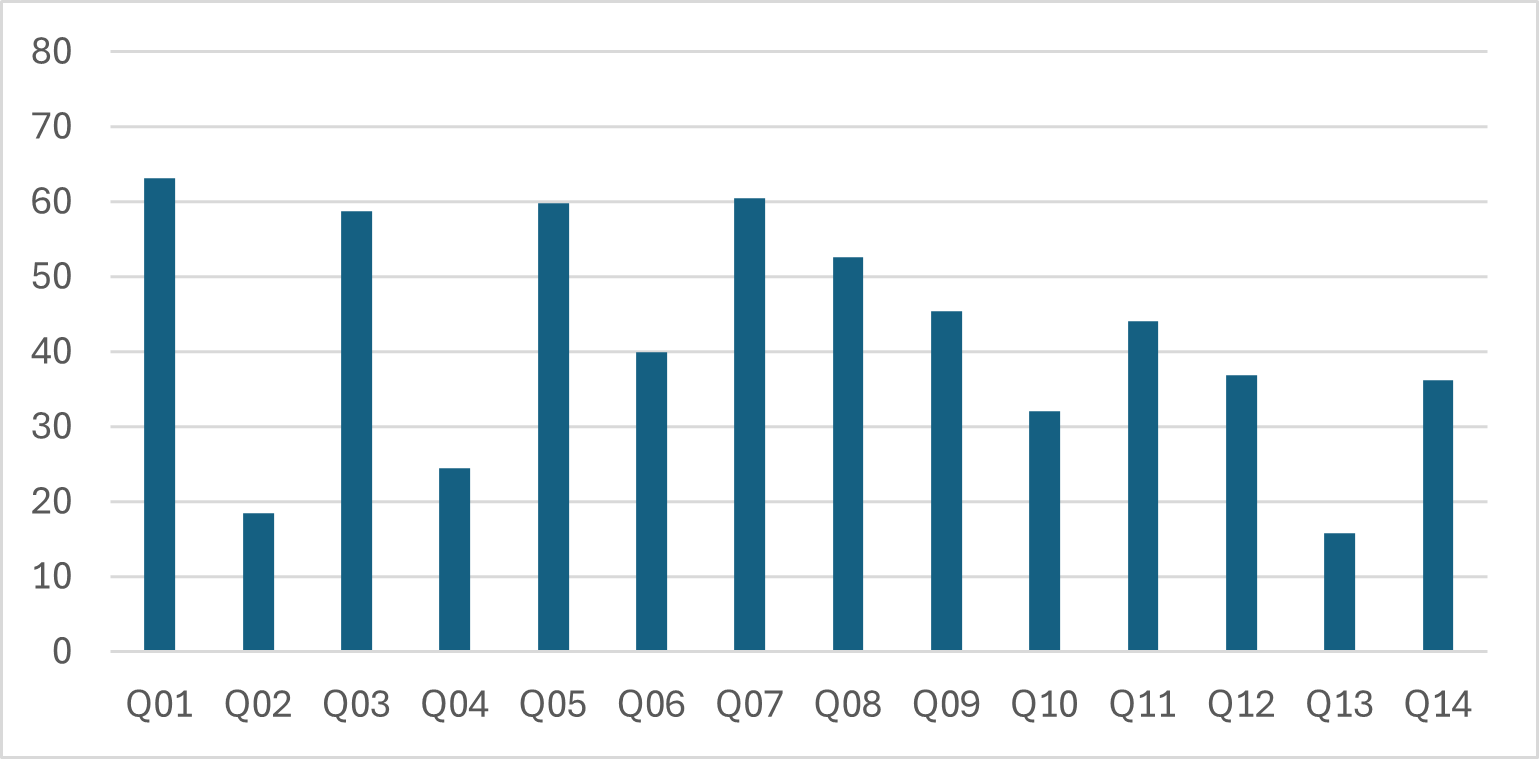}
        \caption{\label{fig:ALE-per-query}Mean RMS F$_1$ scores for the 14 ALE queries, averaged across the same 21 systems.}
    \end{subfigure}
    \caption{\label{fig:fscore}Per-query Relative Mapping Similarity (RMS) F$_1$ scores for ALD (left) and ALE (right). For each query, we compute RMS F$_1$ between each system’s predicted table and the SPARQL gold-standard table, then average the F$_1$ values over the 21 systems. Higher mean RMS F$_1$ scores indicate queries that are easier for the models to reproduce the symbolic SPARQL results, whereas lower scores indicate more challenging queries in the LLM setting.}
\end{figure*}

\begin{figure*}[!htb]
    \centering
    \begin{subfigure}[t]{0.49\linewidth}
        \centering
        \includegraphics[width=\linewidth]{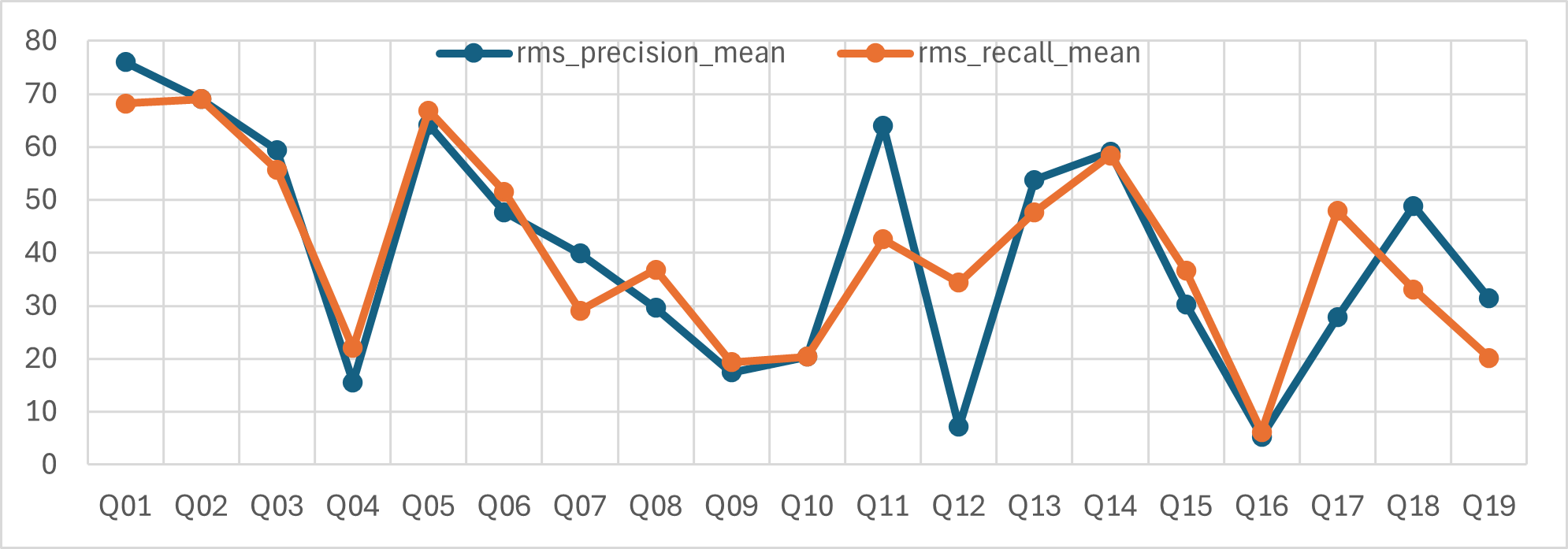}
        \caption{\label{fig:ALD-recall-prec}Mean RMS precision and recall for the 19 ALD queries, averaged across 21 neural and neurosymbolic systems.}
    \end{subfigure}
    \begin{subfigure}[t]{0.49\linewidth}
        \centering
        \includegraphics[height=2.4cm, keepaspectratio]{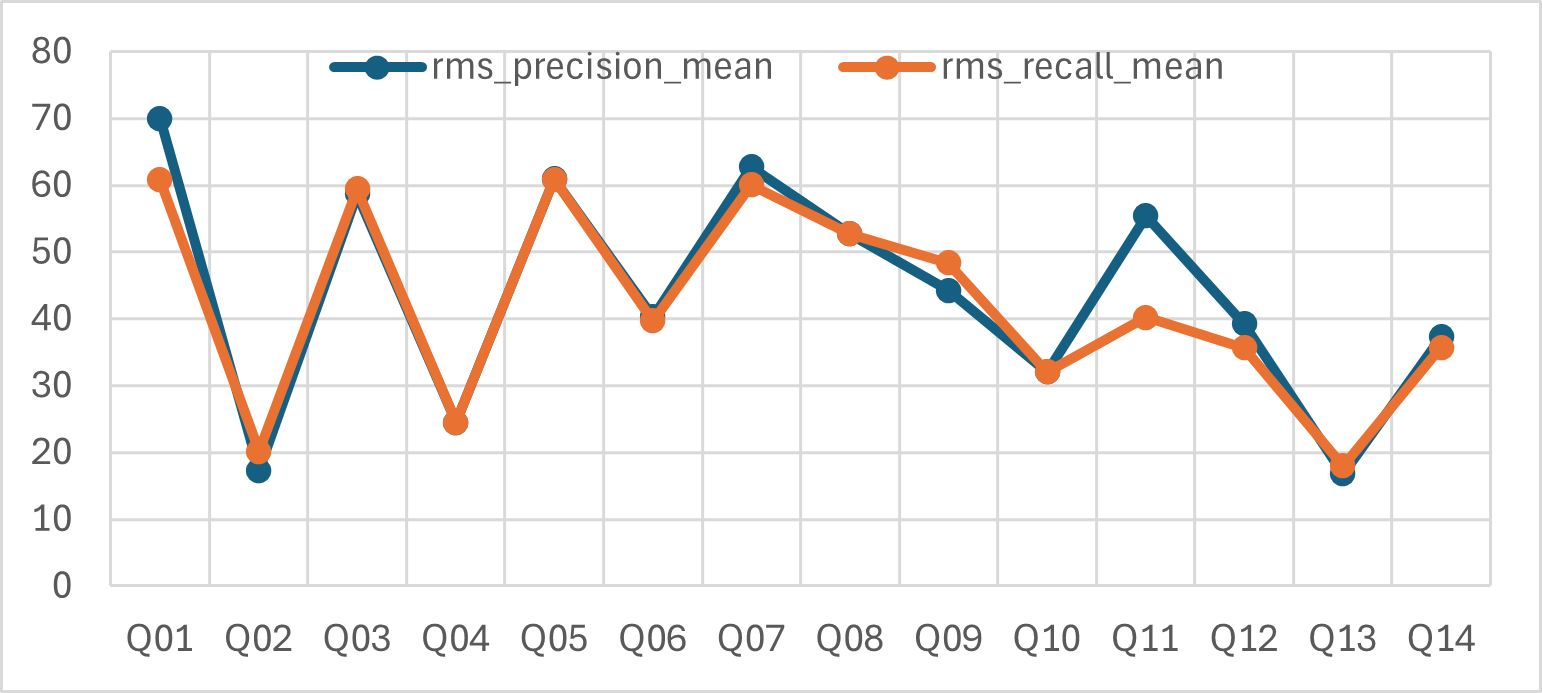}
        \caption{\label{fig:ALE-recall-prec}Mean RMS precision and recall for the 14 ALE queries, averaged across the same 21 systems.}
    \end{subfigure}
    \caption{\label{fig:recall-prec}Per-query Relative Mapping Similarity (RMS) precision and recall for ALD (left) and ALE (right). For each query, we compute RMS precision and recall between each system’s predicted table and the SPARQL gold-standard table, then average these scores across the 21 systems. Points where the mean precision and recall curves coincide indicate that systems tend to reproduce the gold-standard mappings without adding or omitting entries. In contrast, a noticeable gap with substantially lower precision than recall indicates hallucination (many additional entries not supported by SPARQL), whereas higher precision than recall indicates under-coverage, i.e., models tend to produce only a conservative subset of the gold-standard entries.}
\end{figure*}

\subsubsection{Neural querying over ORKG comparisons: benefits of symbolic grounding}

From an ALD/ALE perspective, RQ3 asks a simple question: if we give LLMs the \emph{same kind of structured table} that experts use (the ORKG comparisons), do they get closer to the “ideal’’ SPARQL answers than when they have to read the PDFs by themselves? In contrast to Setting~2 (PDF-based querying), Setting~3 therefore uses the same proprietary and open-weights models but restricts their input to CSV-serialized ORKG comparison tables. Each model receives the exported comparison table together with the detailed natural-language query. This isolates neural reasoning over fully structured, machine-actionable evidence and lets us compare symbolic-context-augmented neural querying directly to the symbolic SPARQL gold standard (Setting~1) and to purely neural PDF-based querying (Setting~2). Relative Mapping Similarity (RMS) scores for all systems (introduced in \autoref{sec:eval-metrics}) are summarised in \autoref{tab:all-settings-results}.

In \autoref{tab:all-settings-results}, the last two main blocks (Settings~3.1 and~3.2) show a clear gain from symbolic grounding. With ORKG tables as context, all three proprietary models improve substantially over their PDF-only counterparts, and the best model now exceeds 70\% RMS F$_1$. Claude-Sonnet-4.5 reaches 74.2\% RMS F$_1$, Gemini-3-Pro 70.3\%, and ChatGPT-5.1 63.0\%. For an ALD/ALE reader, this means that when these systems are given a clean process table rather than a full PDF, roughly two thirds to three quarters of the rows and values in their answers match the SPARQL baseline. In the open-weights regime, Llama-3.3-70B-Instruct remains the strongest model and rises to 49.3\% RMS F$_1$ in Setting~3.2, with Mistral-Large-Instruct-2407 close behind at 48.8\%. Symbolic table context thus lifts performance for \emph{all} model families and noticeably narrows (though does not close) the gap to the SPARQL upper bound of 100\%.

\autoref{fig:neurosymbolic} shows, for each ALD and ALE query, which system achieves the highest RMS F$_1$ when compared to the SPARQL result. Most queries are “won’’ by systems operating in the symbolically grounded Setting~3, not by the PDF-only Settings~2.1–2.2. Put differently: whenever some model in our pool can solve a query reasonably well, it almost always does so when it sees the curated ORKG comparison table rather than the original PDF. This directly supports our central hypothesis that structured, machine-actionable evidence stores are more effective for precise information and knowledge synthesis than relying solely on the generative capabilities of LLMs over unstructured documents.

At the same time, symbolic grounding does not make all questions easy. \autoref{fig:fscore} shows, for each query, the average RMS F$_1$ across all 21 systems. Some questions reach relatively high average scores and can be answered reliably by many models once the ORKG tables are available. Others remain difficult even in Setting~3, with mean RMS F$_1$ sometimes below 20\%. These harder cases tend to combine several ingredients: tight numeric constraints (e.g., simultaneous bounds on temperature, thickness, or etch rate), information scattered across long or multi-page tables, and the need to group or aggregate by higher-level categories (such as precursor families, linker types, or device classes) or to join information across multiple tables. For such multi-step reasoning tasks, small mistakes in filtering, grouping, or averaging quickly translate into missing or incorrect rows and are heavily penalised by the RMS metric.

\autoref{fig:recall-prec} adds a complementary view by plotting mean RMS precision and recall per query. When precision and recall are close, systems tend to reproduce the SPARQL mappings without adding many spurious entries or omitting many gold-standard rows. Noticeably lower precision than recall signals hallucination: models often output extra table entries that are not supported by the SPARQL results. Higher precision than recall indicates under-coverage, where systems return only a conservative subset of the true entries. Moving from PDFs (Setting~2) to ORKG tables (Setting~3) substantially reduces both hallucination and under-coverage, but does not eliminate them entirely, especially for the more complex queries described above.

Overall, RQ3 can be answered positively for ALD/ALE practice: when LLMs are grounded in structured ORKG comparison tables, they move significantly closer to the symbolic SPARQL baseline and behave in a more controlled way than when they work from PDFs alone. At the same time, the remaining errors point to clear avenues for improvement: richer and more complete comparison schemas, better handling of multi-constraint numeric filters and cross-table joins, and interaction patterns that guide models toward systematic filtering and aggregation rather than ad-hoc text generation. All natural-language queries, SPARQL formulations, answer tables, and ORKG comparisons are released as the \textit{ALD/E Query Dataset} on Zenodo (DOI: \url{https://doi.org/10.5281/zenodo.17720429})~\cite{dsouza_ald-e_2025}, enabling inspection, reuse, and extension by the ALD/ALE community.

\section{Discussion}

Our experiments put two very different ways of answering ALD/ALE questions side by side and make their strengths and weaknesses tangible. On the one hand, symbolic querying via SPARQL over ORKG comparisons is fast (microseconds per query on our hardware), deterministic (the same query over the same data always gives the same answer), and operates over an explicit schema of materials, processes, and performance descriptors. This matches how many JVSTA readers already think about process data: as rows and columns in a well-defined table or database.

On the other hand, neural querying over PDFs behaves very differently. When we ask LLMs to read the full review-article PDFs and generate table-style answers, queries take minutes rather than microseconds, results can change from run to run, and the models are prone to hallucinations, misread tables, and basic numerical errors—even when we provide carefully engineered natural-language versions of the SPARQL questions. In several cases, long chains of “step-by-step reasoning’’ cause the model to lose track of its own logic and drift into repetitive or partially inconsistent answers instead of converging on the precise comparison table we need.

Feeding LLMs the ORKG tables improves the situation but does not fully solve it. In the ORKG-augmented setting (Setting~3), we replace PDFs by compact, schema-aligned comparison tables exported as CSV. This clearly boosts answer quality and reduces some hallucinations: models move substantially closer to the SPARQL baseline and behave more predictably. However, they remain slower than direct SPARQL querying, still show some run-to-run variation, and still occasionally misinterpret column semantics or numeric values. In other words, structured tables make LLMs much more useful, but they do not turn them into perfect drop-in replacements for symbolic querying.

These observations underline that our main contribution is to make the case for \emph{keeping} a symbolic layer for scientific knowledge, rather than claiming that we have built a full neurosymbolic system. Our Setting~3 is best understood as symbolically grounded neural question answering: the ORKG comparison (a structured, human-curated table) constrains and guides a language model that remains fundamentally stochastic, i.e., it samples from a probability distribution over possible next tokens in the generated sequence, so that the same question can yield slightly different outputs from run to run. We do not change the model architecture or training objective, and we do not integrate a logical reasoning module inside the network. Instead, we show empirically that even the strongest LLMs should not be treated as substitutes for a symbolic query engine in deterministic scientific applications. At their best, they approximate the outputs of SPARQL and benefit dramatically from being \emph{fed} symbolic structure that has been curated and maintained in a knowledge graph such as the ORKG.

Looking ahead, our results suggest that symbolic question answering and LLM-based question answering should be viewed as \emph{complementary} for ALD/ALE research. Symbolic querying over ORKG via SPARQL is strongest when the goal is exact, repeatable filtering and aggregation of well-defined quantities (e.g., “which ALD recipes for material~X satisfy a given temperature window and growth-per-cycle constraint?”). In these cases, every row and every number matters, and the symbolic layer provides hard guarantees: the same query over the same data always returns the same result, with transparent provenance. LLM-based question answering, by contrast, is at its best for more open-ended, exploratory, or integrative tasks: explaining mechanisms, relating processes to broader trends, or combining scattered observations into a coherent narrative. This view is consistent with recent ALD benchmarking work, which evaluates GPT-4-class models as assistants for conceptual and design questions rather than as exact table-reproducing engines~\cite{yanguas2025benchmarking}, and with MOF-ChemUnity, which couples a literature-derived knowledge base for metal–organic frameworks with LLM-driven assistance for retrieval and structure–property reasoning~\cite{pruyn2025mof}. In both cases, LLMs are most convincing when they sit on top of curated knowledge, not when they are the sole source of quantitative truth.

On the symbolic side, our work also connects to large, literature-derived knowledge graphs in materials science such as MatKG~\cite{venugopal2024matkg}, which autonomously extracts entities and relations from millions of materials papers and exposes them as an RDF/CSV knowledge graph queryable via SPARQL. While MatKG emphasises broad, automatically mined coverage, our ORKG-based comparisons focus on a smaller set of review tables with explicit schema design, expert curation, and process-level detail tailored to ALD/ALE. Together, these efforts illustrate a spectrum of symbolic infrastructures: from wide, automatically generated graphs to high-fidelity, machine-actionable review tables that can serve as gold-standard targets for precise querying and future neurosymbolic integration.

In light of this, a natural next step is to explore \emph{closer integrations of symbolic knowledge with neural models}, given the case for symbolic knowledge made in this paper. One direction would be to train or fine-tune LLMs directly on symbolic artefacts—comparison tables, ORKG triples, and SPARQL query–result pairs—so that they become better at precise, table-like operations while still supporting trend extrapolation and hypothesis generation (for example, suggesting candidate precursors, process windows, or follow-up experiments). For such approaches to be effective, however, the underlying symbolic data must be both \emph{verified} and \emph{large-scale}. Since current LLMs are already trained on massive amounts of natural language text, the bottleneck shifts to constructing sufficiently rich gold-standard symbolic datasets. This calls for workflows that can mine machine-actionable knowledge from large collections of papers with very high accuracy (ideally well above 95\% for table reconstruction and key fields), combining automatic extraction with targeted human validation. Prospectively, we therefore invite the ALD/ALE community to publish machine-actionable versions of their review tables—alongside the traditional PDF—as part of a growing gold-standard set (for example in ORKG and linked repositories). Another complementary direction is to design hybrid systems in which a deterministic symbolic core (e.g., a SPARQL engine over ORKG) remains responsible for exact filtering and aggregation, while neural components help with formulating natural-language queries, ranking and clustering results, and explaining patterns. Across these scenarios, our findings point to a clear design principle for ALD/ALE workflows: reliable, quantitative querying should remain anchored in symbolic substrates such as ORKG, with neural models used as powerful but approximate assistive layers on top—well suited for exploration, explanation, and idea generation, but not treated as standalone sources of truth for process-critical numbers.

\section{Conclusion}
This work showed how ALD/ALE review knowledge can be transformed from static PDF tables into a machine-actionable layer that supports precise, repeatable querying. From nine recent review articles, we converted 18 multi-study process tables into ORKG \emph{comparisons} and defined 33 natural-language questions with corresponding SPARQL queries, releasing them as the ALD/E Query Dataset~\cite{dsouza_ald-e_2025}. A survey with three ALD/ALE experts confirmed that these machine-actionable reviews are scientifically meaningful and practically useful for tasks such as precursor screening, process-window comparison, and mechanism overview, while also highlighting concrete opportunities to enrich schemas and add missing parameters. Comparing three ways of answering the same questions—symbolic querying via SPARQL over ORKG, purely neural querying over review PDFs, and neural querying grounded in ORKG comparison tables—shows a clear division of roles: SPARQL over ORKG remains the gold standard for exact, reproducible answers with transparent provenance, whereas large language models are best used as assistive layers on top, especially when grounded in structured ORKG tables, for formulating questions, exploring patterns, and generating hypotheses.

Looking ahead, our results argue for growing and tightening the connection between symbolic and neural tools. On the symbolic side, we see a need for community workflows that routinely publish machine-actionable versions of review tables alongside PDFs, creating large, verified corpora that can serve both as gold-standard resources and as training material for future neurosymbolic models. On the neural side, hybrid systems in which a deterministic symbolic core remains responsible for exact filtering and aggregation, while LLMs handle query formulation, result exploration, and explanation, appear particularly promising. Finally, this work complements emerging efforts on multimodal scientific AI, where models are expected to interpret not only text and tables but also the figures that carry much of the quantitative insight in ALD/ALE and related fields~\cite{d_souza_2025_17130928}. Taken together, these directions outline a path toward scientific assistants that can reliably answer precise questions, help make sense of complex datasets, and eventually reason across text, tables, and images in an integrated way.

\begin{ack}
Experiments with the five open-weights LLMs were supported by the KISSKI HPC infrastructure~\cite{doosthosseiniSAIASeamlessSlurmNative2025}, in particular via the ChatAI API service (\url{https://kisski.gwdg.de/en/leistungen/2-02-llm-service/}). This centralized service provides access to very large language models with billions of parameters, whose computational demands often exceed the local GPU resources typically available in academic research environments, and thereby alleviates the need for redundant deployments and time-consuming ad hoc querying setups.

This work is supported by the program “AI-Aware Pathways to Sustainable Semiconductor Process and Manufacturing Technologies”, co-sponsored by Intel Corporation \& Merck KGaA, Darmstadt, Germany.
\end{ack}

\section*{Data Availability Statement}

The code accompanying this article is released on Github at \url{https://github.com/sciknoworg/ald-ale-orkg-review} with detailed documentation. The dataset produced and studied in this research is released on Zenodo \cite{dsouza_ald-e_2025} and be accessed at this DOI \url{https://doi.org/10.5281/zenodo.17720429}.

\bibliographystyle{plainnat} 
\bibliography{citation}    


\appendix

\section*{Appendix}

\section{Experimental Setting 2 -- Detailed Technical Description}
\label{appsec:setting2}

In this setting, we evaluate the neural upper bound of performance by treating the PDF of each review article as the only source of evidence. Both proprietary and open-weights language models receive the detailed natural-language equivalent of the SPARQL query together with the article PDF (or its extracted text and tables), but \emph{no} direct access to the ORKG representations. The models must therefore internally parse the PDF, locate the relevant passages and tables, and aggregate the required quantities in a single neural inference step. By comparing their outputs against the SPARQL gold standard from Setting~1, we can quantify how closely purely neural, PDF-based querying can approach symbolic performance for precise, table-style scientific questions.

\begin{enumerate}[label=A.\arabic*]
    \item \textit{Proprietary language model (neural) querying.} Here we use the proprietary chat interfaces of three state-of-the-art conversational generative AI models—ChatGPT-5.1 (OpenAI, release 12.11.2025)\cite{chatgpt51}, Gemini-3-Pro (Google, release 18.11.2025)\cite{gemini3}, and Claude-Sonnet-4.5 (Anthropic, release 29.09.2025)\cite{anthropic2025sonnet45}. As of the date of this submission (13.12.2025), these systems represent the most capable commercially hosted LLMs. Each model is provided with the detailed natural-language equivalent of the SPARQL query and the PDF of the review article. This setting tests whether proprietary     models, operating as black-box systems with undisclosed PDF-processing workflows, can directly answer precise scientific questions from the PDFs alone. However, proprietary models provide limited experimental control: their internal PDF-parsing pipelines are opaque, their context integration strategies cannot be inspected or modified, and their outputs cannot be reproduced or systematically debugged. To obtain full transparency, reproducibility, and fine-grained experimental control, we therefore complement this setting with open-weights models.
    
    \item \textit{Open-weights language model (neural) querying.} This setting mirrors Setting~2, but instead of proprietary models, we evaluate high-performing open-weights LLMs. We extract text and tabular content from each PDF using \href{https://github.com/jsvine/pdfplumber}{pdfplumber}, and provide that content as input to the models. Five open-weights models are evaluated which are: \href{https://huggingface.co/google/gemma-3-27b-it}{gemma-3-27b-it}, \href{https://huggingface.co/meta-llama/Llama-3.3-70B-Instruct}{llama-3.3-70b-instruct}, \href{https://huggingface.co/Qwen/Qwen2.5-VL-72B-Instruct}{qwen2.5-vl-72b-instruct}, \href{https://huggingface.co/Qwen/Qwen3-30B-A3B-Instruct-2507}{qwen3-30b-a3b-instruct-2507}, and \href{https://huggingface.co/mistralai/Mistral-Large-Instruct-2407}{mistral-large-instruct}. These models were selected due to their demonstrated strength in large-scale tabular reasoning benchmarks~\cite{cremaschi2025mammotab}.

    \begin{enumerate}[label=\alph*)]
        \item \textit{Context-injected prompting (single-shot document QA).} In this variant, the full text and all tables of each paper are extracted using \href{https://github.com/sciknoworg/ald-ale-orkg-review/blob/main/llm-experiments/pdf_extractor.py}{\texttt{pdf\_extractor.PDFExtractor}} and concatenated into one long context window. This complete article representation is passed directly to an open-weights model together with the detailed natural-language query, as implemented in \href{https://github.com/sciknoworg/ald-ale-orkg-review/blob/main/llm-experiments/context_injected_doc_qa.py}{\texttt{context\_injected\_doc\_qa.py}}. The model must answer solely from this unfiltered, full-document context. While this approach does not control noise or redundancy, it guarantees that tables remain intact and all evidence is available to the model.
        
        \item \textit{Retrieval-augmented generation (RAG) over PDF segments.} As in Setting~3.1, we first extract all text and tables using \href{https://github.com/sciknoworg/ald-ale-orkg-review/blob/main/llm-experiments/pdf_extractor.py}{\texttt{PDFExtractor}}. Instead of supplying the entire document to the model, we split the extracted text into contiguous segments of approximately 8\,000 characters. In our corpus, a segment typically corresponds to one to three PDF pages—long enough to contain most tables, though some wide or multi-page tables may still be split. Each segment is converted into a dense numerical vector (an ``embedding'') using a local Ollama encoder (default: \href{https://ollama.com/library/nomic-embed-text:latest}{\texttt{nomic-embed-text:latest}}), producing a per-paper vector index, as implemented in \href{https://github.com/sciknoworg/ald-ale-orkg-review/blob/main/llm-experiments/rag_pdf_segments_doc_qa.py}{\texttt{rag\_pdf\_segments\_doc\_qa.py}}.
    
        At query time, the detailed natural-language query is embedded with the same encoder, and cosine similarity \cite{singhal2001modern} is computed against all segment embeddings. We retrieve the single top-ranked segment (\texttt{TOP\_K = 1}) and expand the selection to include neighboring segments in index order, which helps reconstruct table content that may straddle segment boundaries. Only these retrieved segments, together with the detailed query, are provided to the model.
    
        \textit{Potential limitations.}
        Because RAG introduces an explicit retrieval stage, its behavior depends not only on the language model but also on how we index each paper and select segments for a given query. In our implementation, these design choices introduce several potential failure modes:
        
        \begin{itemize}
            \item \textit{Segmentation effects.} The formatted PDF output of \texttt{PDFExtractor} is split into contiguous 8k-character segments, which correspond roughly to a few pages of text and tables. We use zero overlap between segments (\texttt{CHUNK\_OVERLAP = 0}) to avoid duplicating table rows across chunks, but this also means that large review tables can be cut exactly at a segment boundary. Although the retrieval step always expands the top retrieved segment to include its two immediate neighbors on each side, tables that span more than this local window, or that reappear as ``continued'' tables much later in the paper, may still be only partially visible in the model's context.
        
            \item \textit{Embedding sensitivity.} Dense retrieval is driven by an embedding model (\texttt{nomic-embed-text}) that maps each segment and each query into a high-dimensional numerical vector so that semantically similar texts lie close to each other. Retrieval quality therefore depends on how well this general-purpose encoder captures specialised ALD/ALE terminology and fine-grained distinctions between process variants, precursor chemistries, or temperature regimes. If the embedding space does not separate these concepts cleanly, segments that contain the relevant table rows may receive lower similarity scores than less relevant segments.
        
            \item \textit{Top-$k$ retrieval and local windowing.}
            For each query we retrieve only the single most similar segment (\texttt{TOP\_K = 1}) and then expand around this ``anchor'' by adding its immediate neighbours. This favours a compact context window and keeps the RAG setting comparable to the full-context Setting~3.1, but it also assumes that most of the evidence the model needs is concentrated in one contiguous region of the paper. Queries that implicitly refer to several distant regions (for example, all SiO\textsubscript{2} processes in a table that is split across pages, or a table plus an explanatory paragraph elsewhere in the text) may require information scattered across non-adjacent segments that this single-anchor strategy never retrieves together.
        
            \item \textit{Loss of global document context.} In contrast to Setting~3.1, where the model sees the entire extracted paper at once, the RAG variant deliberately restricts the model to a narrow slice of the document and instructs it to answer \emph{only} from the retrieved segments. For table-oriented queries that require aggregating over all rows, comparing multiple materials, or cross-checking constraints stated in different parts of the article, this narrow window can omit complementary rows or conditions that happen to fall outside the retrieved neighbourhood. In such cases, the retrieval stage may remove crucial context that the full-context setting would have made available.
        \end{itemize}
          
        This setting thus tests a retrieval-augmented approach designed to reduce context length and emphasize relevant evidence, while also introducing retrieval- and segmentation-dependent behavior that may influence downstream performance.
    \end{enumerate}
    
\end{enumerate}

\section{Evaluation Metrics for the Neural Q\&A Tabular Results} 
\label{appsec:rms}

To evaluate RQ2 and RQ3, we measure how closely the tables produced by the neural systems match the SPARQL gold-standard tables on a per-query basis. Intuitively, we want a metric that (i) is insensitive to permutations of rows and columns, (ii) distinguishes between minor numeric or textual deviations and completely wrong entries, and (iii) exposes both \emph{precision} (spurious rows or cells) and \emph{recall} (missing rows or cells). 
We adopt the \emph{Relative Mapping Similarity} (RMS) metric introduced by Liu et al.~\cite{liu-etal-2023-deplot} as part of Google research, which is specifically designed for table-level evaluation of chart and table question answering systems. 
Our implementation of the RMS-based evaluation, built on top of the official DePlot metrics, is available at \url{https://github.com/sciknoworg/ald-ale-orkg-review/blob/main/eval_deplot_rms/README.md}. 
This implementation documents how we adapt Google's DePlot code (\url{https://github.com/google-research/google-research/tree/master/deplot}) to our ALD/ALE SPARQL and neural querying results dataset provided as the Zenodo dump \cite{dsouza_ald-e_2025}.

Following Liu et al., RMS builds on the earlier Relative Number Set Similarity (RNSS) metric \cite{luo2021chartocr,masry-etal-2022-chartqa}, which only compares the unordered set of numeric values in the predicted table against those in the gold table. Let $P = \{p_i\}_{i=1}^N$ and $T = \{t_j\}_{j=1}^M$ denote the predicted and target number sets, respectively. The relative numeric error between two numbers is
\begin{equation}
  D(p, t) = \min\!\left( 1, \frac{\lvert p - t \rvert}{\lvert t \rvert} \right).
\end{equation}
Given the $N \times M$ matrix of pairwise distances $D(p_i, t_j)$, a minimum-cost bipartite matching $X \in \{0,1\}^{N \times M}$ is found. RNSS is then defined as
\begin{equation}
  \mathrm{RNSS}
  = 
  1 - \frac{\displaystyle\sum_{i=1}^N \sum_{j=1}^M X_{ij} \, D(p_i, t_j)}{\max(N, M)}.
\end{equation}
While RNSS is useful as a numeric sanity check, it ignores the \emph{position} of values in the table (which row/column they belong to), discards all non-numeric content, and does not distinguish precision from recall.

RMS instead views a table as an unordered collection of \emph{mappings} from row and column headers to a cell value. Each entry in the predicted table is represented as $p_i = (p^r_i, p^c_i, p^v_i)$, where $p^r_i$ and $p^c_i$ are the row and column headers, and $p^v_i$ is the cell value; the gold table is represented analogously as $t_j = (t^r_j, t^c_j, t^v_j)$. To compare two entries $p_i$ and $t_j$, RMS combines (i) a textual similarity between the concatenated keys $p^r_i \Vert p^c_i$ and $t^r_j \Vert t^c_j$, and (ii) a similarity between the cell values $p^v_i$ and $t^v_j$.

Textual similarity is measured with a normalized Levenshtein distance $NL_\tau(\cdot,\cdot)$ with threshold $\tau$, which clips large edit distances to $1$:
\begin{equation}
  s_{\text{key}}(i,j)
  =
  1 - NL_\tau\bigl(p^r_i \Vert p^c_i,\; t^r_j \Vert t^c_j\bigr),
\end{equation}
so that $s_{\text{key}}(i,j) \in [0,1]$. Numeric values are compared using the relative error with a second threshold~$\theta$; the corresponding similarity is
\begin{equation}
  s_{\text{val}}(i,j)
  =
  1 - \min\!\left( 1, \frac{\lvert p^v_i - t^v_j \rvert}{\lvert t^v_j \rvert} \right),
\end{equation}
again clipped at $0$ for very large discrepancies. When a value is non-numeric, $s_{\text{val}}$ falls back to a textual similarity term using $NL_\tau$. The overall similarity between two mappings is then
\begin{equation}
  s(i,j) = s_{\text{key}}(i,j) \, s_{\text{val}}(i,j),
\end{equation}
which is close to $1$ only if both the key (row+column) and the associated value match well.

To compare a predicted table $P = \{p_i\}_{i=1}^N$ with a gold table $T = \{t_j\}_{j=1}^M$, RMS first constructs an $N \times M$ cost matrix based on the key similarity and solves a minimum-cost bipartite matching problem to obtain a binary assignment matrix $X \in \{0,1\}^{N \times M}$. Using this matching, RMS precision and recall are defined as
\begin{subequations}
\begin{eqnarray}
  \mathrm{RMS}_{\text{precision}}
  &=& \frac{1}{N} \sum_{i=1}^N \sum_{j=1}^M X_{ij} \, s(i,j),
  \\
  \mathrm{RMS}_{\text{recall}}
  &=& \frac{1}{M} \sum_{i=1}^N \sum_{j=1}^M X_{ij} \, s(i,j),
  \\
  \mathrm{RMS}_{\mathrm{F1}}
  &=& \frac{2 \cdot \mathrm{RMS}_{\text{precision}} \cdot \mathrm{RMS}_{\text{recall}}}
          {\mathrm{RMS}_{\text{precision}} + \mathrm{RMS}_{\text{recall}}}.
\end{eqnarray}
\end{subequations}
Because RMS treats tables as unordered sets of $(\text{row}, \text{column}, \text{value})$ mappings and considers both the original and transposed layouts, it is by construction invariant to row and column permutations and to table transpositions.

In our experiments, for each query we take the SPARQL result table as the gold table $T$ and the output of each neural system (neural-only and symoblic context-augmented neural settings) as the predicted table $P$. We compute RMS precision, recall, and F1 using the original DePlot implementation's default thresholds (for textual and numeric similarity), as exposed via our evaluation script. We report per-query RMS scores as well as macro-averaged RMS across queries for each system and evaluation setting, thereby quantifying both how many of the gold rows and cells are recovered (recall) and how many of the predicted entries genuinely correspond to SPARQL-derived evidence (precision).

\section{Results -- Detailed analysis of domain expert impressions (RQ1)}
\label{appsec:stats-for-survey}

Here we provide a more detailed account of the expert evaluation results for RQ1. This was based on the survey taken by 3 domain experts wtih the following outcome.

\paragraph{Meaningfulness.}
Across all 33 queries, the questions themselves were judged to be scientifically well-posed. The mean meaningfulness score was $4.54/5$, with 96\% of all ratings at~$\geq 4$, and 32 out of 33 queries achieving an average meaningfulness score of at least~4. ALE queries were rated slightly higher (mean $4.69$) than ALD queries (mean $4.42$), and all ALE ratings fell in the range 4--5. As visualised in Fig.~\ref{fig:meaningful} (see also the per-domain panels in Figs.~\ref{fig:ALD-meaningful} and~\ref{fig:ALE-meaningful}), individual per-query, per-annotator points cluster tightly in the upper part of the scale, with only a few isolated ratings below 3 (e.g., for ALD Q11). This indicates that the query set, as a whole, aligns well with the experts' sense of plausible relationships, appropriate concepts, and relevant use cases in ALD and ALE.

Among the ALD queries, six achieved perfect mean scores of $5.00$ (Q09, Q10, Q13, Q14, Q17, Q18), while the lowest-rated query was Q11 with a mean of $3.67$---the only query across either domain to fall below $4.00$. For ALE, four queries achieved perfect scores of $5.00$ (Q02, Q05, Q07, Q08), and the lowest-rated queries (Q10, Q12, Q13) still attained a mean of $4.33$. At the individual annotator level, mean ratings ranged from $4.21$ to $5.00$ for ALD and from $4.29$ to $4.93$ for ALE, reflecting moderate variation in stringency across evaluators.

Formal inter-annotator agreement, measured via Fleiss' $\kappa$ and Krippendorff's $\alpha_{\text{ordinal}}$, was low (ALD: $\kappa = 0.021$, $\alpha_{\text{ordinal}} = 0.081$; ALE: $\kappa = -0.114$, $\alpha_{\text{ordinal}} = -0.088$). This is a known artefact when ratings are strongly skewed toward one end of the scale: when nearly all scores are 4 or 5, even small disagreements (e.g., 4 vs.\ 5) can produce low or negative $\kappa$ values despite broad substantive consensus. In our case, the high proportion of ratings at 4–5, combined with high exact-match rates (35.7\%–78.6\% across annotator pairs) and full three-way agreement on 28.6\% (ALE) to 36.8\% (ALD) of queries, suggests that annotators agreed that the queries were meaningful, even if they differed slightly in how enthusiastically they expressed this via the 5-point scale.

\paragraph{Usefulness.}
Experts also rated the usefulness of the SPARQL result tables as synthesized knowledge artefacts. Over all queries, the average usefulness score was $4.08/5$, with 69\% of ratings at~$\geq 4$ and more than half at the maximum value of~5. On a per-query basis, 25 of 33 queries achieved a mean usefulness score of at least~4, and only two queries received an average usefulness below~3. ALD queries received a marginally higher usefulness score (mean $4.16$) than ALE queries (mean $3.98$). Figure~\ref{fig:useful} (panels~\ref{fig:ALD-useful} and~\ref{fig:ALE-useful}) shows that usefulness ratings are somewhat more dispersed than meaningfulness ratings, especially for a small subset of queries whose tables are sparse or missing key parameters, but the majority of points still lie in the 4--5 band.

Among ALD queries, twelve achieved perfect mean usefulness scores of $5.00$ (Q01, Q02, Q04, Q05, Q06, Q13, Q14, Q15, Q16, Q17, Q18, Q19), while the lowest-rated query was Q08 with a mean of $3.00$. For ALE, the highest-rated query was Q06 with a mean of $4.67$, and the lowest-rated queries were Q12 (mean $2.67$) and Q14 (mean $3.00$)---the only two queries across both domains to fall below $3.00$. At the individual annotator level, mean usefulness ratings ranged from $4.21$ to $5.00$ for ALD and from $2.93$ to $4.36$ for ALE, indicating greater variation in evaluator stringency for the usefulness criterion than for meaningfulness.

Formal inter-annotator agreement was again low: for ALD, Fleiss' $\kappa = -0.035$ and $\alpha_{\text{ordinal}} = 0.018$; for ALE, $\kappa = 0.151$ and $\alpha_{\text{ordinal}} = -0.041$. In the ALD case, these values are largely driven by a ceiling effect: two of three annotators assigned maximum scores to nearly all queries, so there is little variability for the statistics to capture, even though 63.2\% of queries show full three-way agreement. ALE usefulness ratings exhibit greater dispersion, with one annotator providing systematically lower scores (mean $2.93$) than the other two (means $4.36$ and $4.21$). As a result, full three-way agreement drops to 21.4\%. Pairwise agreement between the two higher-scoring annotators remains substantial (78.6\% exact match, Cohen's $\kappa = 0.576$), suggesting that low overall agreement is driven mainly by different baseline expectations for what constitutes a “useful’’ synthesis, rather than inconsistent application of the criterion.

Taken together, these quantitative results suggest that domain experts not only regard the questions as meaningful but also find the machine-actionable syntheses produced by the ORKG comparisons genuinely helpful for reasoning about processes, mechanisms, and material–performance trade-offs.

\paragraph{Qualitative feedback.}
Free-text explanations for lower usefulness ratings point primarily to limitations in the current data modelling and coverage rather than to the concept of machine-actionable reviews itself. Experts frequently asked for additional parameters (e.g., sticking coefficients, surface orientation and termination, clearer precursor/co-reactant distinctions, explicit dopant columns, error bars or standard deviations on growth-per-cycle, and more detailed timing parameters per ALD/ALE cycle), more consistent categorisation of mechanisms and precursor families, and improved result presentation (e.g., removing duplicate rows, compacting very long tables, or explaining derived indices such as the efficiency metric in the rare-earth MOSLED example). In several cases, low scores were attributed to incomplete or sparsely populated ORKG comparisons, or to queries that, while technically correct, remained too generic to support concrete decision-making. This pattern indicates that dissatisfaction is concentrated on specific modelling choices and missing annotations, not on the idea of querying symbolic review knowledge.

The open-ended questions corroborate this interpretation and highlight both enthusiasm and pragmatic concerns. When asked to rate the overall usefulness of machine-actionable scientific knowledge, all three experts gave scores between 4 and 5 (mean $4.67$), emphasising the value of faster access to synthesized insights and the ability to compare conditions across papers instantly instead of manually reconstructing spreadsheets from multiple reviews. Their “wish-list’’ questions extend this theme: examples include queries over the most common surface orientations and terminations for a given material, multi-parameter aggregations over ALD recipes grouped by precursor/co-reactant, temperature, plasma power, and reactor type, or systematic overviews of process windows and performance metrics for specific device architectures. In other words, experts immediately project the ORKG comparisons onto richer, multi-dimensional queries that mirror the way they already reason about processes and materials.

At the same time, experts foresee several barriers to broader adoption. These include (i) the additional time and effort needed to enter machine-actionable tables alongside traditional PDFs, (ii) uneven reporting practices in the literature, which limit what can be modelled consistently, (iii) concerns about data quality and trust if tables were to be (partially) auto-extracted, and (iv) the lack of established community standards in ALD/ALE for which process parameters and performance metrics should be captured. On the positive side, they identify clear incentives that would motivate them to use or publish such tables: demonstrable time savings in literature review, community-wide standardisation of schemas, user-friendly tooling for authoring and querying ORKG comparisons, and clear benefits for their own workflows (e.g., being able to answer complex comparative questions “in one click’’).

\end{document}